\documentclass[runningheads]{llncs}

\pdfoutput=1
\usepackage{grffile}

\usepackage{graphicx}
\usepackage{comment}
\usepackage{amsmath,amssymb} %
\usepackage{color}
\usepackage{comment}

\def\bfg{{\boldsymbol{g}}} 
\def\bfR{{\boldsymbol{R}}} 
\def\bft{{\boldsymbol{t}}} 
\def\bfT{{\boldsymbol{T}}}

\def\bfp{{\boldsymbol{p}}}

\usepackage{pdfpages}
\usepackage{afterpage}
\usepackage{pdflscape}
\usepackage{subcaption}

\newcommand{\etal}{\textit{et al.}}

\begin{document}

\pagestyle{headings}
\mainmatter

\def\ECCVSubNumber{2616} 
\title{DEMEA: Deep Mesh Autoencoders for Non-Rigidly Deforming Objects}

\titlerunning{DEMEA: Deep Mesh Autoencoders}

\author{Edgar Tretschk\inst{1}%
\hspace{15pt}
Ayush Tewari\inst{1}%
\\
Michael Zollh\"ofer\inst{2}
\hspace{10pt}
Vladislav Golyanik\inst{1}%
\hspace{10pt}
Christian Theobalt\inst{1}
} 
\authorrunning{E. Tretschk et al.}
\institute{Max Planck Institute for Informatics, Saarland Informatics Campus%
\and
Stanford University
}

\maketitle

\setcounter{footnote}{0}
\begin{abstract} 
   Mesh autoencoders are commonly used for dimensionality reduction, sampling and mesh modeling.
   We propose a general-purpose DEep MEsh Autoencoder \hbox{(DEMEA)} which adds a novel embedded deformation layer to a graph-convolutional mesh autoencoder.
   The embedded deformation layer (EDL) is a differentiable deformable geometric proxy which explicitly models point displacements of non-rigid deformations in a lower dimensional space and serves as a local rigidity regularizer.
   DEMEA decouples the parameterization of the deformation from the final mesh resolution since the deformation is defined over a lower dimensional embedded deformation graph.
   We perform a large-scale study on four different datasets of deformable objects.
   Reasoning about the local rigidity of meshes using EDL allows us to achieve higher-quality results for highly deformable objects, compared to directly regressing vertex positions.
   We demonstrate multiple applications of DEMEA, including non-rigid 3D reconstruction from depth and shading cues, non-rigid surface tracking, as well as the transfer of deformations over different meshes. 
\end{abstract}

\section{Introduction}\label{sec:intro} 
\begin{figure}
\begin{center}
\includegraphics*[width=1.0\linewidth]{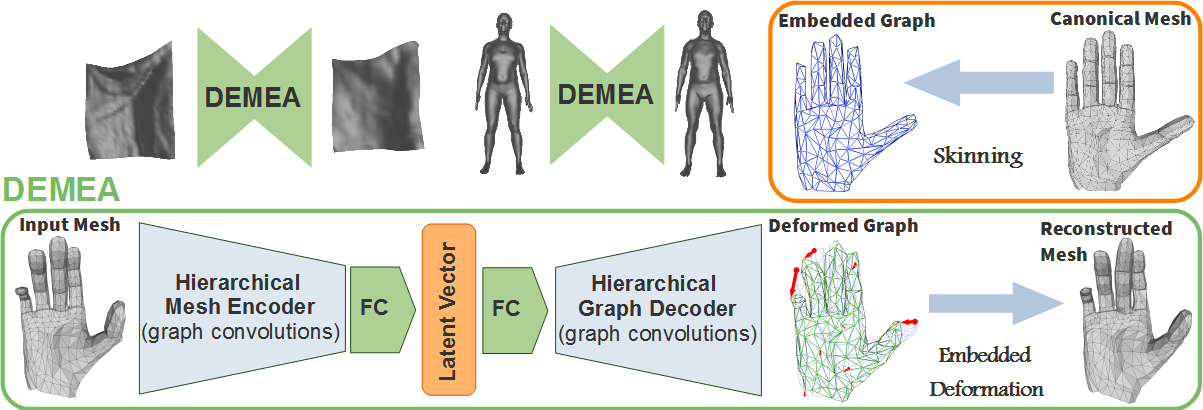}
\caption{Pipeline: DEMEA encodes a mesh using graph convolutions on a mesh hierarchy. The graph decoder first maps the latent vector to node features of the coarsest graph level. A number of upsampling and graph convolution modules infer the node translations and rotations of the embedded graph. An embedded deformation layer applies the node translations to a template graph, against which a template mesh is skinned. With the node rotations and the skinning, this deformed graph allows reconstructing a deformed mesh. %
}
\label{fig:pipeline}
\end{center} 
\end{figure}
With the increasing volume of datasets of deforming objects enabled by modern 3D acquisition technology, the demand for compact data representations and compression grows. Dimensionality reduction of mesh data has multiple applications in computer graphics and vision, including shape retrieval, generation, interpolation, and completion. Recently, deep convolutional autoencoder networks were shown to produce compact mesh representations~\cite{Bagautdinov2018,Tan2018AAAI,Ranjan2018,Bouritsas2019}. 

Dynamic real-world objects do not deform arbitrarily.
While deforming, they preserve topology, and nearby points are more likely to deform similarly compared to more distant points.
Current convolutional mesh autoencoders exploit this coherence by learning the deformation properties of objects directly from data and are already suitable for mesh compression and representation learning.
On the other hand, they do not explicitly reason about the deformation field in terms of local rotations and translations.
We show that explicitly reasoning about the local rigidity of meshes enables higher-quality results for highly deformable objects, compared to directly regressing vertex positions.

At the other end of the spectrum, mesh manipulation techniques such as As-Rigid-As-Possible Deformation~\cite{SorkineAlexa2007} and Embedded Deformation~\cite{Sumner2007} only require a single mesh and enforce deformation properties, such as smoothness and local rigidity, based on a set of hand-crafted priors.
These hand-crafted priors are effective and work surprisingly well, but since they do not model the real-world deformation behavior of the physical object, they often lead to unrealistic deformations and artifacts in the reconstructions.

In this paper, we propose a general-purpose mesh autoencoder with a model-based deformation layer, combining the best of both worlds, \textit{i.e.,} supervised learning with deformable meshes and a novel \textit{differentiable embedded deformation} layer that models the deformable meshes using lower-dimensional deformation graphs with physically interpretable deformation parameters. While the core of our DEep MEsh Autoencoder (DEMEA) learns the deformation model of objects from data using the state-of-the-art convolutional mesh autoencoder (CoMA) \cite{Ranjan2018}, the novel embedded deformation layer decouples the parameterization of object motion from the mesh resolution and introduces local spatial coherence via vertex skinning.
DEMEA is trained on mesh datasets of moderate sizes that have recently become available \cite{Loper2014,Bogo2017,Bednarik2018,Malik2018}. DEMEA is a general mesh autoencoding approach that can be trained for any deformable object class. 
We evaluate our approach on datasets of three objects with large deformations like articulated deformations  (body, hand) and large non-linear deformations (cloth), and one object with small localized deformations (face).  
Quantitatively, DEMEA outperforms standard convolutional mesh autoencoder architectures in terms of vertex-to-vertex distance error. 
Qualitatively, we show that DEMEA produces visually higher fidelity results due to the physically based embedded deformation layer. 
We show several applications of DEMEA in computer vision and graphics. Once trained, the decoder of our autoencoders can be used for shape compression, high-quality depth-to-mesh reconstruction of human bodies and hands, and even poorly textured RGB-image-to-mesh reconstruction for deforming cloth.
The low-dimensional latent space learned by our approach is meaningful and well-behaved, which we demonstrate by different applications of latent space arithmetic.
Thus, DEMEA provides us a well-behaved general-purpose category-specific generative model of highly deformable objects.

\section{Related Work}

\textbf{Mesh Manipulation and Tracking.} Our embedded deformation layer is inspired by as-rigid-as-possible modelling~\cite{SorkineAlexa2007} and the method of Sumner \textit{et al.}~\cite{Sumner2007} for mesh editing and manipulation. 
While these methods have been shown to be very useful for mesh manipulation in computer graphics, to the best of our knowledge, this is the first time a model-based regularizer is used in a mesh autoencoder.
Using a template for non-rigid object tracking from depth maps was extensively studied in the model-based setting \cite{Li2009,Zollhoefer2014}. 
Recently, Litany \textit{et al.}~\cite{Litany2018} demonstrated a neural network-based approach for the completion of human body shapes from a single depth map. %
\\ 
\textbf{Graph Convolutions. } %
The encoder-decoder approach to dimensionality reduction with neural networks (NNs) for images was introduced in \cite{HintonSalakhutdinov2006}.
Deep convolutional neural networks (CNNs) allow to effectively capture contextual information of input data modalities and can be trained for various tasks. 
Lately, convolutions operating on regular grids have been generalized to more general topologically connected structures such as meshes and two-dimensional manifolds \cite{Bruna2013,Niepert2016}, enabling learning of correspondences between shapes, shape retrieval \cite{Masci2015,Boscaini2016,Monti2017}, and segmentation \cite{Yi2017}.  
Masci \textit{et al.}~\cite{Masci2015} proposed geodesic CNNs operating on Riemannian manifolds for shape description, retrieval, and correspondence estimation.
Boscani \textit{et al.}~\cite{Boscaini2016} introduced spatial weighting functions based on simulated heat propagation and projected anisotropic convolutions. %
Monti \textit{et al.}~\cite{Monti2017} extended graph convolutions to variable patches through Gaussian mixture model CNNs. 
In FeaSTNet~\cite{Verma2018}, the correspondences between filter weights and graph neighborhoods with arbitrary connectivities are established dynamically from the learned features. 
The localized spectral interpretation of Defferrard \textit{et al.}~\cite{Defferrard2016} is based on recursive feature learning with Chebyshev polynomials and has linear evaluation complexity. 
Focusing on mesh autoencoding, Bouritsas \textit{et al.}~\cite{Bouritsas2019} exploited the fixed ordering of neighboring vertices.
\\
\textbf{Learning Mesh-Based 3D Autoencoders. } 
Very recently, several mesh autoencoders with various applications were proposed. 
A new hierarchical variational mesh autoencoder with fully connected layers for facial geometry parameterization learns an accurate face model from small databases and accomplishes depth-to-mesh fitting tasks \cite{Bagautdinov2018}. 
Tan and coworkers \cite{Tan2018} introduced a mesh autoencoder with a rotation-invariant mesh representation as a generative model. 
Their network can generate new meshes by sampling in the latent space and perform mesh interpolation.
To cope with meshes of arbitrary connectivity, they used fully-connected layers and did not explicitly encode neighbor relations. 
Tan \textit{et al.}~\cite{Tan2018AAAI} trained a network with graph convolutions to extract sparse localized deformation components from meshes. 
Their method is suitable for large-scale deformations and meshes with irregular connectivity. %
Gao \textit{et al.}~\cite{Gao2018} transferred mesh deformations by training a generative adversarial network with a cycle consistency loss to map shapes in the latent space, while a variational mesh autoencoder encodes deformations.
The Convolutional facial Mesh Autoencoder (CoMA) of Ranjan \textit{et al.}~\cite{Ranjan2018} allows to model and sample stronger deformations compared to previous methods and supports asymmetric facial expressions. 
The Neural 3DMM of Bouritsas \textit{et al.}~\cite{Bouritsas2019} improves quantitatively over CoMA due to better training parameters and task-specific graph convolutions.
Similar to CoMA~\cite{Ranjan2018}, our DEMEA uses spectral graph convolutions but additionally employs the embedded deformation layer as a model-based regularizer.
\\
\textbf{Learning 3D Reconstruction. } 
Several supervised methods reconstruct rigid objects in 3D. 
Given a depth image, the network of Sinha \textit{et al.}~\cite{Sinha2017} reconstructs the observed surface of non-rigid objects. 
In its 3D reconstruction mode, their method reconstructs rigid objects from single images. 
Similarly, Groueix \textit{et al.}~\cite{Groueix2018} reconstructed object surfaces from a point cloud or single monocular image with an atlas parameterization. 
The approaches of Kurenkov \textit{et al.}~\cite{Kurenkov2018} and Jack \textit{et al.}~\cite{Jack2018} deform a predefined object-class template to match the observed object appearance in an image. 
Similarly, Kanazawa \textit{et al.}~\cite{Kanazawa2018} deformed a template to match the object appearance but additionally support object texture. 
The Pixel2Mesh approach of Wang \textit{et al.}~\cite{Wang2018} reconstructs an accurate mesh of an object in a segmented image. 
Initializing the 3D reconstruction with an ellipsoid, their method gradually deforms it until the appearance matches the observation. 
The template-based approaches \cite{Kurenkov2018,Jack2018,Kanazawa2018}, as well as Pixel2Mesh \cite{Wang2018}, produce complete 3D meshes. %
\\
\textbf{Learning Monocular Non-Rigid Surface Regression. } %
Only a few supervised learning approaches for 3D reconstruction from monocular images tackle the deformable nature of non-rigid objects. 
Several methods \cite{Pumarola2018,Golyanik2018,shimada2019ismo} train networks for deformation models with synthetic thin plates datasets.
These approaches can infer non-rigid states of the observed surfaces such as paper sheets or membranes.
Still, their accuracy and robustness on real images are limited. 
Bedna\v{r}\'{i}k \textit{et al.}~\cite{Bednarik2018} proposed an encoder-decoder network for texture-less surfaces relying on shading cues.
They trained on a real dataset and showed an enhanced reconstruction accuracy on real images, but support only trained object classes. 
Fuentes-Jimenez \textit{et al.}~\cite{FuentesJimenez2018} trained a network to deform an object template for depth map recovery. 
They achieved impressive results on real image sequences but require an accurate 3D model of every object in the scene, which restricts the method's practicality. 
One of the applications of DEMEA is the recovery of texture-less surfaces from RGB images. 
Since a depth map as a data modality is closer to images with shaded surfaces,
we train DEMEA in the depth-to-mesh mode on images instead of depth maps. 
As a result, we can regress surface geometry from shading cue.

\section{Approach}\label{sec:approach} 
In this section, we describe the architecture of the proposed DEMEA. %
We employ an embedded deformation layer to decouple the complexity of the learned deformation field from the actual mesh resolution.
The deformation is represented relative to a canonical mesh $\mathcal{M} = (\mathbf{V}, \mathbf{E})$ with $N_v$ vertices $\mathbf{V} = \{\mathbf{v}_i \}_{i=1}^{N_v}$, and edges $\mathbf{E}$.
To this end, we define the encoder-decoder on a coarse deformation graph and use the embedded deformation layer to drive the deformation of the final high-resolution mesh, see Fig.~\ref{fig:pipeline}.
Our architecture is based on graph convolutions that are defined on a multi-resolution mesh hierarchy. 
In the following, we describe all components in more detail.
We describe the employed spiral graph convolutions \cite{Bouritsas2019} in the supplemental document.

\subsection{Mesh Hierarchy}
The up- and downsampling in the convolutional mesh autoencoder is defined over a multi-resolution mesh hierarchy, similar to the CoMA \cite{Ranjan2018} architecture.
We compute the mesh hierarchy fully automatically based on quadric edge collapses \cite{Garland:1997}, \textit{i.e.,} each hierarchy level is a simplified version of the input mesh.
We employ a hierarchy with five resolution levels, where the finest level is the mesh.
Given the multi-resolution mesh hierarchy, we define up- and downsampling operations \cite{Ranjan2018} for feature maps defined on the graph.
To this end, during downsampling, we enforce the nodes of the coarser level to be a subset of the nodes of the next finer level.
We transfer a feature map to the next coarser level by a similar subsampling operation.
The inverse operation, \textit{i.e.,} feature map upsampling, is implemented based on a barycentric interpolation of close features.
During edge collapse, we project each collapsed node onto the closest triangle of the coarser level.
We use the barycentric coordinates of this closest point with respect to the triangle's vertices to define the interpolation weights.

\subsection{Embedded Deformation Layer (EDL)} \label{ssec:EDL} 
Given a canonical mesh, we have to pick a corresponding coarse embedded deformation graph.
We employ MeshLab's \cite{Meshlab} more sophisticated implementation of quadric edge collapse to fully automatically generate the graph. See the supplemental document for details.
\begin{figure}
\begin{center}
\includegraphics*[width=\linewidth]{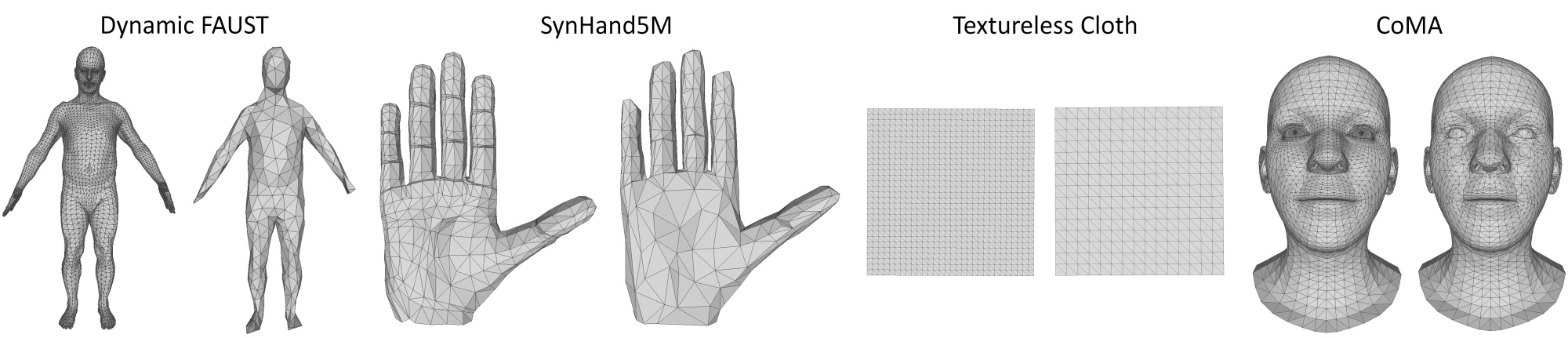}
\caption{Template mesh and the corresponding embedded deformation graph pairs automatically generated using \cite{Meshlab}.%
}
\label{fig:datasetGraphs}
\end{center} 
\end{figure}
The deformation graph is used as one of the two levels immediately below the mesh in the mesh hierarchy (depending on the resolution of the graph) of the autoencoder.
When generating the mesh hierarchy, we need to enforce the subset relationship between levels. However, the quadric edge collapse algorithm of \cite{Ranjan2018} might delete nodes of the embedded graph when computing intermediate levels between the mesh and the embedded graph. We ensure that those nodes are not removed by setting the cost of removing them from levels that are at least as fine as the embedded graph to infinity.

Our embedded deformation layer models a space deformation that maps the vertices of the canonical template mesh $\mathbf{V}$ to a deformed version $\hat{\mathbf{V}}$.
Suppose $\mathcal{G} = (\mathbf{N}, \mathbf{E})$ is the embedded deformation graph \cite{Sumner2007} with $L$ canonical nodes $\mathbf{N} = \{\bfg_l\}_{i=1}^{L}$ and $K$ edges $\mathbf{E}$, with $\bfg_l \in \mathbb{R}^3$.
The global space deformation is defined by a set of local, rigid, per-graph node transformations.
Each local rigid space transformation is defined by a tuple $\bfT_l = (\bfR_l, \bft_l)$, with $\bfR_l \in \mathbf{SO}(3)$ being a rotation matrix and $\bft_l \in \mathbb{R}^3$ being a translation vector.
We enforce that $\bfR_l^\mathsf{T} = \bfR_l^{-1}$ and $\operatorname{det}(\bfR_l) = 1$ by parameterizing the rotation matrices based on three Euler angles.
Each $\bfT_l$ is anchored at the canonical node position $\bfg_l$ and maps every point $\bfp \in \mathbb{R}^3$ to a new position in the following manner \cite{Sumner2007}:
\begin{equation}\label{eq:deformation_layer} 
  \bfT_l(\bfp) =  \bfR_l [\bfp - \bfg_l] + \bfg_l + \bft_l. 
\end{equation} 
To obtain the final global space deformation $\mathbf{G}$, the local per-node transformations are linearly combined:
\begin{equation}
  \mathbf{G}(\bfp) = \sum_{l \in \mathcal{N}_\bfp }{w_{l}(\bfp) \cdot \bfT_l(\bfp)} \enspace{.}
\end{equation}
Here, $\mathcal{N}_\bfp$ is the set of approximate closest deformation nodes. 
The linear blending weights ${w_{l}}(\bfp)$ for each position are based on the distance to the respective deformation node \cite{Sumner2007}. 
Please refer to the supplemental for more details. 

The deformed mesh $\hat{\mathbf{V}} = \mathbf{G}(\mathbf{V})$ is obtained by applying the global space deformation to the canonical template mesh $\mathbf{V}$.
The free parameters are the local per-node rotations $\bfR_l$ and translations $\bft_l$, \textit{i.e.,} $6L$ parameters with $L$ being the number of nodes in the graph.
These parameters are input to our deformation layer and are regressed by the graph convolutional decoder.

\subsection{Differentiable Space Deformation} 
Our novel EDL is fully differentiable and can be used during network training to decouple the parameterization of the space deformation from the resolution of the final high-resolution output mesh.
This enables us to define the reconstruction loss on the final high-resolution output mesh and backpropagate the errors via the skinning transform to the coarse parameterization of the space deformation.
Thus, our approach enables finding the best space deformation by only supervising the final output mesh.

\subsection{Training}
We train our approach end-to-end in Tensorflow \cite{tensorflow2015-whitepaper} using Adam \cite{kingma:adam}.
As loss we employ a dense geometric per-vertex $\ell_1$-loss with respect to the ground-truth mesh.
For all experiments, we use a learning rate of $10^{-4}$ and default parameters $\beta_1=0.9$, $\beta_2=0.999$, $\epsilon=10^{-8}$ for Adam.
We train for $50$ epochs for Dynamic Faust, $30$ epochs for SynHand5M, $50$ epochs for the CoMA dataset and $300$ epochs for the Cloth dataset.
We employ a batch size of $8$.

\subsection{Reconstructing Meshes from Images/Depth}
The image/depth-to-mesh network consists of an image encoder and a mesh decoder, see Fig.~\ref{fig:pipeline2}.
The mesh decoder is initialized from the corresponding mesh auto-encoder, the image/depth encoder is based on a ResNet-50 \cite{HeZRS16} architecture, and the latent code is shared between the encoder and decoder.
We initialize the ResNet-50 component using pre-trained weights from ImageNet \cite{imagenet_cvpr09}.
To obtain training data, we render synthetic depth maps from the meshes.
We train with the same settings as for mesh auto-encoding.

\subsection{Network Architecture Details} \label{ssec:DEMEA_architecture} 
In the following, we provide more details of our encoder-decoder architectures. %
\\
\textbf{Encoding Meshes.}
Input to the first layer of our mesh encoder is an $N_v \times 3$ tensor that stacks the coordinates of all $N_v$ vertices.
We apply four \emph{downsampling modules}.
Each module applies a graph convolution and is followed by a downsampling to the next coarser level of the mesh hierarchy.
We use spiral graph convolutions~\cite{Bouritsas2019} and similarly apply an ELU non-linearity after each convolution.
Finally, we take the output of the final module and apply a fully connected layer followed by an ELU non-linearity to obtain a latent space embedding. %
\\
\textbf{Encoding Images/Depth.} 
To encode images/depth, we employ a 2D convolutional network to map color/depth input to a latent space embedding.
Input to our encoder are images of resolution $256 \times 256$ pixels.
We modified the ResNet-50 \cite{HeZRS16} architecture to take single or three-channel input image.
We furthermore added two additional convolution layers at the end, which are followed by global average pooling.
Finally, a fully connected layer with a subsequent ELU non-linearity maps the activations to the latent space. %
\\
\textbf{Decoding Graphs.} 
The task of the graph decoder is to map from the latent space back to the embedded deformation graph.
First, we employ a fully connected layer in combination with reshaping to obtain the input to the graph convolutional \emph{upsampling modules}.
We apply a sequence of three or four upsampling modules until the resolution level of the embedded graph is reached.
Each upsampling module first up-samples the features to the next finer graph resolution and then performs a graph convolution, which is then followed by an ELU non-linearity.
Then, we apply two graph convolutions with ELUs for refinement and a final convolution without an activation function.
The resulting tensor is passed to our embedded deformation layer.

\section{Experiments} 
We evaluate DEMEA quantitatively and qualitatively %
on several challenging datasets and demonstrate state-of-the-art results for mesh auto-encoding. 
In Sec.~\ref{sec:applications}, we show reconstruction from RGB images and depth maps and that the learned latent space enables well-behaved latent arithmetic.
We use Tensorflow 1.5.0 \cite{tensorflow2015-whitepaper} on Debian with an NVIDIA Tesla  V100 GPU.
\\
\textbf{Datasets.}
We demonstrate DEMEA's generality on experiments with body (Dynamic Faust, DFaust \cite{Bogo2017}), hand (SynHand5M \cite{Malik2018}), textureless cloth (Cloth \cite{Bednarik2018}), and face (CoMA \cite{Ranjan2018}) datasets. 
\begin{table}[]
\parbox{.42\linewidth}{
\centering
\scalebox{0.8}{
\begin{tabular}{l|l|l|l|l|l|}
\cline{2-6}
                                & Mesh & 1st           & 2nd          & 3rd            & 4th \\ \hline
\multicolumn{1}{|l|}{DFaust \cite{Bogo2017}}    & 6890 & 1723          & \textbf{431} & 108  & 27  \\ \hline
\multicolumn{1}{|l|}{CoMA \cite{Ranjan2018}}      & 5023 & \textbf{1256} & 314          & 79 & 20  \\ \hline
\multicolumn{1}{|l|}{SynHand5M \cite{Malik2018}} & 1193 & \textbf{299}  & 75          & 19  & 5   \\ \hline
\multicolumn{1}{|l|}{Cloth \cite{Bednarik2018}}     & 961  & \textbf{256}  & 100          & 36  & 16  \\ \hline
\end{tabular}}
\caption{Number of vertices on each level of the mesh hierarchy. Bold levels denote the embedded graph. Note that except for Cloth these values were computed automatically based on \cite{Ranjan2018}.
}
\label{tab:hierarchysizes} 
}
\hfill
\centering
\parbox{.55\linewidth}{
\scalebox{0.85}{
\begin{tabular}{l | l  l | l  l | l  l | l  l |}
 & \multicolumn{2}{|c|}{DFaust} & \multicolumn{2}{|c|}{SynHand5M} & \multicolumn{2}{|c|}{Cloth} & \multicolumn{2}{|c|}{CoMA}\\
\cline{2-9}
                                   & 8   & 32   & 8  & 32  & 8  & 32  & 8  & 32  \\ \hline
\multicolumn{1}{|l|}{CA}           & 6.35 & \textbf{2.07} &  8.12 & 2.60 &  \textbf{11.21} & 6.50 &  \textbf{1.17} & 0.72  \\ %
\multicolumn{1}{|l|}{MCA}          & \textbf{6.21} & 2.13 &  \textbf{8.11} & 2.67 &  11.64 & 6.59 &  1.20 & 0.71  \\ %
\multicolumn{1}{|l|}{Ours}         & 6.69 & 2.23 &  8.12 & \textbf{2.51} &  11.28 & 6.40 &  1.23 & 0.81  \\ %
\multicolumn{1}{|l|}{FCA}          & 6.51 & 2.17 &  15.10 & 2.95 &  15.63 & 5.99 &  1.77 & \textbf{0.67}  \\ %
\multicolumn{1}{|l|}{FCED}         & 6.26 & 2.14 &  14.61 & 2.75 &  15.87 & \textbf{5.94} &  1.81 & 0.73  \\ \hline
\end{tabular}}
\caption{Average per-vertex errors on the test sets of DFaust ($cm$), SynHand5M ($mm$), textureless cloth ($mm$) and CoMA ($mm$) for 8 and 32 latent dimensions.}
\label{tab:quantAuto}}
\end{table} 
Table \ref{tab:hierarchysizes} gives the number of graph nodes used on each level of our hierarchical architecture. 
All meshes live in metric space. %
\\
\textbf{DFaust \cite{Bogo2017}}. 
The training set consists of 28,294 meshes. 
For the tests, we split off two identities (female 50004, male 50002) and two dynamic performances, \textit{i.e.,} \textit{one-leg jump} and \textit{chicken wings}. 
Overall, this results in a test set with $12,926$ elements. 
For the depth-to-mesh results, we found the synthetic depth maps from the DFaust training set to be insufficient for generalization, \textit{i.e.,} the test error was high. 
Thus, we add more pose variety to DFaust for the depth-to-mesh experiments. 
Specifically, we add $28k$ randomly sampled poses from the CMU Mocap\footnote{\url{mocap.cs.cmu.edu}} dataset to the training data, where the identities are randomly sampled from the SMPL \cite{Loper2015} model (14$k$ female, 14$k$ male). 
We also add 12$k$ such samples to the test set (6$k$ female, 6$k$ male). %
\\
\textbf{Textureless Cloth \cite{Bednarik2018}.} 
For evaluating our approach on general non-rigidly deforming surfaces, we use the \textit{textureless cloth} data set of Bedna\v{r}\'{i}k \textit{et al.}~\cite{Bednarik2018}.
It contains real depth maps and images of a white deformable sheet ---  observed in different states and differently shaded --- as well as ground-truth meshes. 
In total, we select 3,861 meshes with consistent edge lengths.
3,167 meshes are used for training and $700$ meshes are reserved for evaluation.
Since the canonical mesh is a perfectly flat sheet, it lacks geometric features, which causes downsampling methods like \cite{Garland:1997}, \cite{Ranjan2018} and \cite{Meshlab} to introduce severe artifacts. Hence, we generate the entire mesh hierarchy for this dataset, see the supplemental. This hierarchy is also used to train the other methods for the performed comparisons. %
\\ 
\textbf{SynHand5M \cite{Malik2018}.} 
For the experiments with hands, we take $100k$ random meshes from the synthetic \textit{SynHand5M} dataset of Malik \textit{et al.}~\cite{Malik2018}. 
We render the corresponding depth maps. 
The training set is comprised of $90k$ meshes, and the remaining $10k$ meshes are used for evaluation. %
\\ 
\textbf{CoMA \cite{Ranjan2018}.}
The training set contains 17,794 meshes of the human face in various expressions \cite{Ranjan2018}. 
For tests, we select two challenging expressions, \textit{i.e.,} \textit{high smile} and \textit{mouth extreme}. 
Thus, our test set contains 2,671 meshes in total. 

\subsection{Baseline Architectures}\label{ssec:baseline_architectures} 
We compare DEMEA to a number of strong baselines. 
\\
\textbf{Convolutional Baseline. } 
We consider a version of our proposed architecture, \emph{convolutional ablation (CA)}, where the ED layer is replaced by learned upsampling modules that upsample to the mesh resolution. 
In this case, the extra refinement convolutions occur on the level of the embedded graph. 
We also consider \emph{modified CA (MCA)}, an architecture where the refinement convolutions are moved to the end of the network, such that they operate on mesh resolution. %
\\
\textbf{Fully-Connected Baseline.} 
We also consider an almost-linear baseline, \emph{FC ablation (FCA)}.
The input is given to a fully-connected layer, after which an ELU is applied. 
The resulting latent vector is decoded using another FC layer that maps to the output space. 
Finally, we also consider an \emph{FCED} network where the fully-connected decoder maps to the deformation graph, which the embedded deformation layer (EDL) in turn maps to the full-resolution mesh. 

\subsection{Evaluation Settings}\label{ssec:evaluation_settings}

\begin{table}
\parbox{.48\linewidth}{
\centering
\scalebox{0.75}{
\begin{tabular}{l | l  l | l  l | l  l | l  l |}
 & \multicolumn{2}{|c|}{DFaust} & \multicolumn{2}{|c|}{SynHand5M} & \multicolumn{2}{|c|}{Cloth } & \multicolumn{2}{|c|}{CoMA }\\
\cline{2-9}
                                   & 8   & 32  & 8  & 32  & 8  & 32  & 8  & 32 \\ \hline
\multicolumn{1}{|l|}{w/ GL} & 8.92 & 2.75 & 9.02 & 2.95 & \textbf{11.26} & 6.45 & 1.38 & 0.99  \\ %
\multicolumn{1}{|l|}{w/ LP} & 7.71 & \textbf{2.22} & \textbf{8.00} & 2.52 & 11.46 & 7.96 & 1.25 & \textbf{0.79}  \\ %
\multicolumn{1}{|l|}{Ours}       & \textbf{6.69} & 2.23 & 8.12 & \textbf{2.51} & 11.28 & \textbf{6.40} & \textbf{1.23} & 0.81  \\ \hline
\end{tabular}}
\caption{Average per-vertex errors on the test sets of DFaust ($cm$), SynHand5M ($mm$), textureless cloth ($mm$) and CoMA ($mm$) for 8 and 32 latent dimensions.}
\label{tab:edlexp}
}
\hfill
\parbox{.48\linewidth}{
\centering
\scalebox{0.75}{
\begin{tabular}{l | l  l | l  l | l  l | l  l |}
 & \multicolumn{2}{|c|}{DFaust} & \multicolumn{2}{|c|}{SynHand5M} & \multicolumn{2}{|c|}{Cloth} & \multicolumn{2}{|c|}{CoMA}\\
\cline{2-9}
                                   & 8   & 32  & 8  & 32  & 8  & 32  & 8  & 32 \\ \hline
\multicolumn{1}{|l|}{N. 3DMM}     & 7.09 & \textbf{1.99} & 8.50 & 2.58 & 12.64 & 6.49 & 1.34 & \textbf{0.71} \\ %
\multicolumn{1}{|l|}{Ours}                     & \textbf{6.69} & 2.23 & \textbf{8.12} & \textbf{2.51} & \textbf{11.28} & \textbf{6.40} & \textbf{1.23} & 0.81 \\ \hline
\end{tabular}}
\caption{Average per-vertex errors on the test sets of DFaust (in $cm$), SynHand5M (in $mm$), textureless cloth (in $mm$) and CoMA (in $mm$) for 8 and 32 latent dimensions, compared with Neural 3DMM~\cite{Bouritsas2019}. %
}
\label{tab:quantSpiralPaper}
}
\end{table}

We first determine how to integrate the EDL into the training. %
Our proposed architecture regresses node positions and rotations and then uses the EDL to obtain the deformed mesh, on which the reconstruction loss is applied.

As an alternative, we consider the \emph{graph loss (GL)} with the $\ell_1$ reconstruction loss directly on the graph node positions (where the vertex positions of the input mesh that correspond to the graph nodes are used as ground-truth).
The GL setting uses the EDL only at test time to map to the full mesh, but not for training.
Although the trained network predicts graph node positions $\boldsymbol{t}_l$ at test time, it does not regress graph node rotations $\boldsymbol{R}_l$ which are necessary for the EDL.
We compute the missing rotation for each graph node $l$ as follows:
assuming that each node's neighborhood transforms roughly rigidly, 
we solve a small Procrustes problem that computes the rigid rotation between the 1-ring neighborhoods of $l$ in the template graph and in the regressed network output.
We directly use this rotation as $\boldsymbol{R}_l$.

We also consider the alternative of estimating the local Procrustes rotation \emph{inside the network during training (LP)}. 
We add a reconstruction loss on the deformed mesh as computed by the EDL.
Here, we do not back-propagate through the rotation computation to avoid training instabilities.

Table \ref{tab:edlexp} %
shows the quantitative results using the average per-vertex Euclidean error. 
Using the EDL during training leads to better quantitative results, as the network is aware of the skinning function and can move the graph nodes accordingly. 
In addition to being an order of magnitude faster than LP, regressing rotations either gives the best results or is close to the best.
We use the EDL with regressed rotation parameters during training in all further experiments.

We use spiral graph convolutions \cite{Bouritsas2019}, but show in the supplemental document that spectral graph convolutions \cite{Defferrard2016} also give similar results. %

\subsection{Evaluations of the Autoencoder}\label{ssec:evaluations_of_autoencoder} 
\begin{figure}
\begin{center}
\includegraphics*[width=0.9\linewidth]{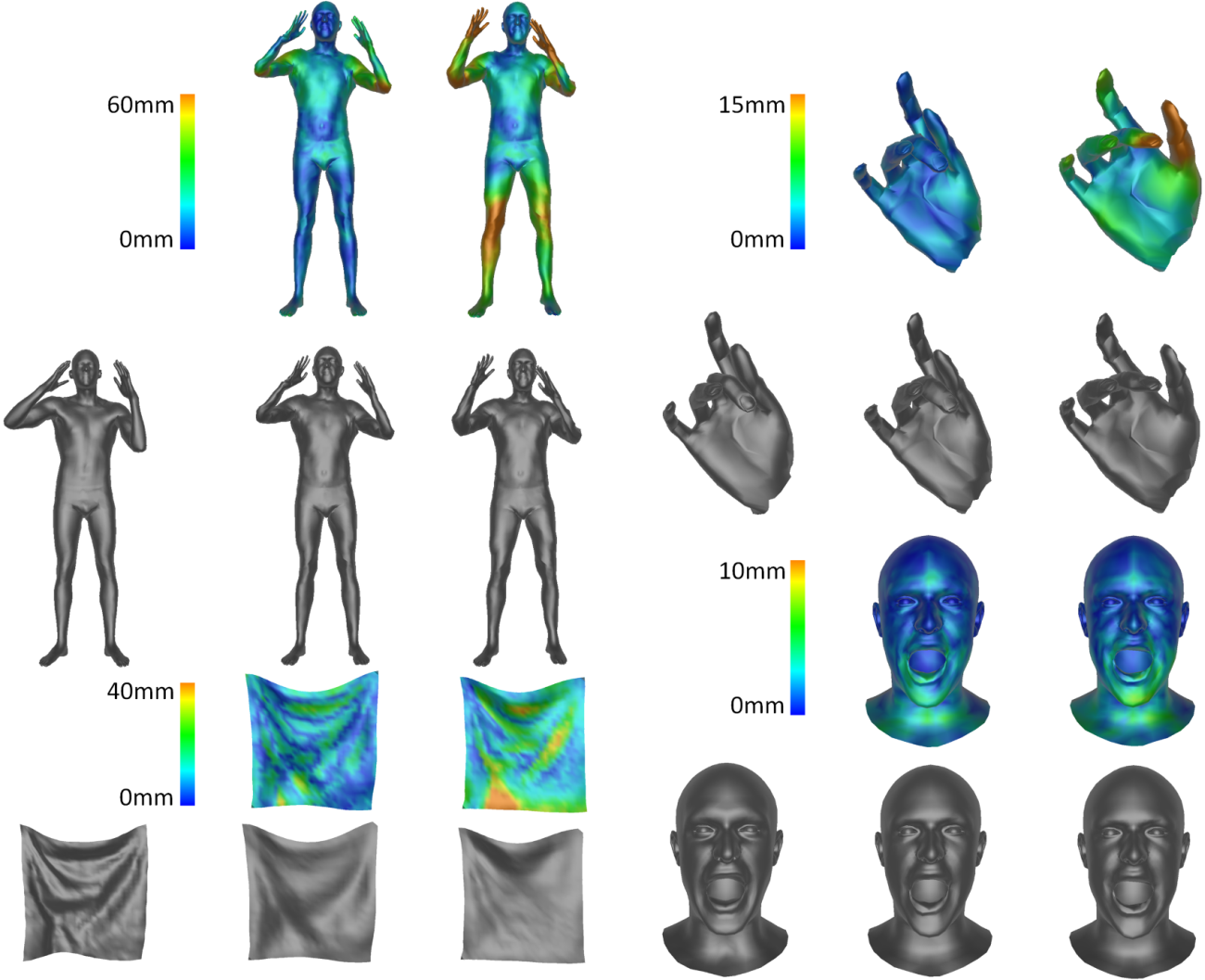}
\caption{Auto-encoding results on all four datasets. From left to right: ground-truth, ours with latent dimension 32, ours with latent dimension 8.}
\label{fig:meshautoencoding}
\end{center} 
\end{figure}

\textbf{Qualitative Evaluations.}
Our architecture significantly outperforms the baselines  qualitatively on the DFaust and SynHand5M datasets, as seen in Figs.~\ref{fig:artifacts} and~\ref{fig:meshautoencoding}. 
Convolutional architectures without an embedded graph produce strong artifacts in the hand, feet and face regions in the presence of large deformations. %
Since EDL explicitly models deformations, we preserve fine details of the template under strong non-linear deformations and  %
articulations of extremities. %
\\ %
\textbf{Quantitative Evaluations. }
We compare the proposed DEMEA to the baselines on the autoencoding task, see Table~\ref{tab:quantAuto}.
\begin{figure}
\begin{center}
\includegraphics*[width=0.7\linewidth]{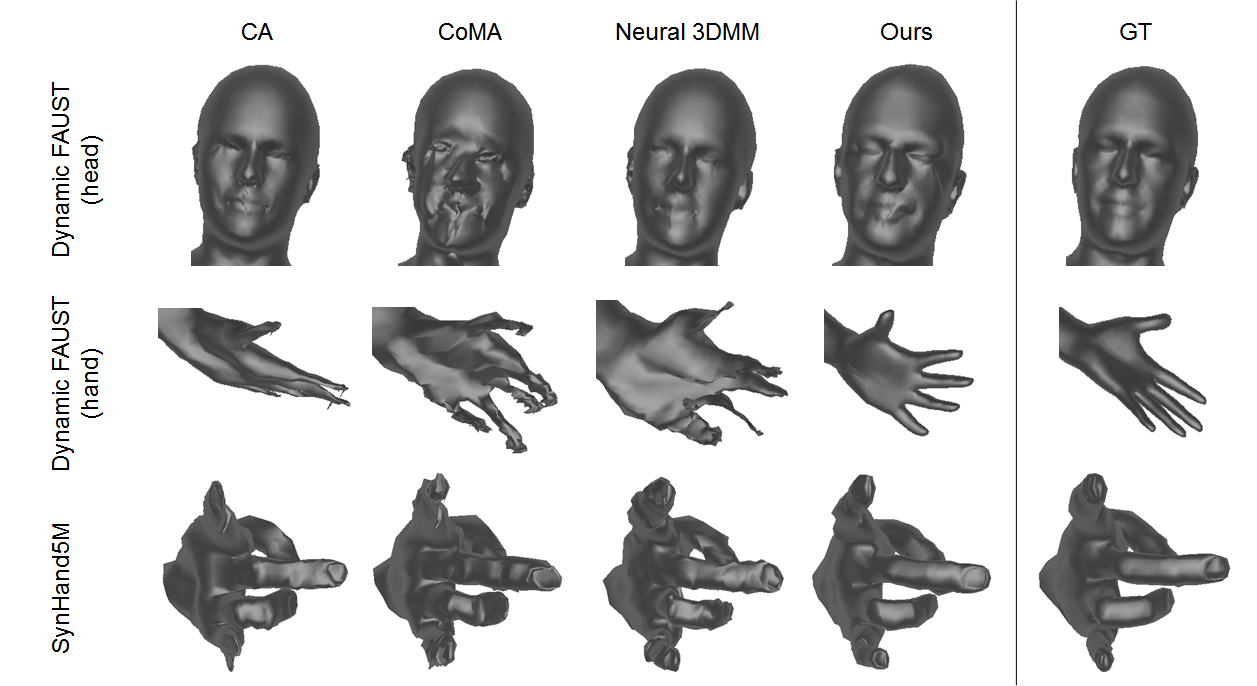}
\caption{In contrast to graph-convolutional networks that directly regress vertex positions, our embedded graph layer does not show artifacts. These results use a latent dimension of $32$. } 
\label{fig:artifacts}
\end{center} 
\end{figure}
While the fully-connected baselines are competitive for larger dimensions of the latent space, their memory demand increases drastically. 
On the other hand, they perform significantly worse for low dimensions on all datasets, except for DFaust. 
In this work, we are interested in low latent dimensions, \textit{e.g.} less than 32, as we want to learn mesh representations that are as compact as possible. %
We also observe that adding EDL to the fully-connected baselines maintains their performance.
Furthermore, the lower test errors of FCED on Cloth indicate that network capacity (and not EDL) limits the quantitative results.

On SynHand5M, Cloth and CoMA, the convolutional baselines perform on par with DEMEA. 
On DFaust, our performance is slightly worse, perhaps because other architectures can also fit to the high-frequency details and noise. 
EDL regularizes deformations to avoid artifacts, which also prevents fitting to high-frequency or small details. 
Thus, explicitly modelling deformations via the EDL and thereby avoiding artifacts has no negative impact on the quantitative performance. 
Since CoMA mainly contains small and local deformations, DEMEA does not lead to any quantitative improvement. 
This is more evident in the case of latent dimension 32, as the baselines can better reproduce noise and other high-frequency deformations.  

\noindent\textbf{Comparisons.} 
In  extensive comparisons with several competitive baselines, we have demonstrated the usefulness of our approach for autoencoding strong non-linear deformations and articulated motion. 
Next, we compare DEMEA to the existing state-of-the-art CoMA approach~\cite{Ranjan2018}. 
We train their architecture on all mentioned datasets with a latent dimension of $8$, which is also used in~\cite{Ranjan2018}. 
We outperform their method quantitatively on DFaust ($6.7cm$ \textit{vs.}~$8.4cm$), on SynHand5M ($8.12mm$ \textit{vs.}~$8.93mm$), on Cloth ($1.13cm$ \textit{vs.}~$1.44cm$), and even on CoMA ($1.23mm$ \textit{vs.}~$1.42mm$), where the deformations are not large. 
We also compare to Neural 3DMM \cite{Bouritsas2019} on latent dimensions $8$ and $32$, similarly to \cite{Ranjan2018} on their proposed hierarchy.
See Table~\ref{tab:quantSpiralPaper} for the results.
DEMEA performs better than Neural 3DMM in almost all cases. 
In Fig.~\ref{fig:artifacts}, we show that DEMEA avoids many of the artifacts present in the case of \cite{Ranjan2018}, \cite{Bouritsas2019} and other baselines.

\section{Applications}\label{sec:applications} 

\subsection{RGB to Mesh} 
\begin{figure*}
\begin{subfigure}{0.53\textwidth}
\begin{center}
\includegraphics*[width=\linewidth]{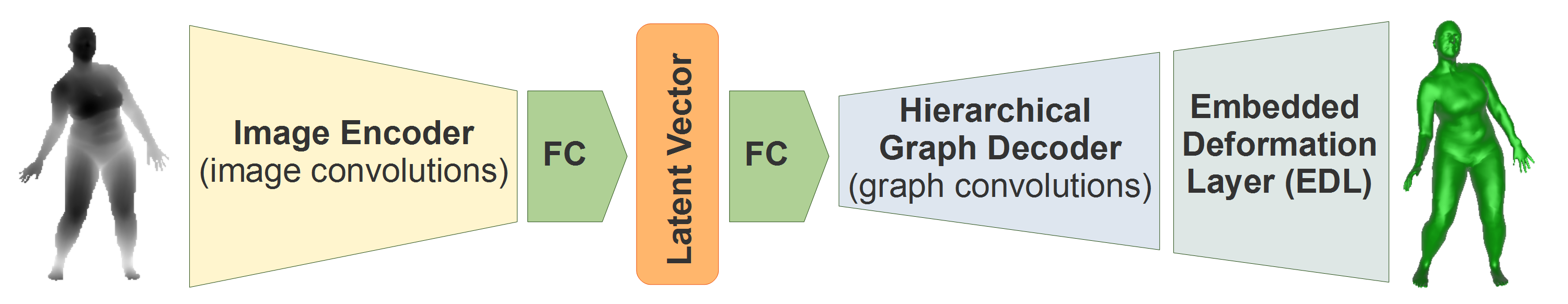}
\caption{}
\label{fig:pipeline2}
\end{center} 
\end{subfigure}
\hfill
\begin{subfigure}{0.45\textwidth}
\begin{center}
\includegraphics*[width=\linewidth]{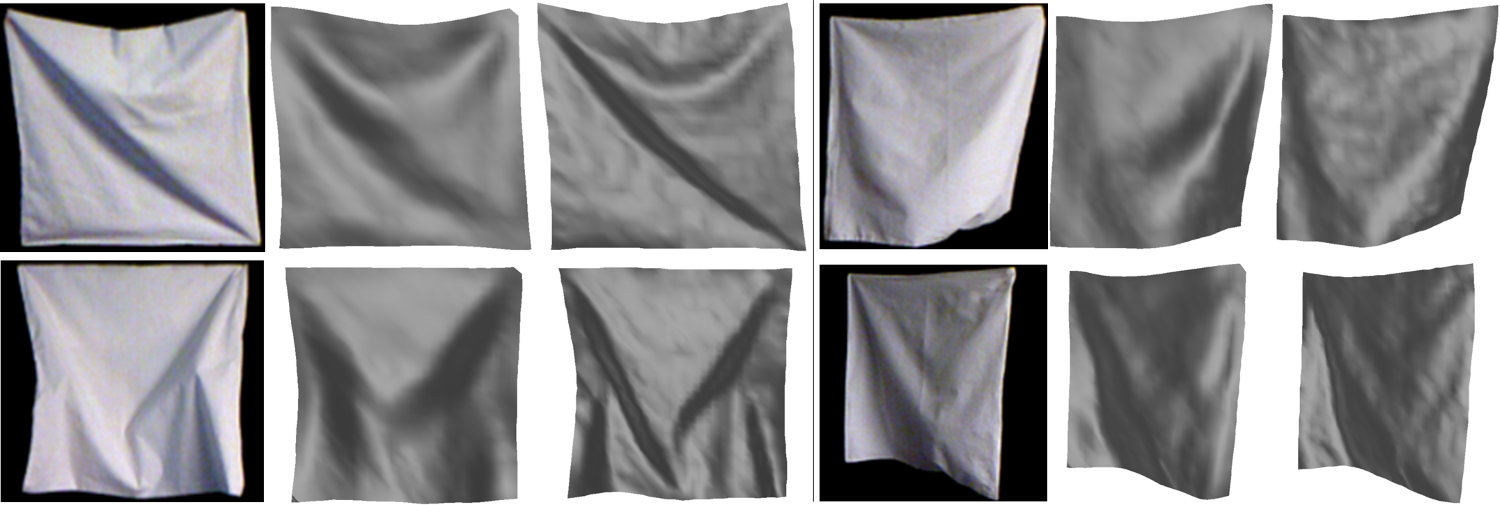}
\caption{}
\label{fig:rgbcloth}
\end{center} 
\end{subfigure}
\caption{(left) Image/depth-to-mesh pipeline: To train an image/depth-to-mesh reconstruction network, we employ a convolutional image encoder and initialize the decoder to a pre-trained graph decoder. (right) RGB-to-mesh results on our test set. From left to right: real RGB image, our reconstruction, ground-truth.}
\end{figure*}

On the Cloth~\cite{Bednarik2018} dataset, we show that DEMEA can reconstruct meshes from RGB images. 
See Fig.~\ref{fig:rgbcloth} for qualitative examples with a latent dimension of $32$. 

On our test set, our proposed architecture achieves RGB-to-mesh reconstruction errors of $16.1mm$ and $14.5mm$ for latent dimensions $8$ and $32$, respectively.
Bedna\v{r}\'{i}k \textit{et al.} \cite{Bednarik2018}, who use a different split than us, report an error of $21.48mm$.
The authors of IsMo-GAN \cite{shimada2019ismo} report results on their own split for IsMo-GAN and the Hybrid Deformation Model Network (HDM-net) \cite{Golyanik2018}. %
On their split, HDM-Net achieves an error of $17.65mm$ after training for 100 epochs using a batch size of 4. 
IsMo-GAN obtains an error of $15.79mm$. 
Under the same settings as HDM-Net, we re-train our approach without pre-training the mesh decoder. 
Our approach achieves test errors of $16.6mm$ and $13.8mm$ using latent dimensions of $8$ and $32$, respectively.

\subsection{Depth to Mesh}
\begin{figure*}
\begin{subfigure}{0.48\textwidth}
\begin{center}
\includegraphics*[width=\linewidth]{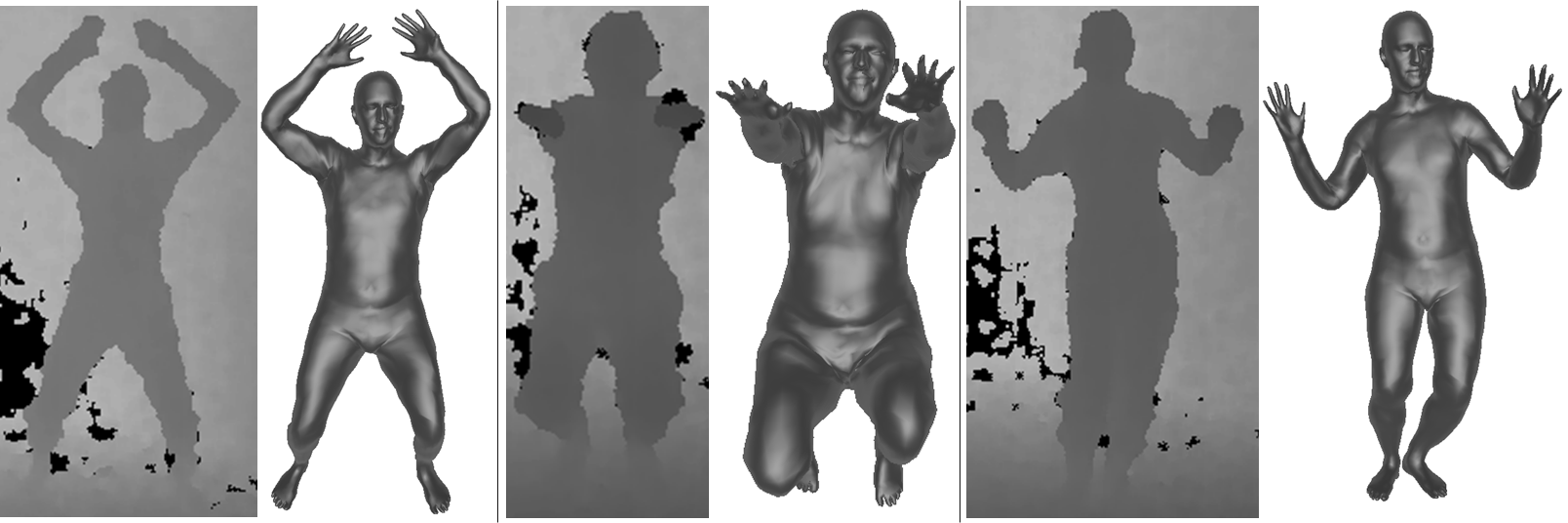}
\caption{DEMEA (right) on real Kinect depth images (left) with latent dimension 32.}
\label{fig:bodiesd2m}
\end{center} 
\end{subfigure}
\begin{subfigure}{0.48\textwidth}
\begin{center}
\includegraphics*[width=\linewidth]{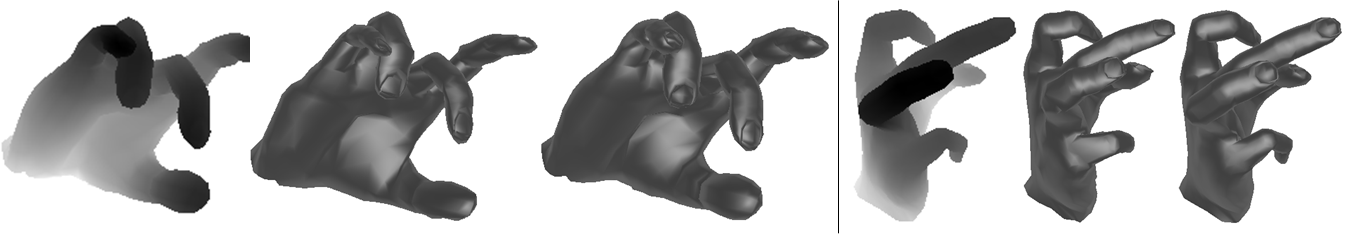}
\caption{Reconstruction results from synthetic depth images of hands using a latent dimension of $32$. From left to right: depth, our reconstruction, ground-truth.}
\label{fig:handsDepth}
\end{center} 
\end{subfigure}
\caption{Reconstruction from a single depth image.}
\end{figure*}

\textbf{Bodies.}
We train a network with a latent space dimension of $32$. %
Quantitatively, we obtain an error of $2.3cm$ %
on un-augmented synthetic data. 
Besides, we also apply our approach to real data, see Fig.~\ref{fig:bodiesd2m}. 
To this end, we found it necessary to augment the depth images with artificial noise to lessen the domain gap. 
Video results are included in the supplementary. %
\\ 
\textbf{Hands. }
DEMEA can reconstruct hands from depth as well, see Fig.~\ref{fig:handsDepth}. 
We achieve a reconstruction error of $6.73mm$ for a latent dimension of $32$. 
Malik~\etal~\cite{Malik2018} report an error of $11.8$ $mm$.
Our test set is composed of a random sample of fully randomly generated hands from the dataset, which is very challenging. 
We use $256\times256$, whereas \cite{Malik2018} use images of size $96\times96$.

\subsection{Latent Space Arithmetic} 
\begin{figure*}
\begin{subfigure}{0.48\textwidth}
\begin{center}
\includegraphics*[width=\linewidth]{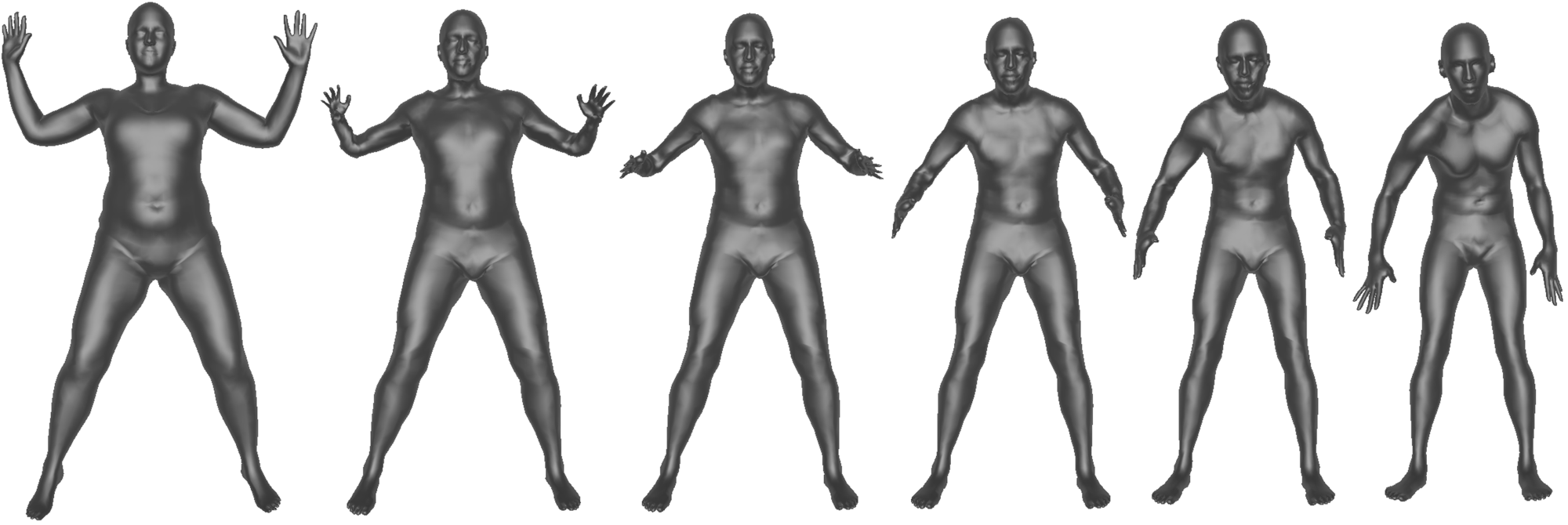}
\caption{Interpolation results, from left to right: source mesh, $\alpha=0.2$, $\alpha=0.4$, $\alpha=0.6$, $\alpha=0.8$, target mesh.}
\label{fig:interpolation}
\end{center} 
\end{subfigure}
\begin{subfigure}{0.48\textwidth}
\begin{center}
 \includegraphics*[width=\linewidth]{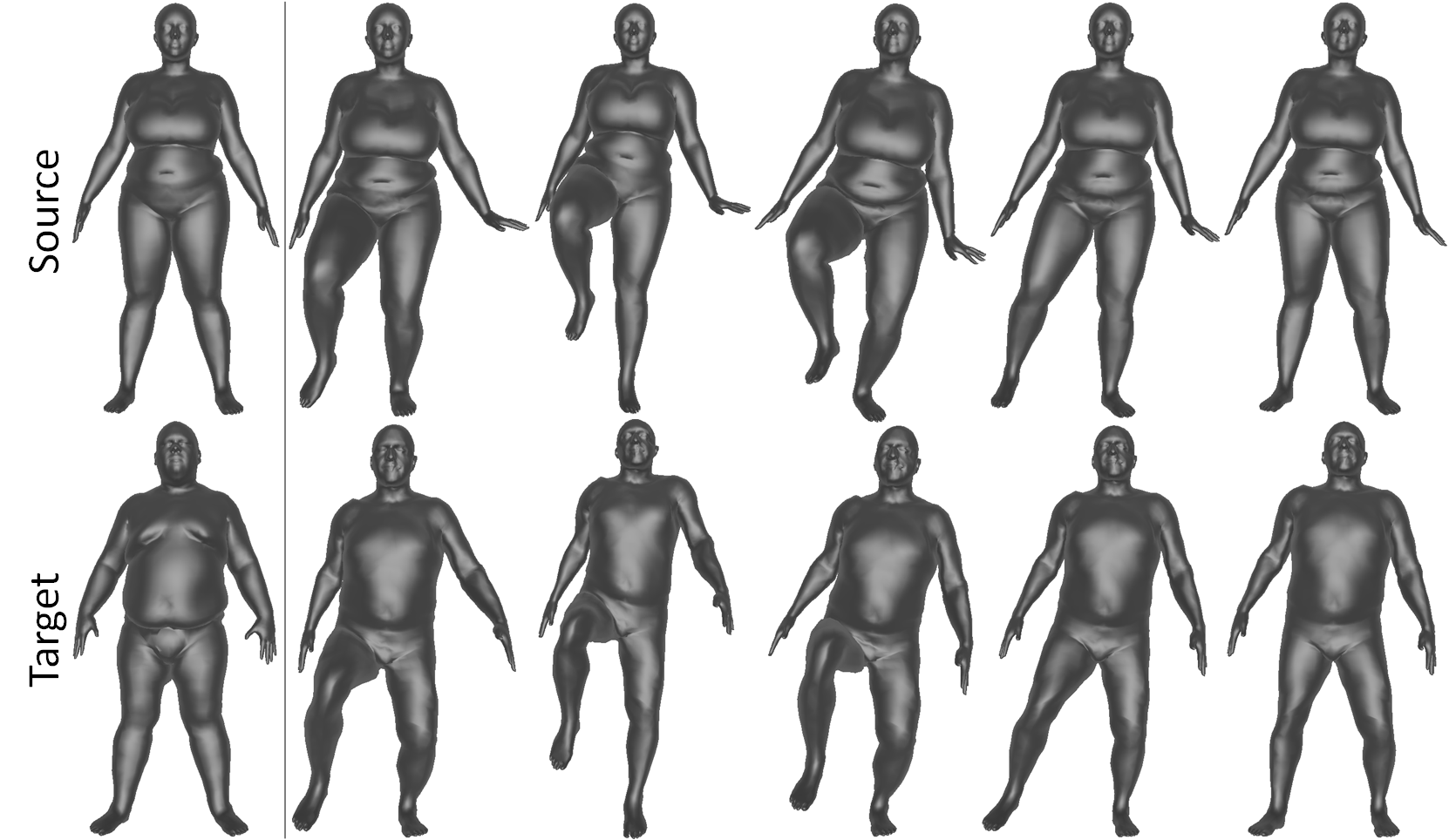}
 \caption{Deformation transfer from a source sequence to a target identity. The first column shows $\textbf{M}_0$ and $\textbf{M}'_0$.}
 \label{fig:transfer}
 \end{center} 
\end{subfigure}
\caption{Latent space arithmetic.}
\end{figure*}

Although we do not employ any regularization on the latent space, we found empirically that the network learns a well-behaved latent space.
As we show in the supplemental document and video, this allows DEMEA to temporally smooth tracked meshes from a depth stream.
\\ 
\textbf{Latent Interpolation.}
We can linearly interpolate the latent vectors $\mathcal{S}$ and $\mathcal{T}$ of a source and a target mesh: $\mathcal{I}(\alpha) = (1-\alpha)\mathcal{S} + \alpha\mathcal{T}$.
Even for highly different poses and identities, these $\mathcal{I}(\alpha)$ yield plausible in-between meshes, see Fig.~\ref{fig:interpolation}.

\noindent\textbf{Deformation Transfer.}
 The learned latent space allows to transfer poses between different identities on DFaust. %
 Let a sequence of source meshes $\mathbf{S} = \{\textbf{M}_i\}_{i}$ of person $A$ and a target mesh $\textbf{M}'_0$ of person $B$ be given, where w.l.o.g. $\textbf{M}_0$ and $\textbf{M}'_0$ correspond to the same pose.
 We now seek a sequence of target meshes $\mathbf{S}'= \{\textbf{M}'_i\}_{i}$ of person $B$ performing the same poses as person $A$ in $\mathbf{S}$.
 We encode $\mathbf{S}$ and $M'_0$ into the latent space of the mesh auto-encoder, yielding the corresponding latent vectors $\{\mathcal{M}_i\}_{i}$ and $\mathcal{M}'_0$.
 We define the identity difference $d = \mathcal{M}'_0 - \mathcal{M}_0$ and set $\mathcal{M}'_i = \mathcal{M}_i + d$ for $i > 0$.
 Decoding $\{\mathcal{M}'_i\}_i$ using the mesh decoder than yields $\mathbf{S'}$.
 See Fig.~\ref{fig:transfer} and the supplement for qualitative results.

\section{Limitations}
While the embedded deformation graph excels on highly articulated, non-rigid motions, it has difficulties accounting for very subtle actions. 
Since the faces in the CoMA~\cite{Ranjan2018} dataset do not undergo large deformations, our EDL-based architecture does not offer a significant advantage.
Similar to all other 3D deep learning techniques, our approach also requires reasonably sized mesh datasets for supervised training, which might be difficult to capture or model.
We train our network in an object-specific manner.
Generalizing our approach across different object categories is an interesting direction for future work. 
\section{Conclusion} 
We proposed DEMEA --- the first deep mesh autoencoder for highly deformable and articulated scenes, such as human bodies, hands, and deformable surfaces, that builds on a new differentiable embedded deformation layer. 
The deformation layer reasons about local rigidity of the mesh and allows us to achieve higher quality autoencoding results compared to several baselines and existing approaches.
We have shown multiple applications of our architecture including non-rigid reconstruction from real depth maps and 3D reconstruction of textureless surfaces from images. 

\textbf{Acknowledgments.}
This work was supported by  
the ERC Consolidator Grant 4DReply (770784), 
the Max Planck Center for Visual Computing and Communications (MPC-VCC), and 
an Oculus research grant.

{\small
\bibliographystyle{splncs04}
\bibliography{main}
}

\clearpage
\title{Supplementary Material}
\titlerunning{DEMEA: Supplementary Material}
\author{}
\institute{}
\maketitle

\setcounter{section}{0}
\renewcommand\thesection{S.\arabic{section}}

\setcounter{figure}{0}
\renewcommand{\thefigure}{S.\arabic{figure}}

\setcounter{table}{0}
\renewcommand{\thetable}{S.\arabic{table}}

In this supplementary material, we expand on several points from the main paper.
In Sec.~\ref{sec:tracking}, we describe how we apply temporal smoothing in latent space.
Sec.~\ref{sec:skinning} contains details about skinning template meshes to embedded graphs.
Sec.~\ref{sec:artifacts} shows more examples of the artifacts seen in purely convolutional architectures.
Sec.~\ref{sec:architecture} provides low-level details of our architecture.
Sec.~\ref{sec:losses} gives the mathematical description of the losses we train with.
In Sec.~\ref{sec:normalization}, we describe how we normalize depth maps and meshes (for reconstruction from real depth data). 
An expanded version of Table 2 from the main manuscript is in Sec.~\ref{sec:std}.
In Sec.~\ref{sec:hierarchy}, we describe how to generate embedded graphs.
In Sec.~\ref{sec:conv}, we give further details and experiments on the graph convolutions. 
Sec.~\ref{sec:fca_ca} contains more results of FCA and CA.
We show artifacts due to coarse embedded graphs in Sec.~\ref{sec:coarse}.

\section{Depth-to-Mesh Tracking}\label{sec:tracking}

We can apply temporal smoothing to the reconstruction of a sequence of real depth images $\{\textbf{D}_i\}_{i}$, by decoding a running (causal) exponential average of the latent vectors of this sequence.
First, we encode the sequence into latent vectors $\{\mathcal{D}_i\}_{i}$.
We then define a smoothed sequence of latent vectors $\{\mathcal{D}'_i\}_{i}$ as follows: let $\mathcal{D}'_0 = \mathcal{D}_0$ and set $\mathcal{D}'_i = \alpha\cdot \mathcal{D}_{i} + (1-\alpha) \cdot \mathcal{D}'_{i-1}$ for $i>0$ for some $\alpha \in [0,1]$.
The smoothed sequence of meshes $\{\textbf{M}_i\}_{i}$ is obtained by decoding $\{\mathcal{D}'_i\}_{i}$. 
See the supplemental video for examples.

\section{Skinning}\label{sec:skinning}

We compute the linear blending weight $w_{l}(\bfp) $ as \cite{Sumner2007}:
\begin{equation}
    w_{l}(\bfp) = \exp\left(\frac{-\lVert \boldsymbol{g}_l - \bfp \rVert^2}{2 \cdot \sigma ^2}\right), 
\end{equation}
where we determine $\sigma \in \mathbb{R}$ heuristically as follows:
\begin{equation}
    \sigma = \sigma_0 \cdot d_{max} \cdot \frac{1}{\sqrt{L}}, 
\end{equation}
where $\sigma_0 = \frac{2}{3}$ and $d_{max}$ is the maximum Euclidean distance between any two mesh vertices, a proxy for the absolute scale of the mesh.%

Note that $\mathcal{N}_\bfp$, ${w_{l}}$ and $\sigma$ are pre-computed on the template mesh and graph and are kept fixed within each dataset.
We use $\lvert \mathcal{N}_{\bfp} \rvert = 6$ for embedded graphs on the first level and $\lvert \mathcal{N}_{\bfp} \rvert = 12$ for embedded graphs on the second level.

\section{Artifacts}\label{sec:artifacts}

Tables~\ref{fig:artifacts} and \ref{fig:artifacts2} show additional examples of artifacts that occur when not integrating the embedded graph into the network.
Even the most competitive network against which we compare (\textit{i.e.~}our ablation) suffers from visually unpleasing artifacts due to large non-rigid deformations, which most visibly occur on the hands and feet of DFaust.
However, due to the localized nature of the artifacts, they do not have a large impact on the quantitative errors.

\begin{table}
\centering
\begin{tabular}{c|c}
Convolutional Ablation & DEMEA\\\hline

\includegraphics[trim={500 800 450 0},clip,height=2cm]
{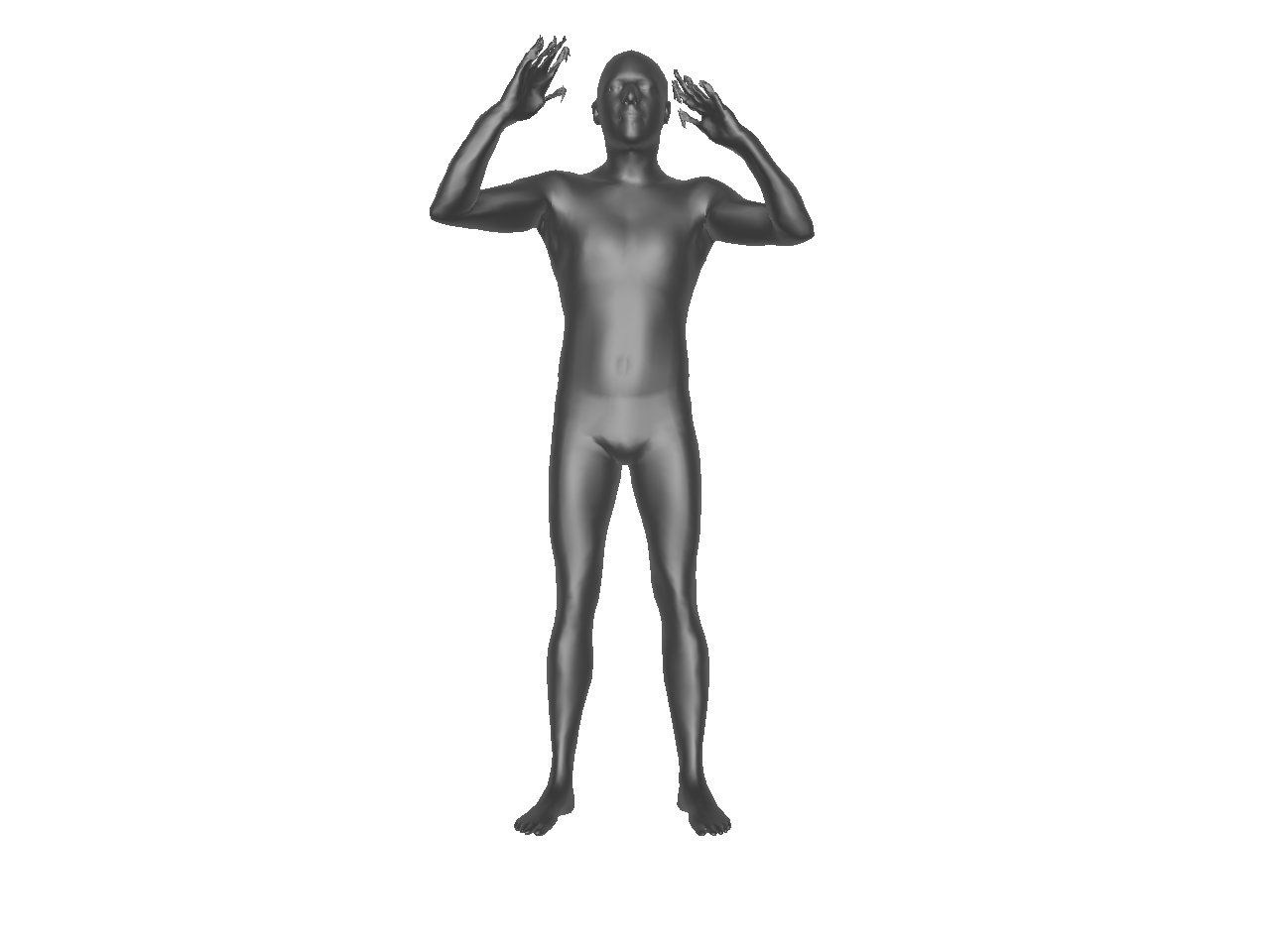} &
\includegraphics[trim={500 800 450 0},clip,height=2cm]
{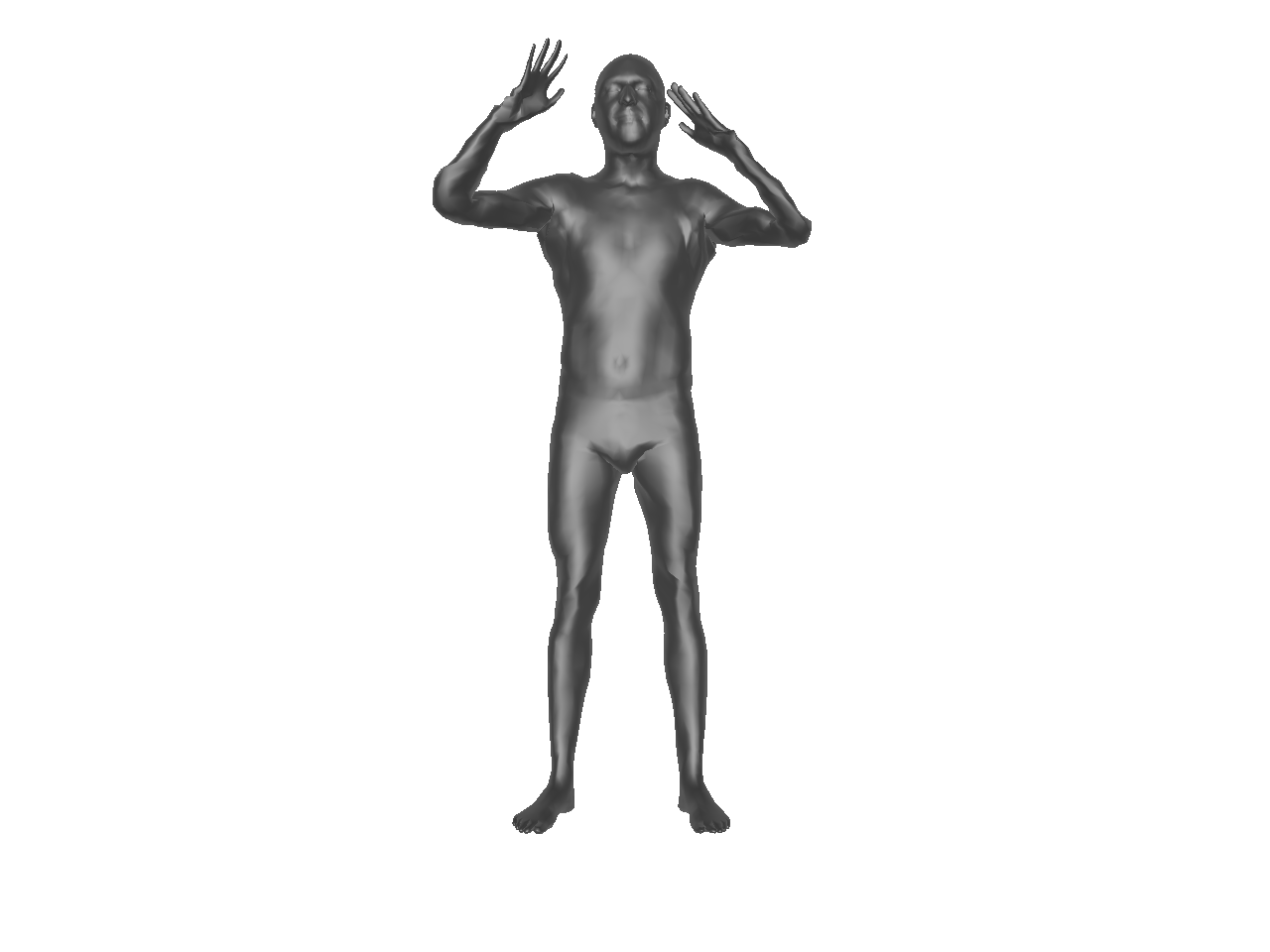}\\

\includegraphics[trim={360 450 760 350},clip,height=2cm]
{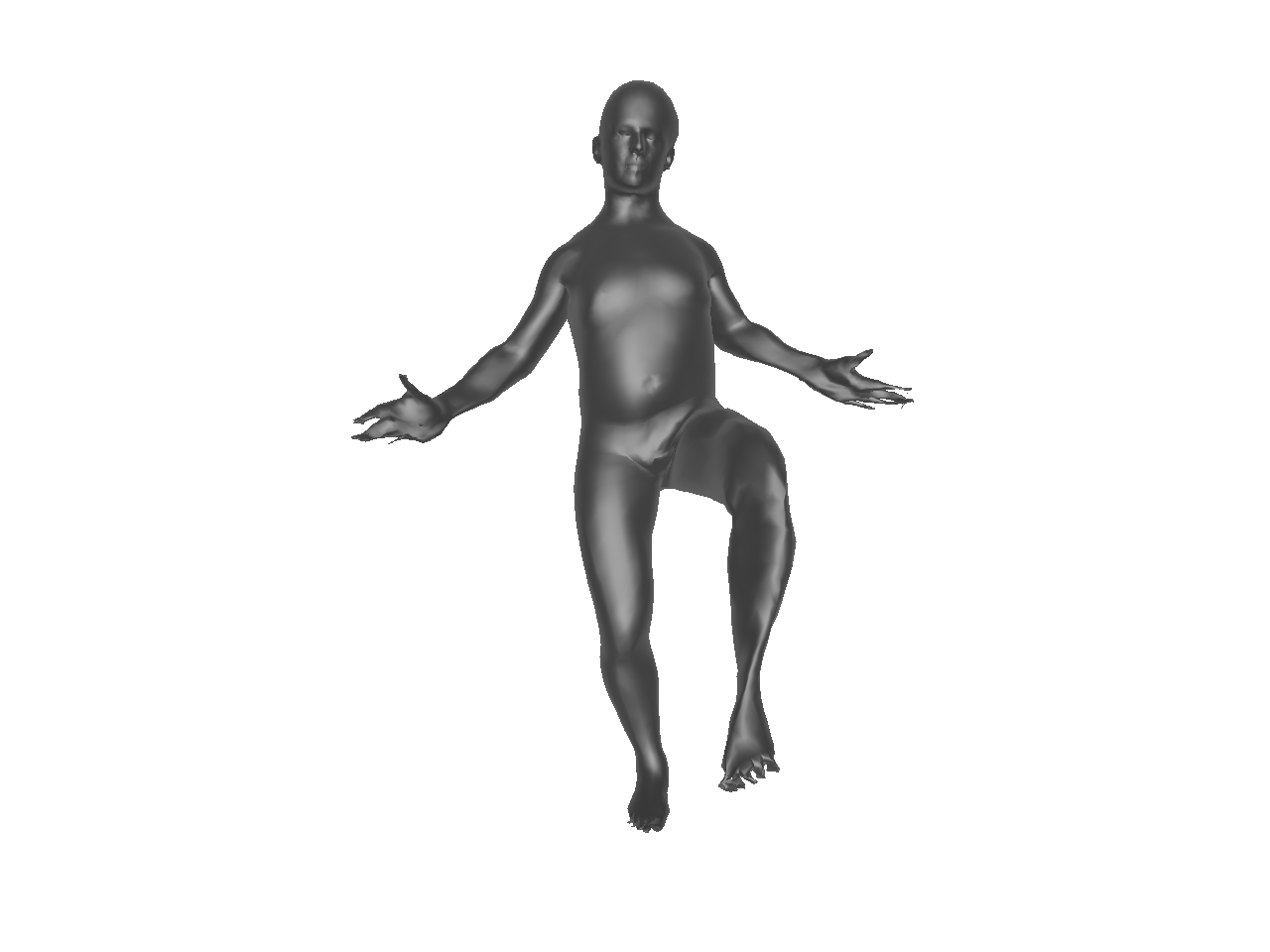} &
\includegraphics[trim={360 450 760 350},clip,height=2cm]
{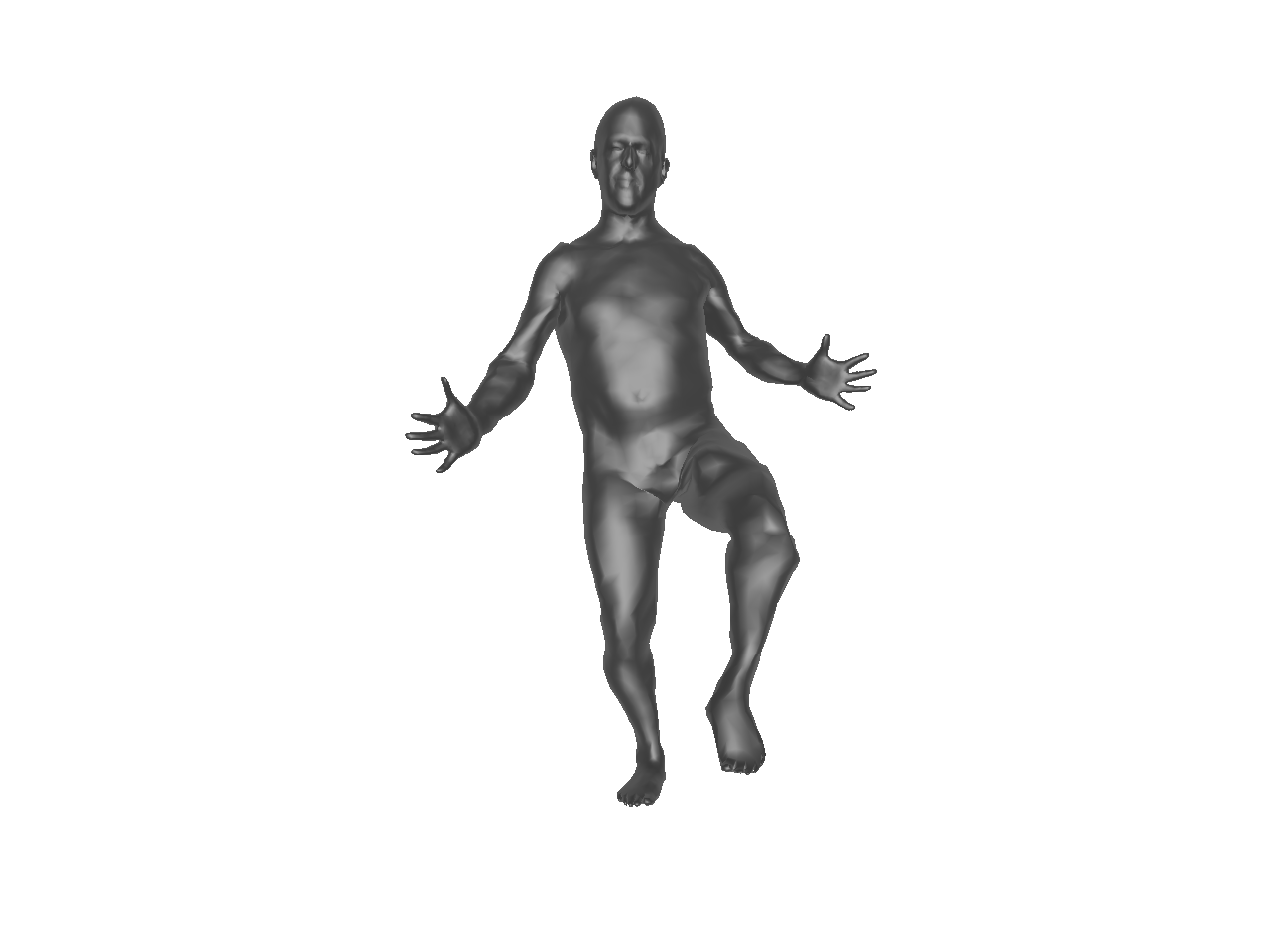}\\

\includegraphics[trim={700 120 500 700},clip,height=2cm]
{FIGURES/artifacts/CA_32/model_130.png}&
\includegraphics[trim={700 120 500 700},clip,height=2cm]
{FIGURES/artifacts/DEMEA_32/model_130.png}\\

\includegraphics[trim={550 80 300 700},clip,height=2cm]
{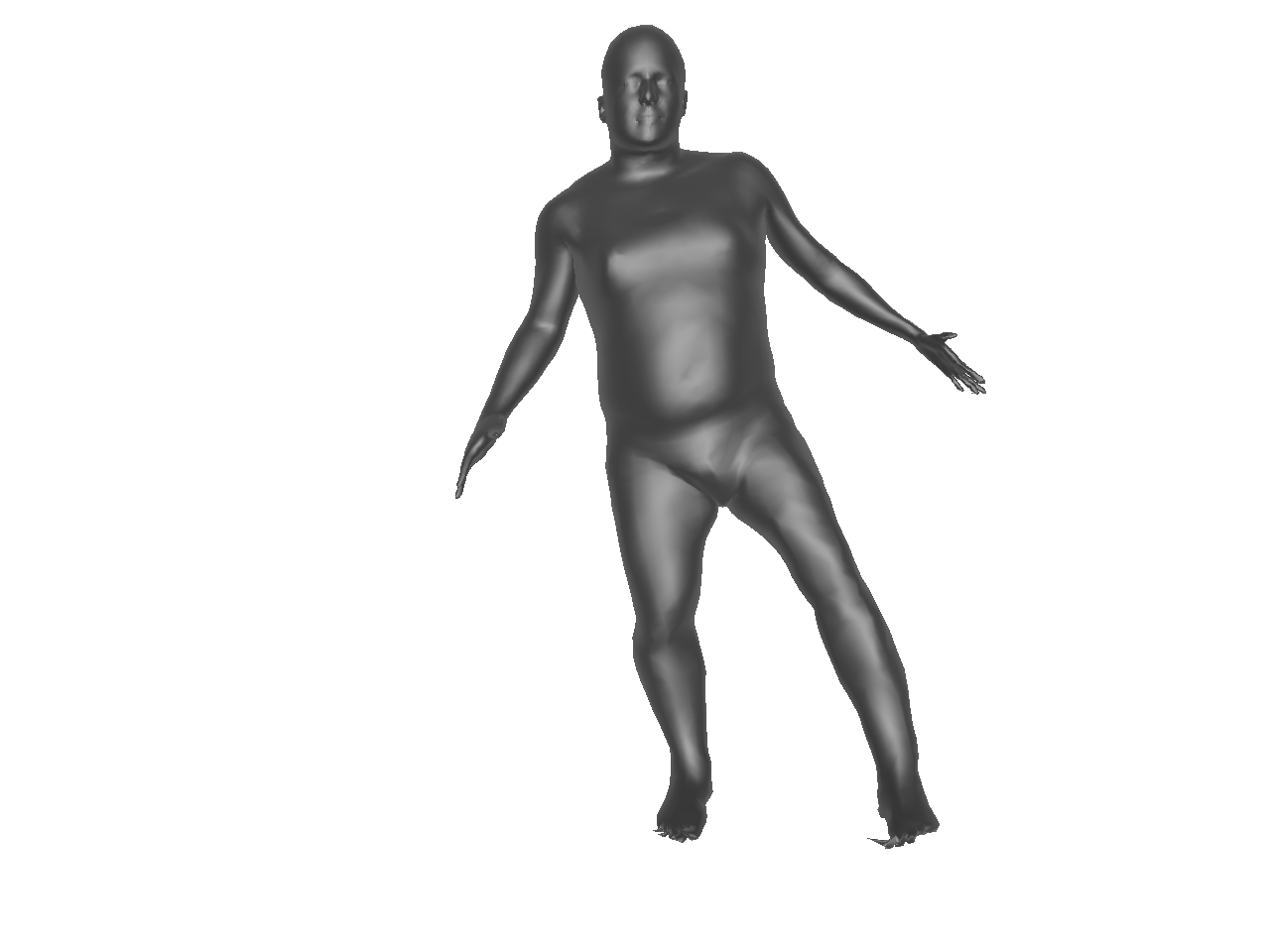}
&
\includegraphics[trim={550 80 300 700},clip,height=2cm]
{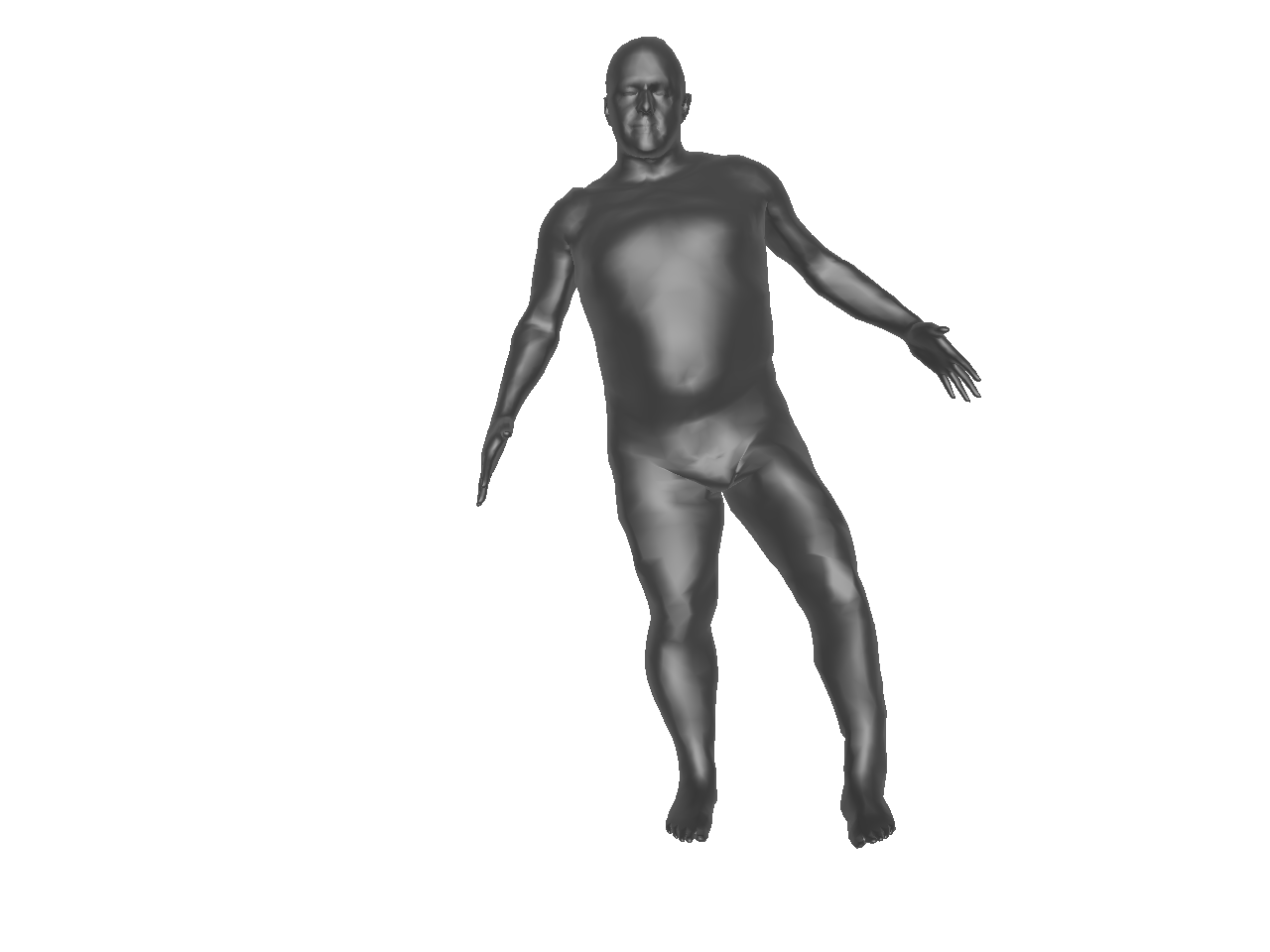}\\

\includegraphics[trim={200 600 900 200},clip,height=2cm]
{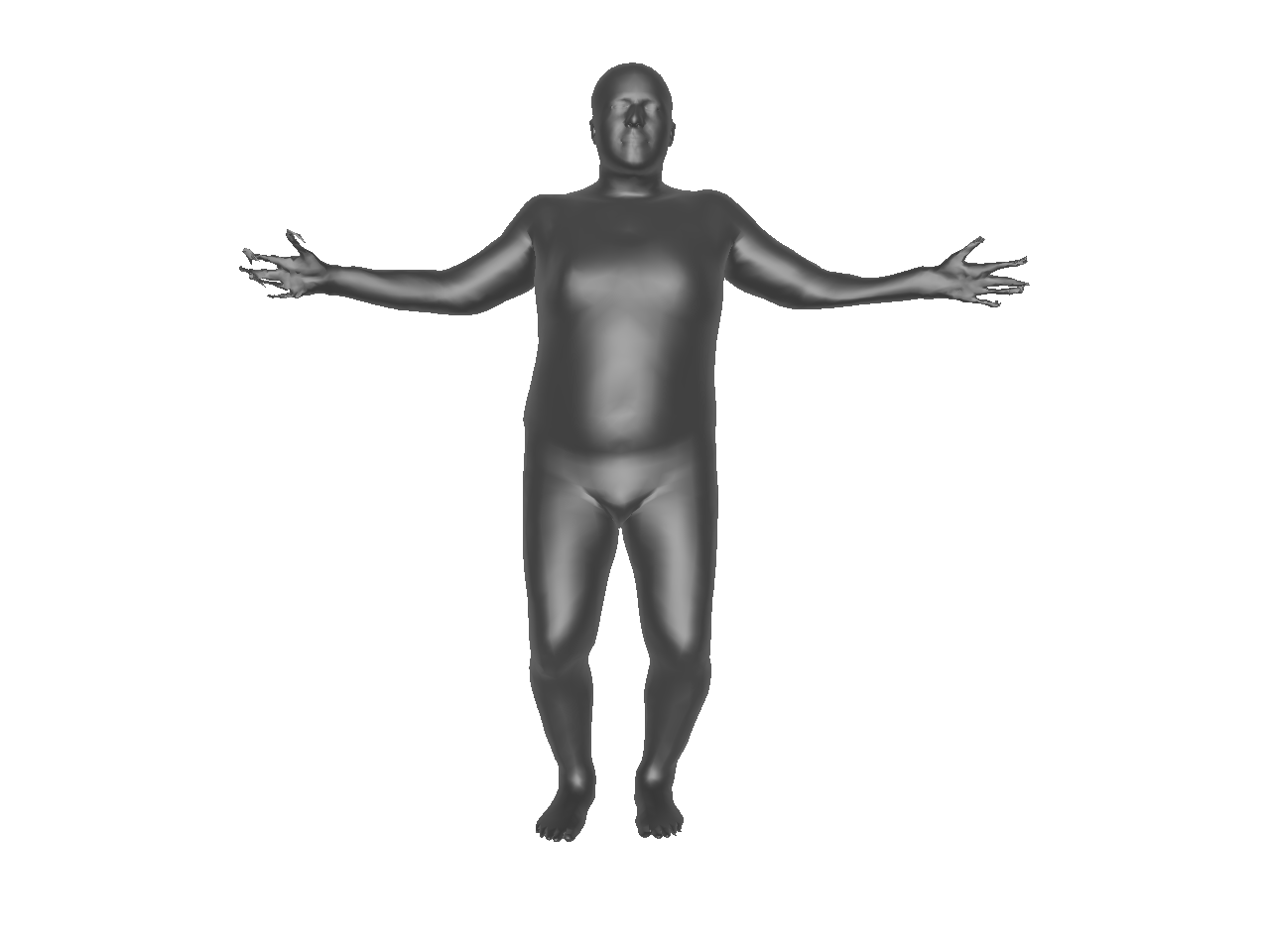}&
\includegraphics[trim={200 600 900 200},clip,height=2cm]
{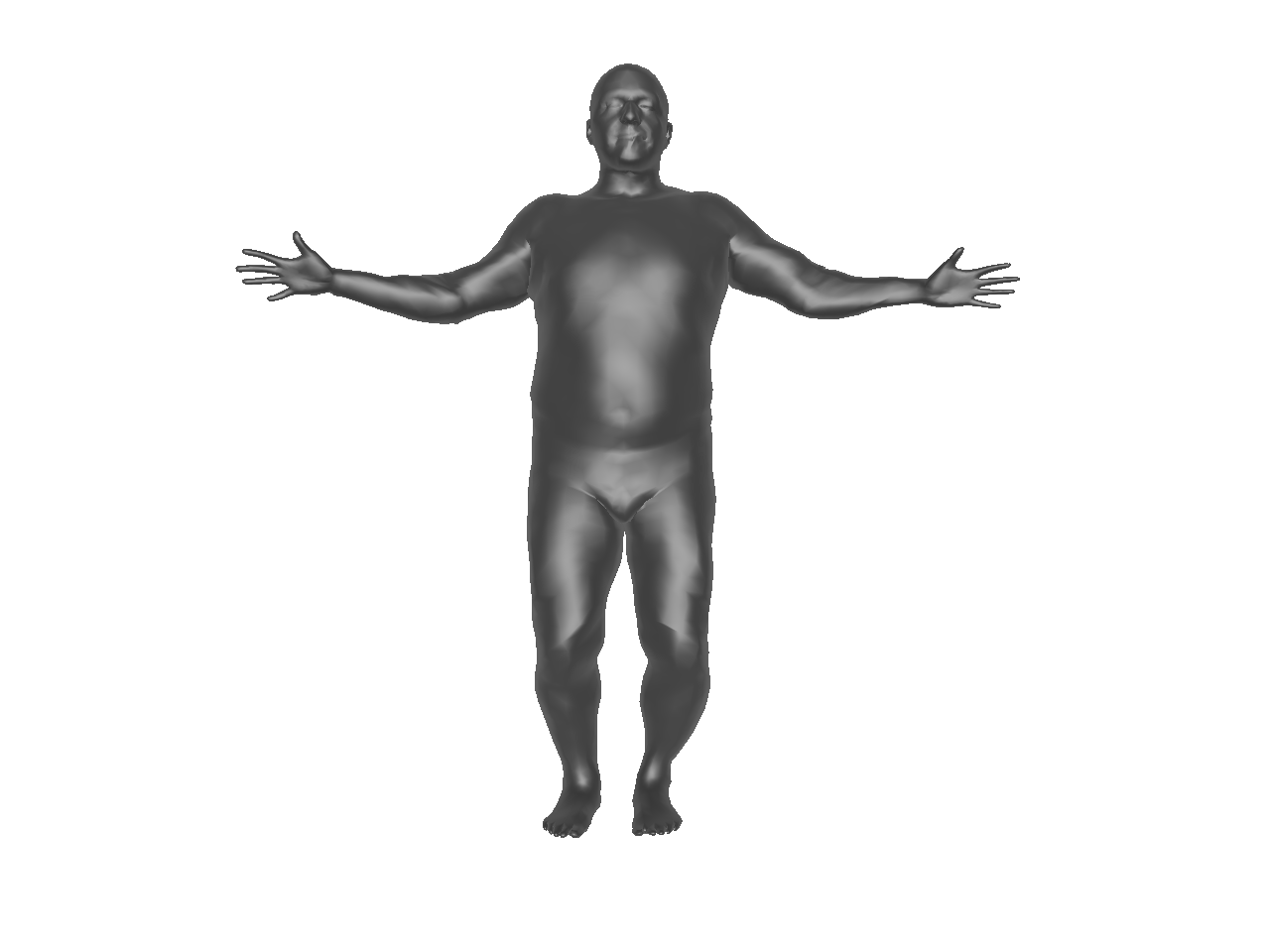}\\

\includegraphics[trim={530 50 600 720},clip,height=2cm]
{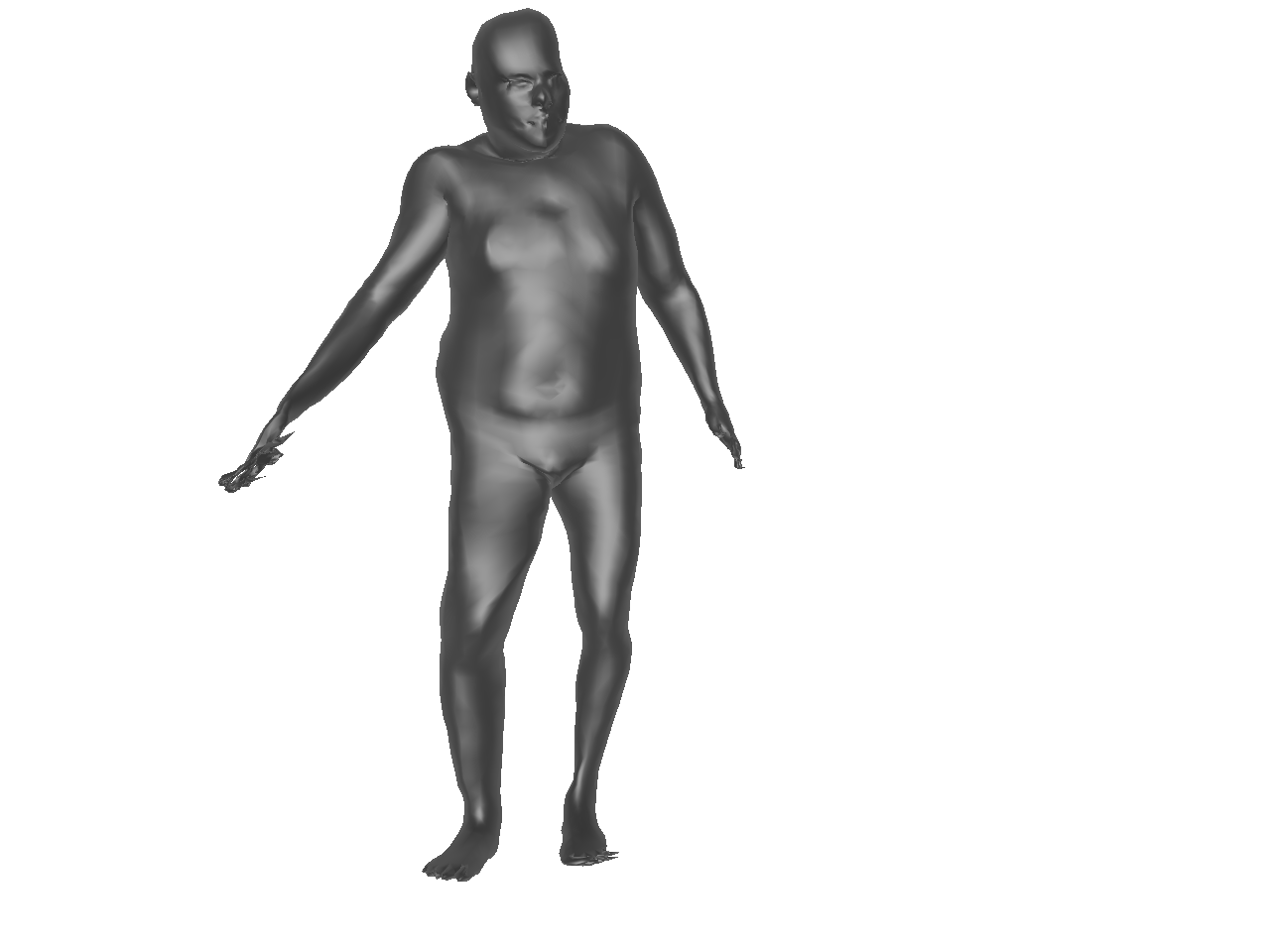}&
\includegraphics[trim={530 50 600 720},clip,height=2cm]
{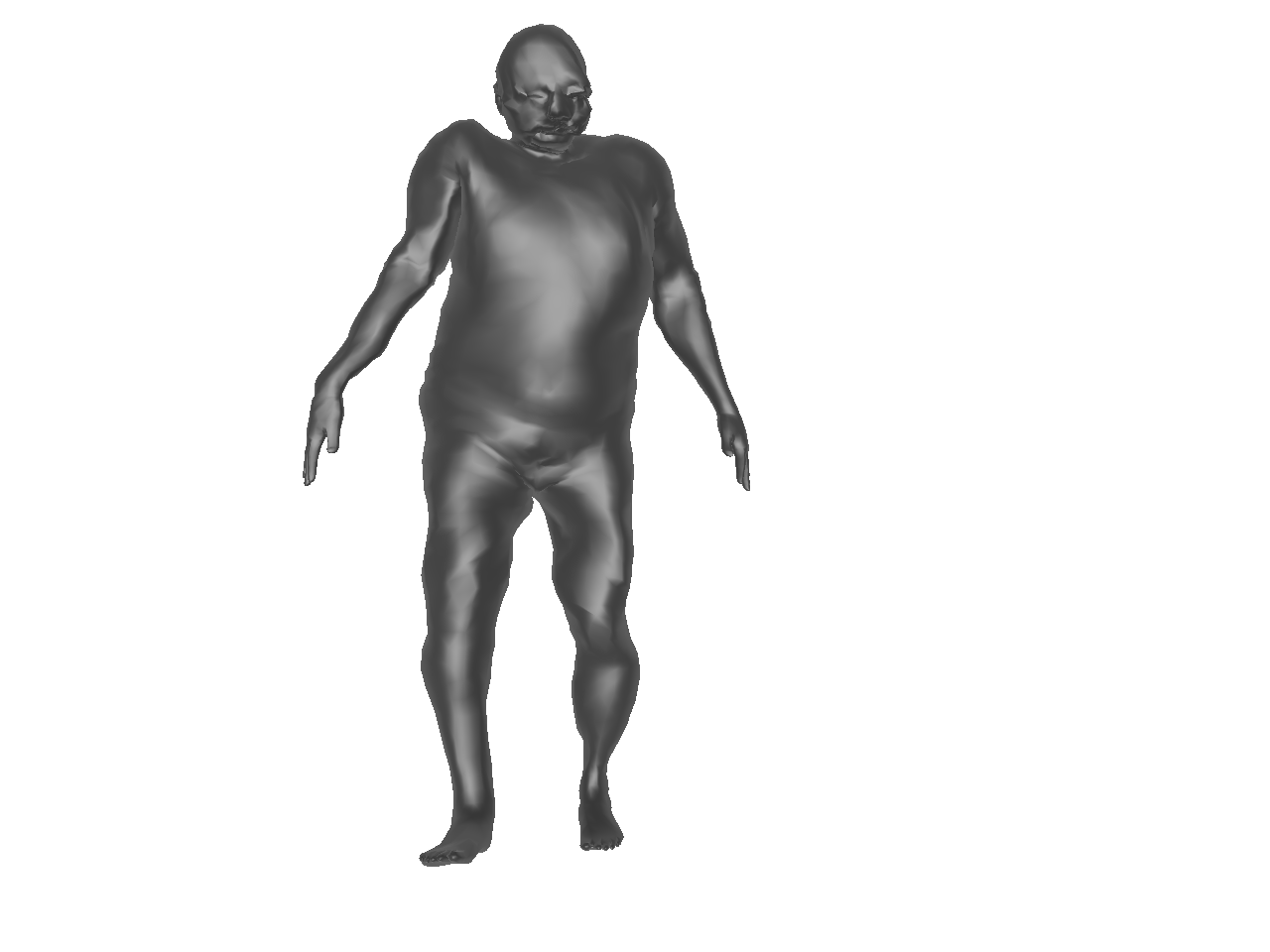}\\

\includegraphics[trim={370 800 750 0},clip,height=2cm]
{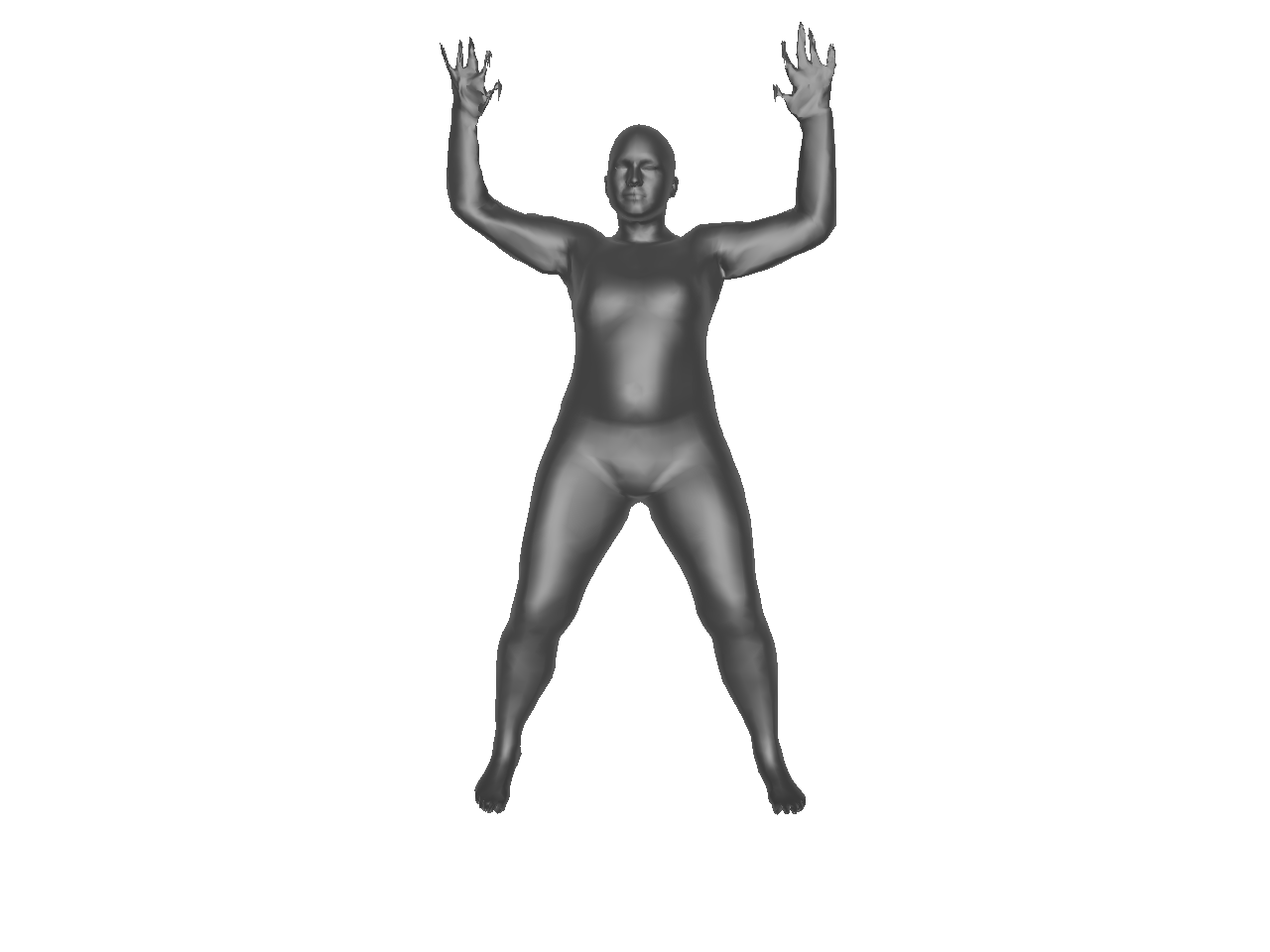}&
\includegraphics[trim={370 800 750 0},clip,height=2cm]
{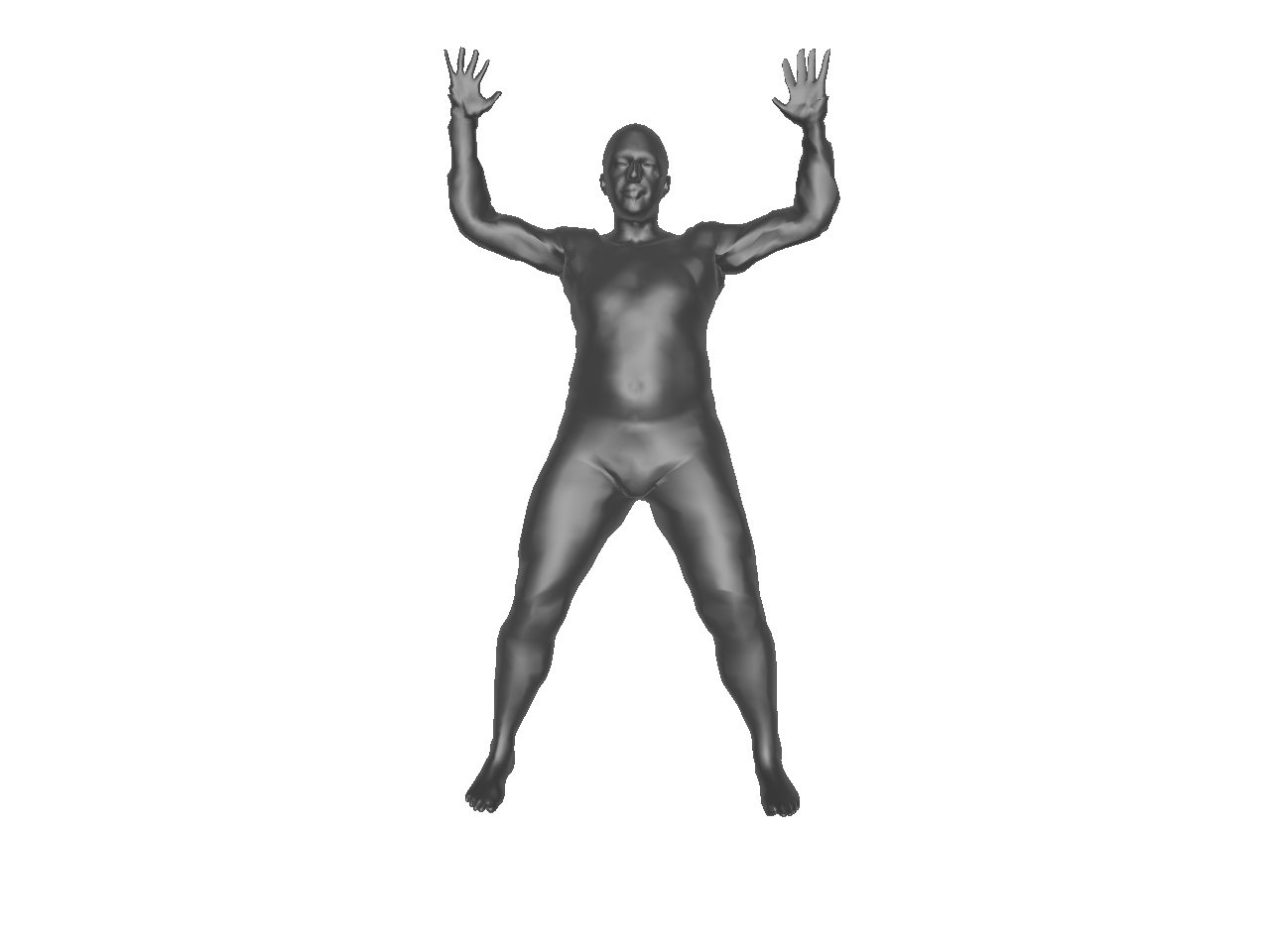}\\

\hline

\includegraphics[trim={500 800 450 0},clip,height=2cm]
{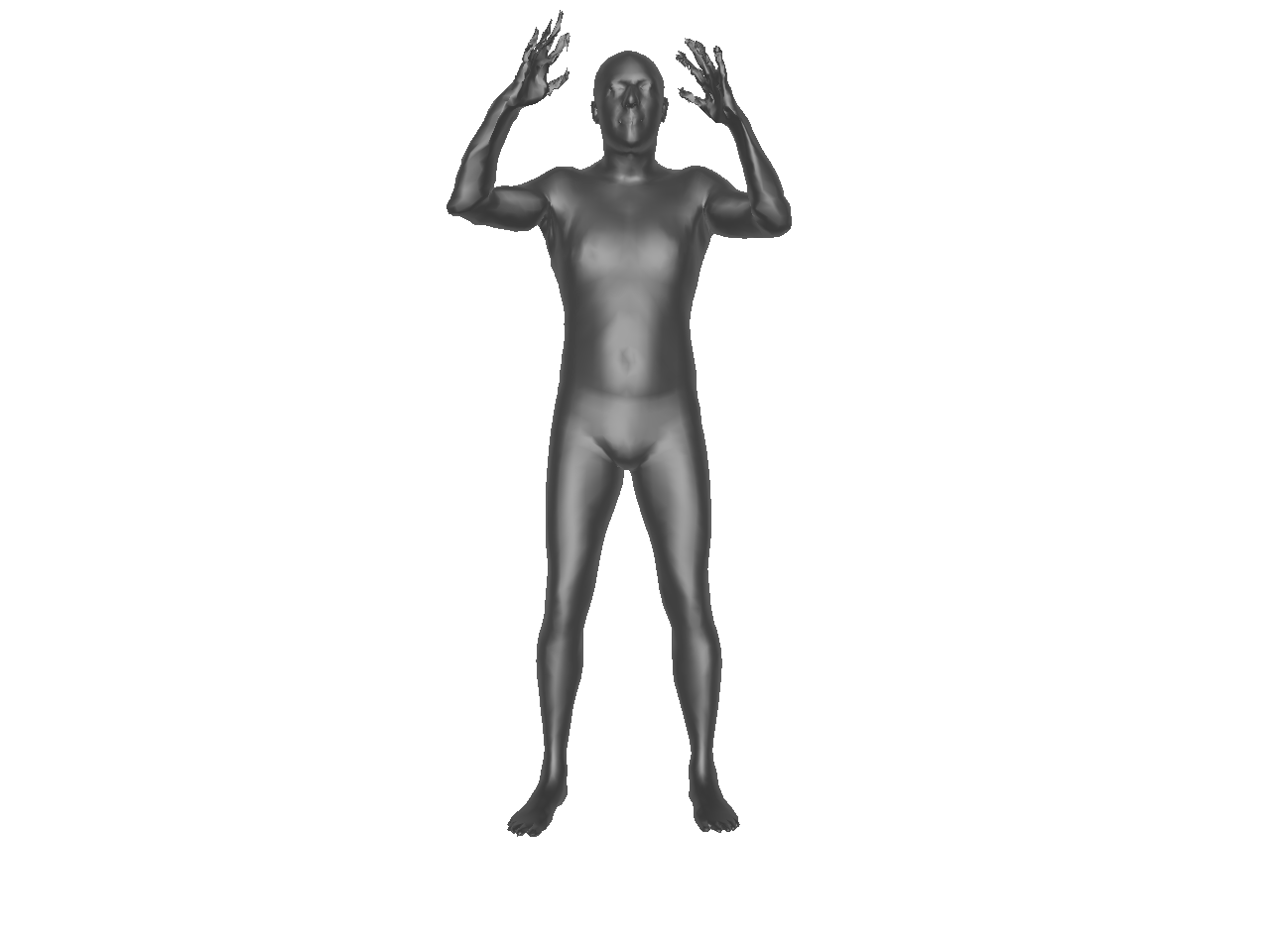}&
\includegraphics[trim={500 800 450 0},clip,height=2cm]
{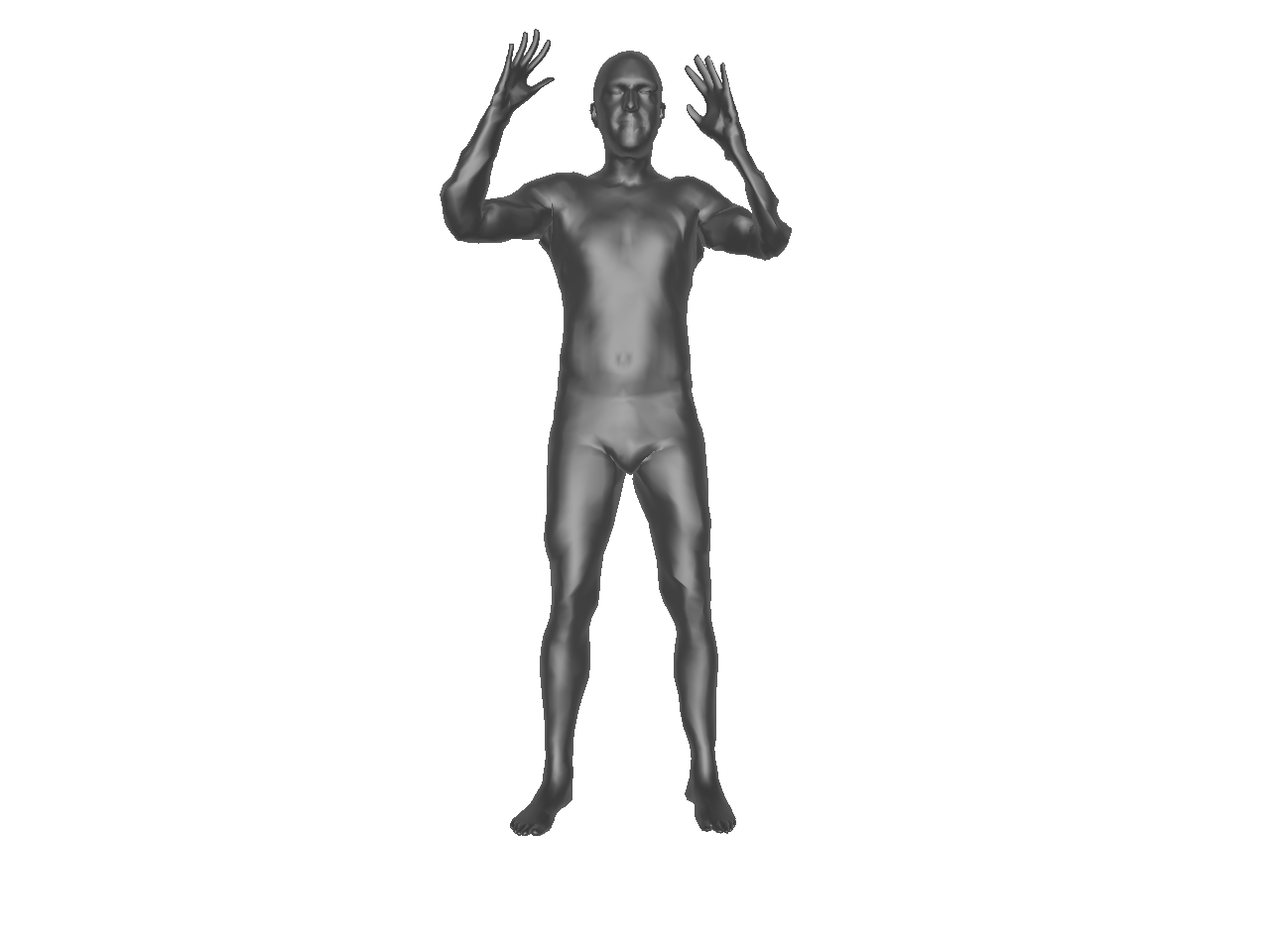}\\

\includegraphics[trim={500 250 550 480},clip,height=2cm]
{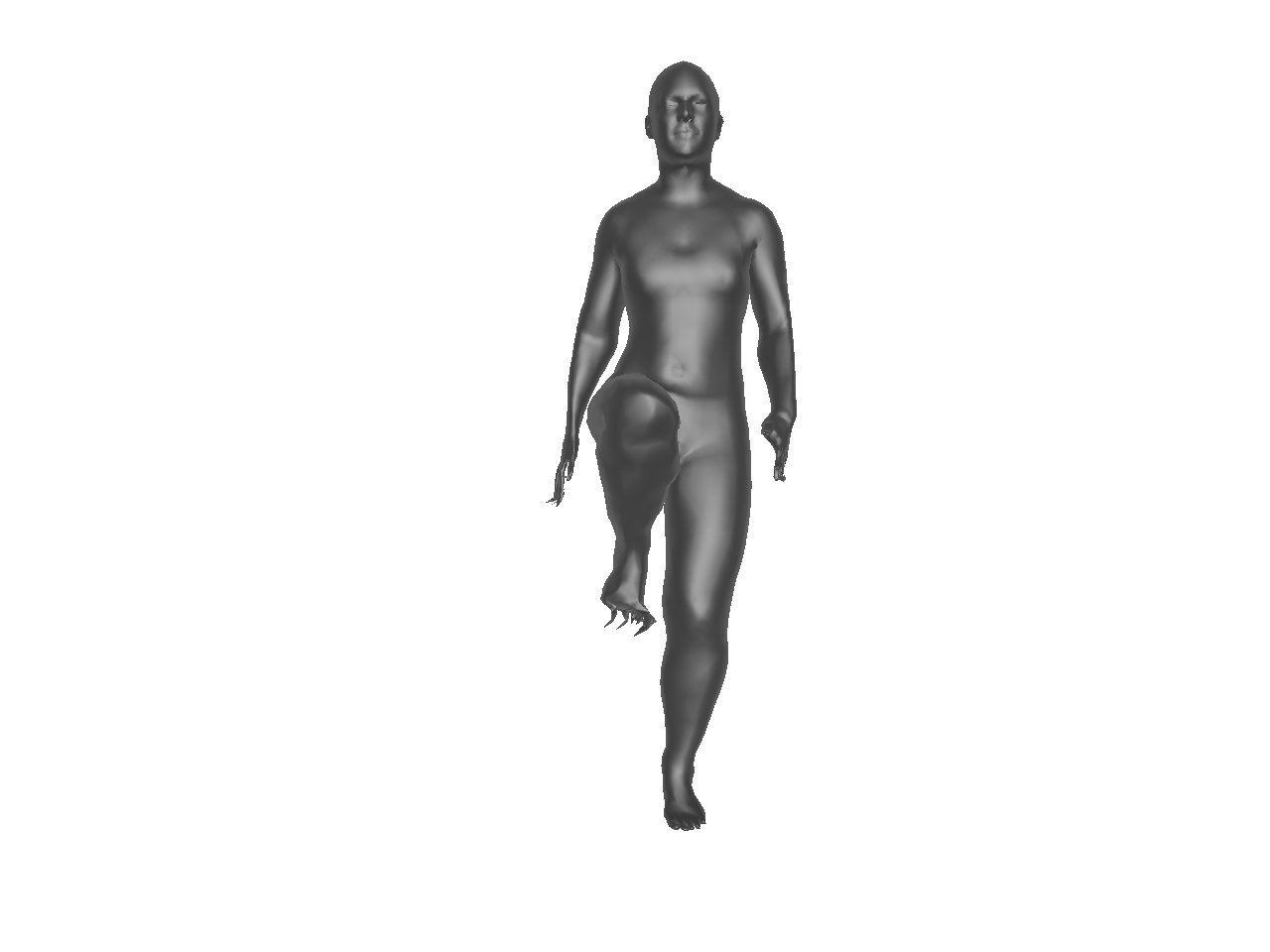}&
\includegraphics[trim={500 250 550 480},clip,height=2cm]
{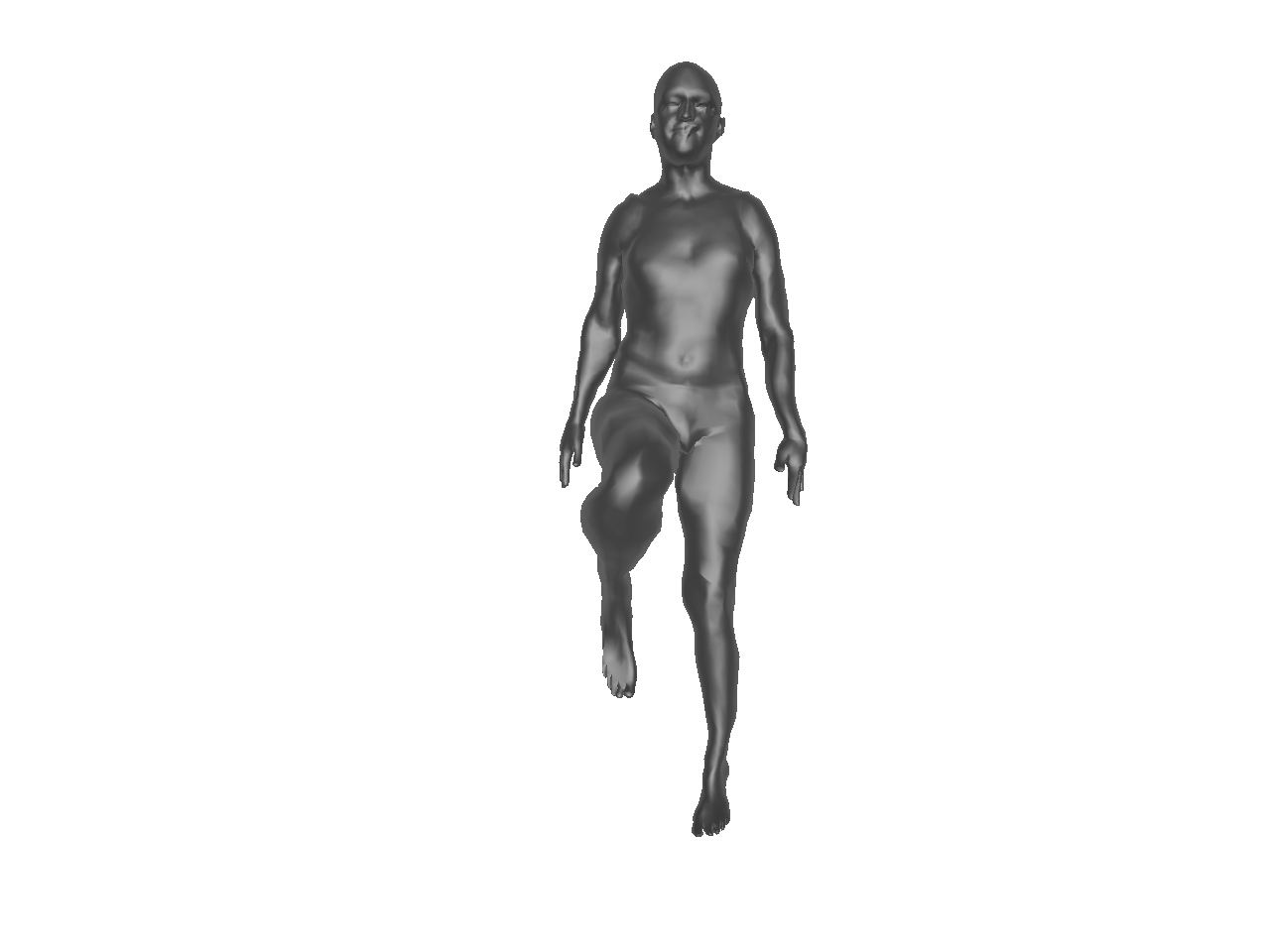}\\

\end{tabular}

\caption{Artifacts. The top rows use 32 latent dimensions, the last two rows are for 8.}
\label{fig:artifacts}
\end{table}

\begin{table}
\centering
\resizebox{\columnwidth}{!}{
\begin{tabular}{c|c|c|c}
Convolutional Ablation 8 & DEMEA 8 & Convolutional Ablation 32 & DEMEA 32\\\hline

\includegraphics[trim={500 200 500 350},clip,height=3cm]
{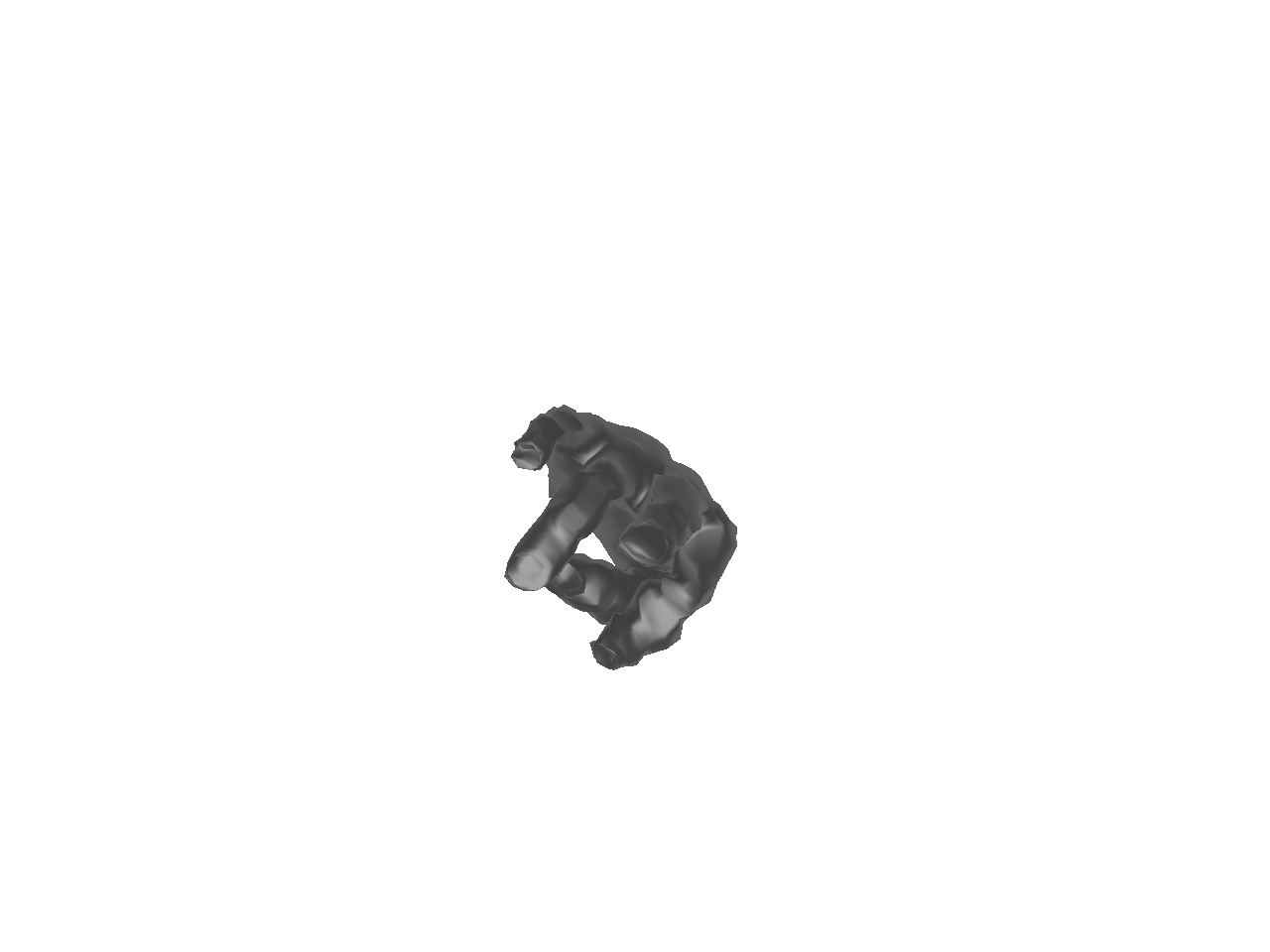} &
\includegraphics[trim={500 200 500 350},clip,height=3cm]
{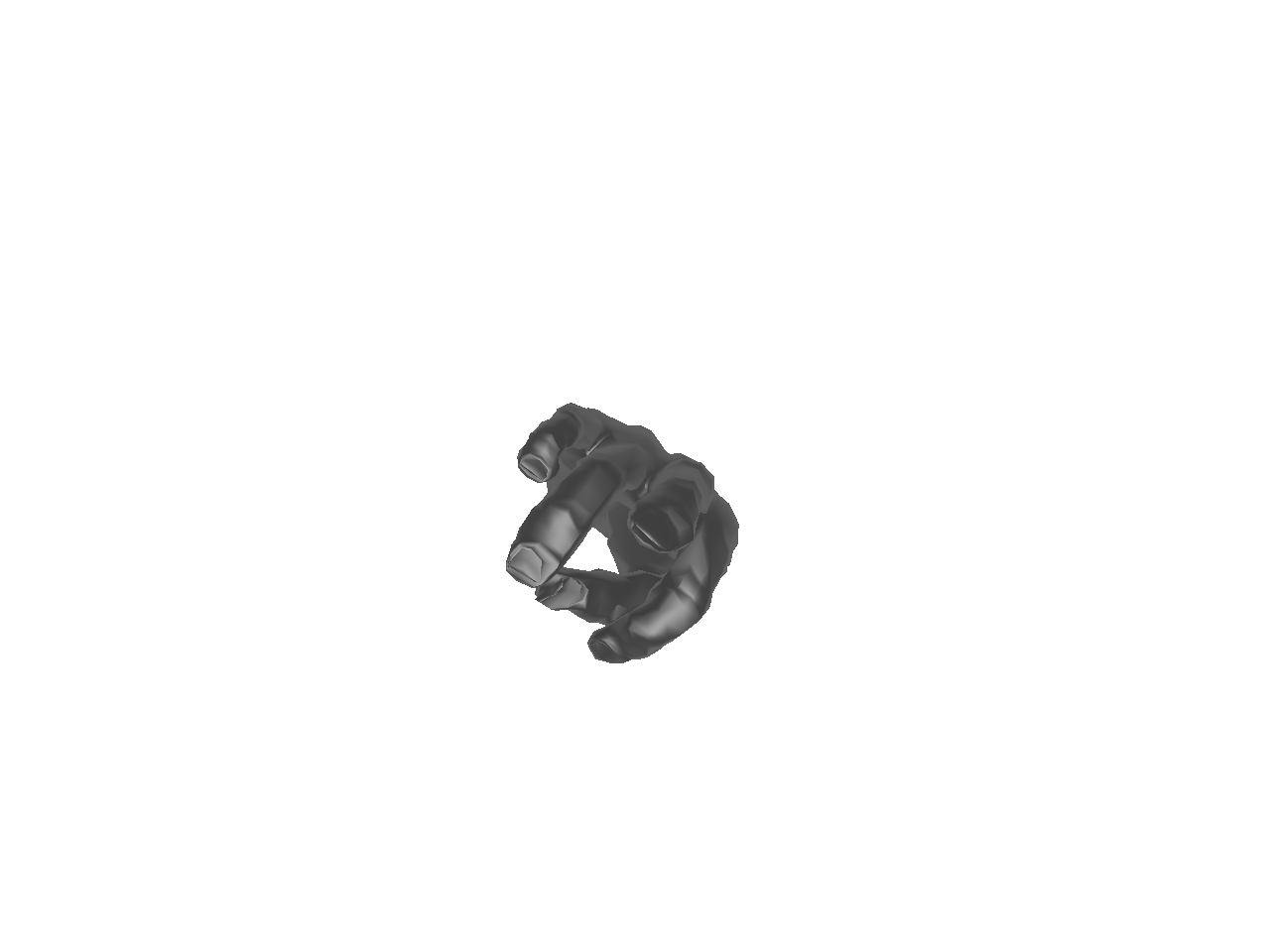} &
\includegraphics[trim={500 200 500 350},clip,height=3cm]
{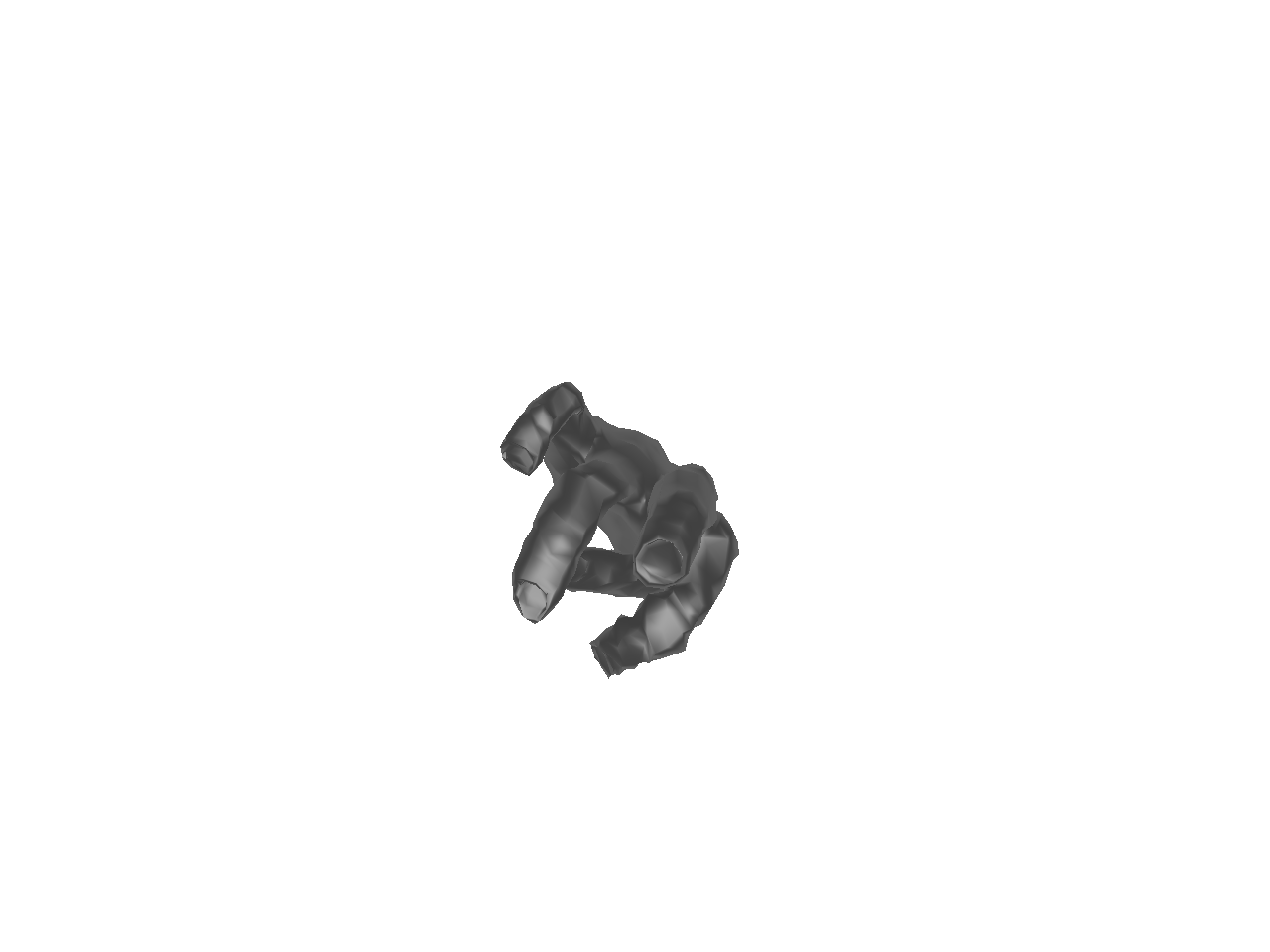} &
\includegraphics[trim={500 200 500 350},clip,height=3cm]
{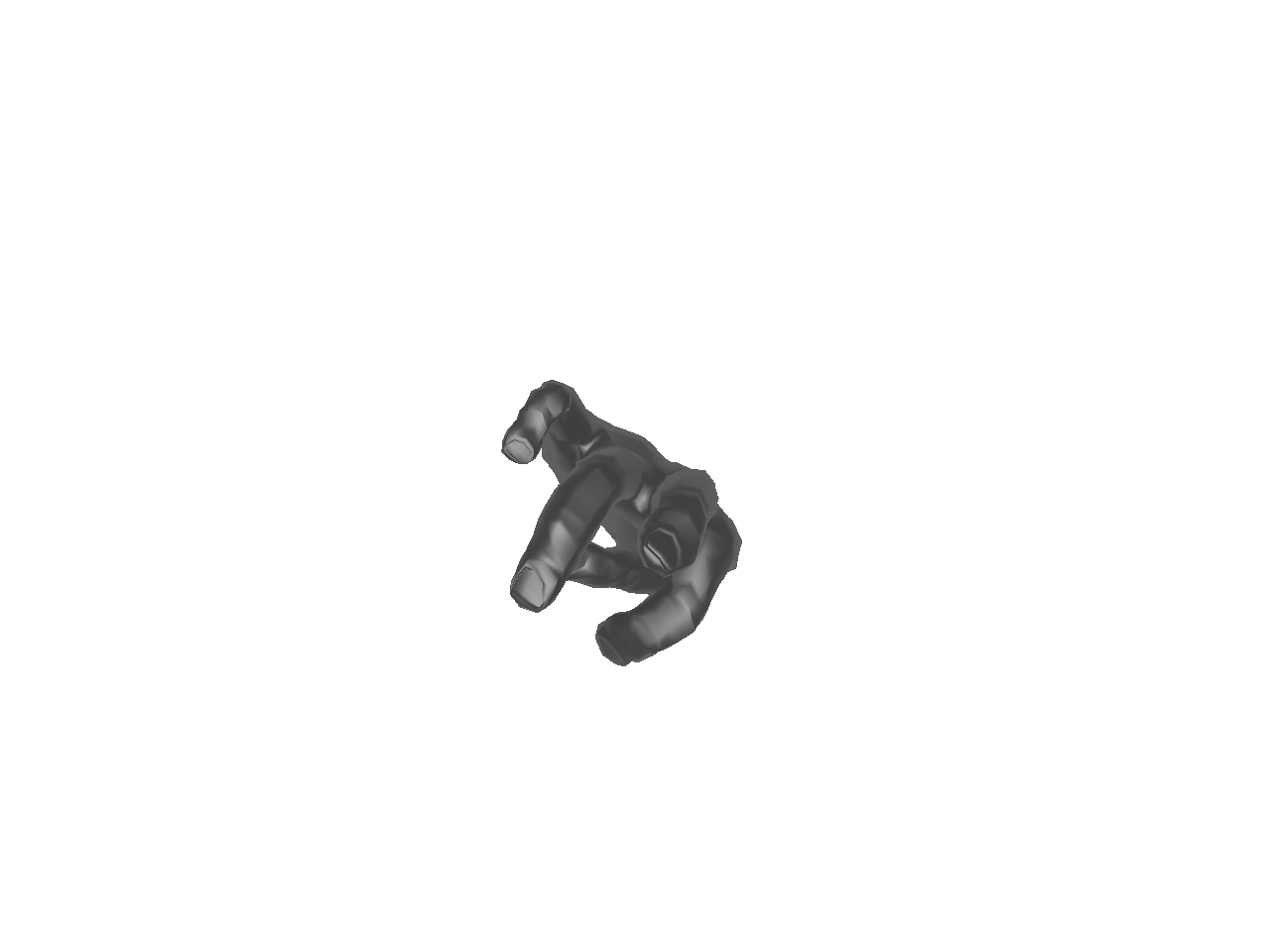}\\

\includegraphics[trim={400 200 400 400},clip,height=2.7cm,angle=90,origin=c]
{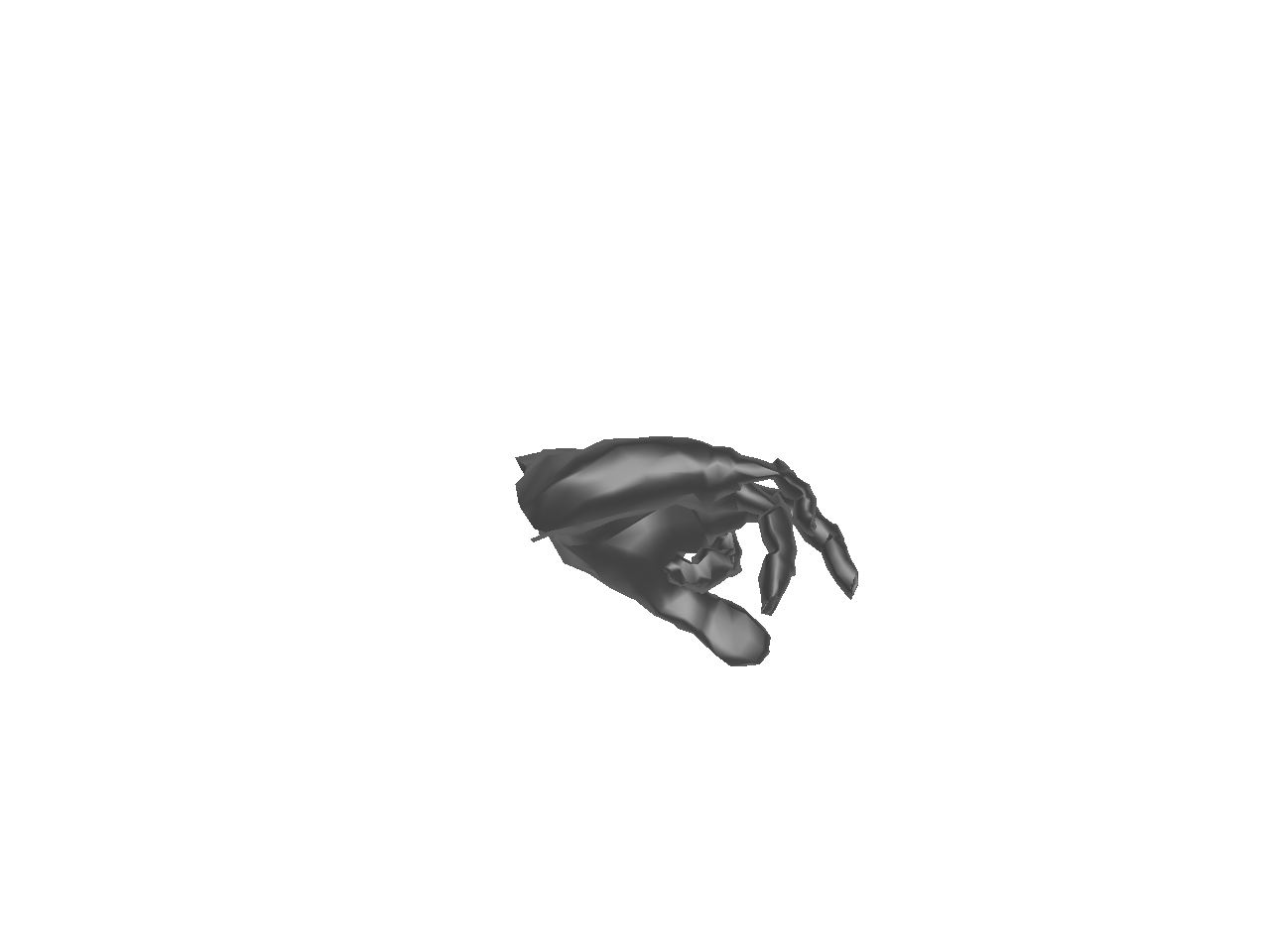} &
\includegraphics[trim={400 200 400 400},clip,height=2.7cm,angle=90,origin=c]
{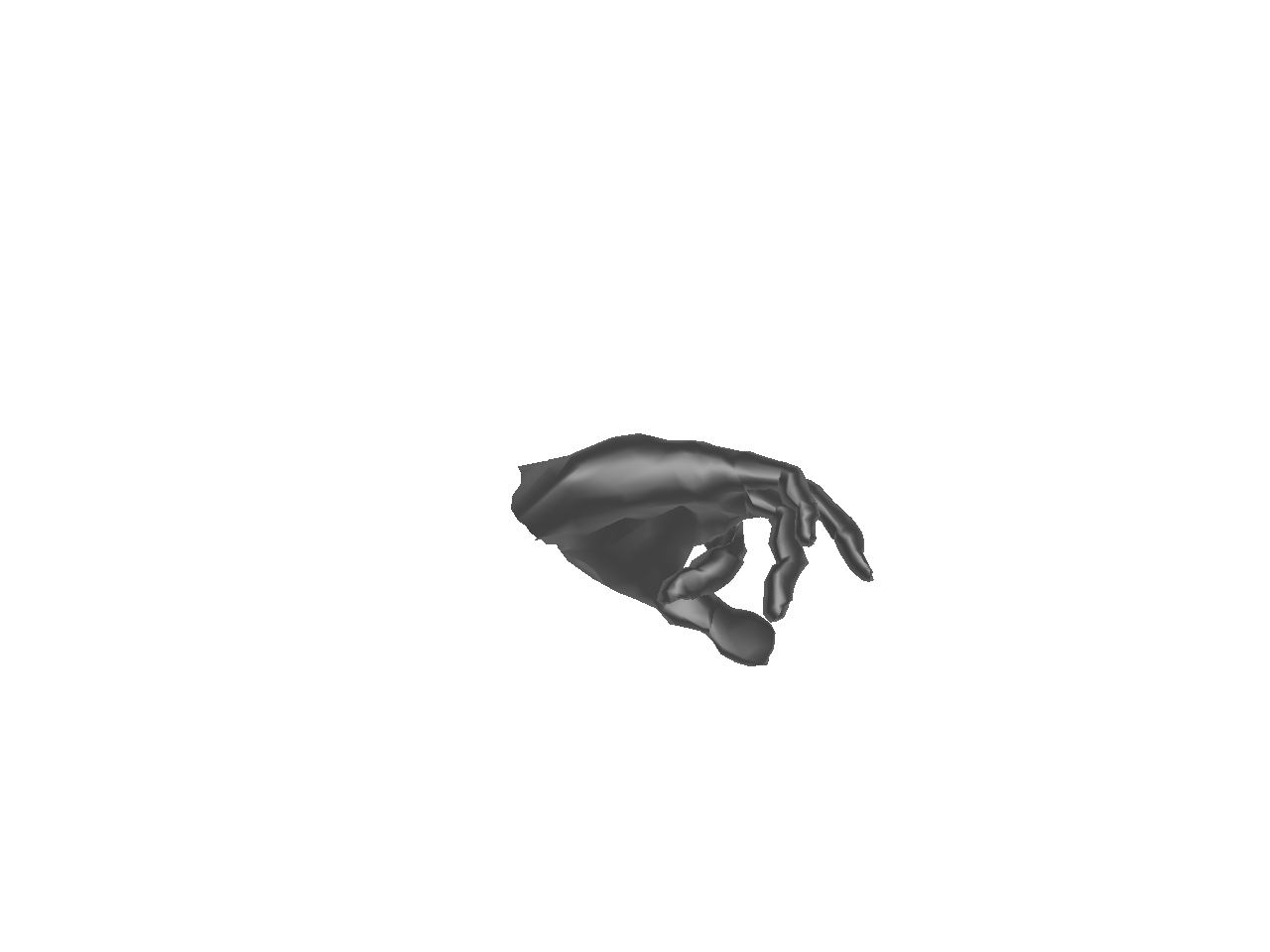} &
\includegraphics[trim={400 200 400 400},clip,height=2.7cm,angle=90,origin=c]
{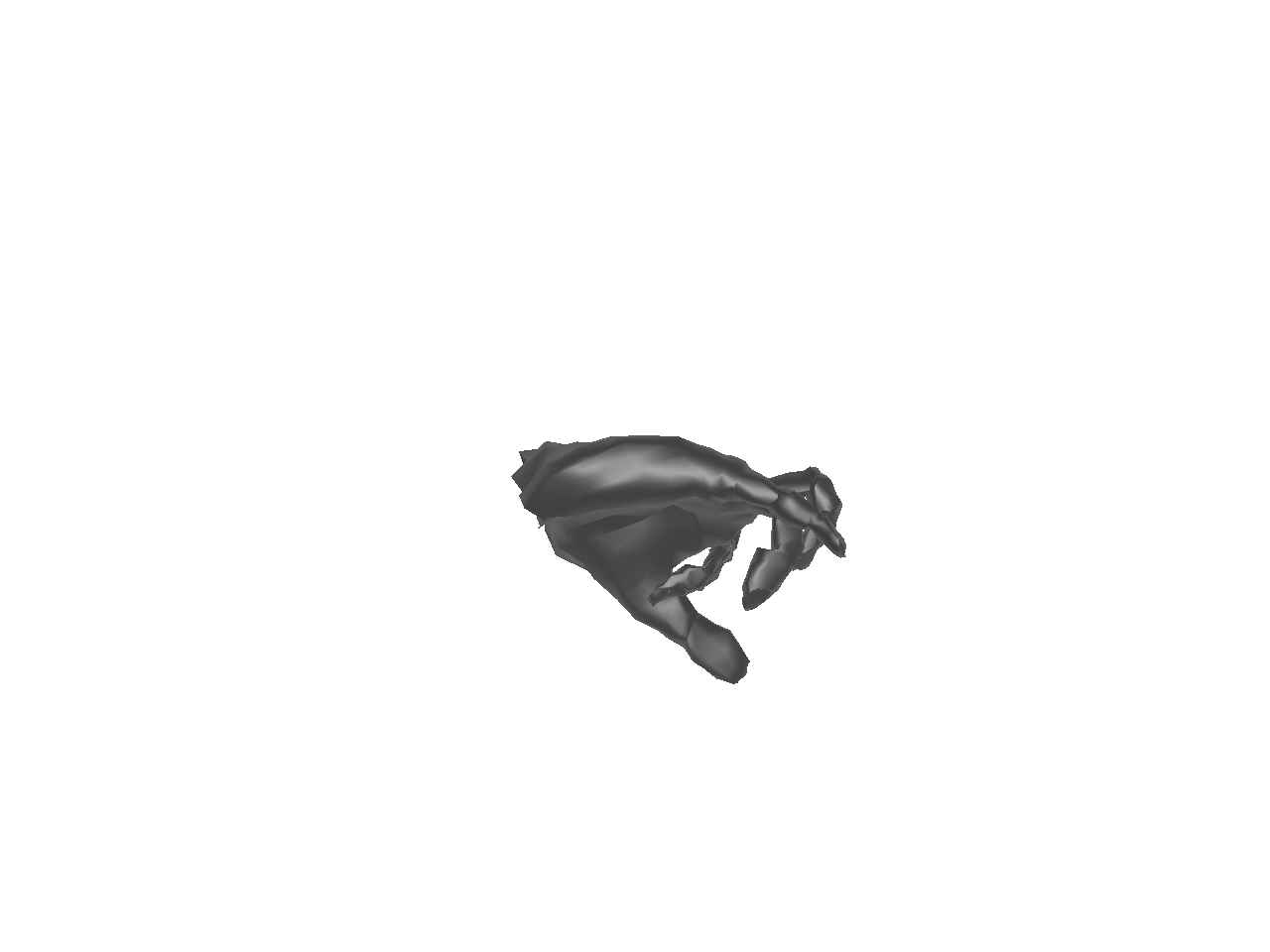} &
\includegraphics[trim={400 200 400 400},clip,height=2.7cm,angle=90,origin=c]
{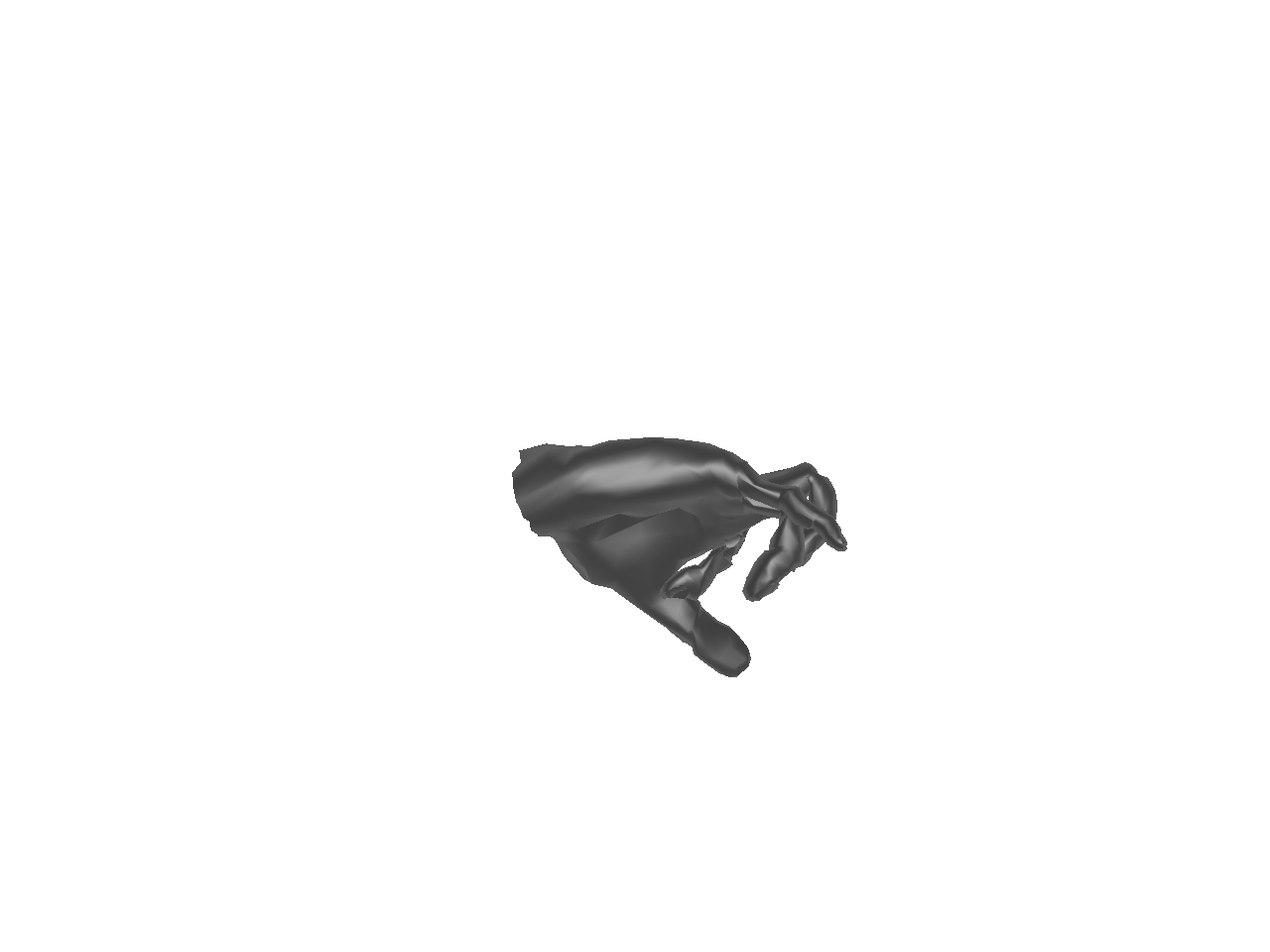}\\

\end{tabular}
}

\caption{Artifacts on SynHand5M. In contrast to CA, DEMEA yields a smooth index finger in the examples shown above.}
\label{fig:artifacts2}
\end{table}

\section{Architecture}\label{sec:architecture}

Fig.~\ref{fig:pipeline} contains our low-level architecture. 
\begin{figure*}
\begin{center}
\includegraphics*[width=\linewidth]{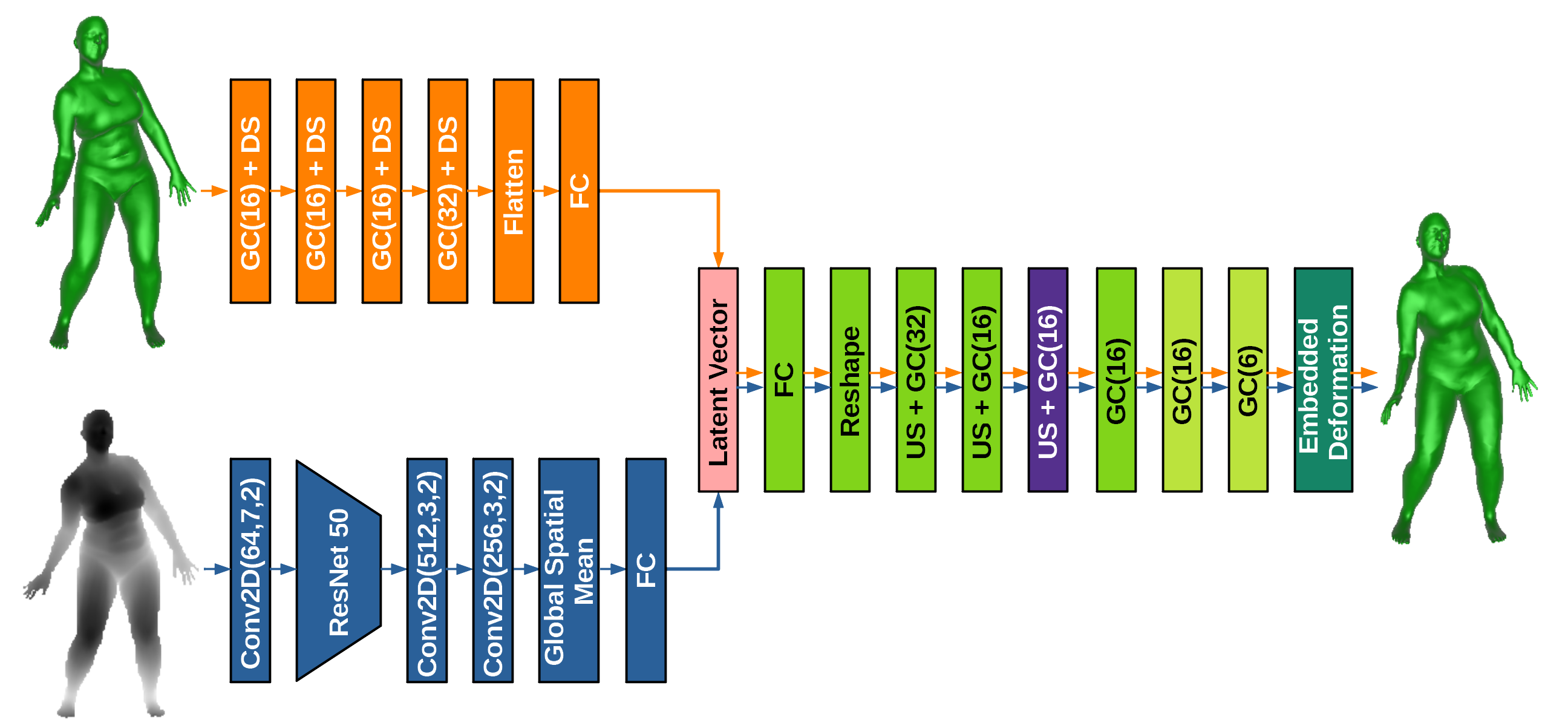}
\caption{The low-level architecture of DEMEA (orange path) and the depth-to-mesh network (blue path). Note that the two paths are not trained simultaneously.
}
\label{fig:pipeline}
\end{center} 
\end{figure*}
\emph{GC(f)} is a graph-convolutional layer with $f$ output features. 
\emph{DS} is a downsampling layer and \emph{US} is an upsampling layer. 
\emph{Conv2D(f,k,s)} is a 2D convolution with $f$ output features, kernel size $k\times k$ and stride $s$. 
We modified ResNetV2 50 by removing its first convolutional layer and its final non-convolutional layers, and keep its architecture unmodified otherwise.
We use ELU non-linearities after every graph-convolutional, 2D convolutional and fully-connected layer except for the first 2D convolutional layer in the depth encoder and the last graph-convolutional layer. 
The third upsampling module (\textit{i.e.} upsampling layer followed by a graph convolution) is only used for higher-resolution embedded graphs. 
In the case of spiral graph convolutions, we use the default settings of \cite{Bouritsas2019} to determine the length of the spirals.
In the case of spectral graph convolutions, we always use $K=6$, except for the last two layers, which use $K=2$ for local refinement.

\section{Losses}\label{sec:losses}

The vertex loss mentioned in Sec.~3.4 is:
\begin{equation}\label{eq:vertex}
    \mathcal{L}_{vertex}=\frac{1}{N_v} \sum_{i=1}^{N_v} \lVert \mathbf{\hat{v}}_i - \mathbf{v}^*_i  \rVert_1,
\end{equation}
where $\mathbf{\hat{v}}_i$ is the $i$-th deformed vertex and $\mathbf{v}^*_i$ is the $i$-th ground-truth vertex. 
The loss is averaged across the batch.

The deformed vertex can be either directly regressed (CA, MCA, FCA) or it can be computed via EDL (DEMEA, FCED). 
In the latter case, we follow Eq.~2 from the main manuscript.
While DEMEA regresses all EDL parameters, we also try out a modification that instead uses local Procrustes inside the network, called LP, see Sec.~3.4.
In that case, we use the same loss (Eq.~\ref{eq:vertex}) as for DEMEA but only regress the translation parameters, $\boldsymbol{t}_l$, while computing the rotation parameters of EDL, $\boldsymbol{R}_l$, using local Procrustes as described in Sec.~3.4.
Note that we do not backpropagate through the rotation computation in this case.

The graph loss described in Sec.~3.4 of the main manuscript operates on the graph nodes.
It encourages the regressed graph nodes positions to be close to the ground-truth vertex positions.
The graph nodes $\mathbf{N}$ are a subset of the mesh vertices $\mathbf{V}$ and we denote the vertex index corresponding to a graph node $l$ as $i_l$.
Then, the graph loss is:
\begin{equation}
    \mathcal{L}_{graph}=\frac{1}{L} \sum_{l=1}^{L} \lVert \mathbf{t}_l - \mathbf{v}^*_{i_l}  \rVert_1,
\end{equation}

\section{Normalization}\label{sec:normalization}
\paragraph{Depth}
All depth-to-mesh networks rescale the depth values of the input depth map from between $0.3m$ and $7m$ to $[-1,1]$. 
\paragraph{Bodies: Depth}
For our depth-to-mesh network on bodies, we employ a number of additional normalization steps to focus on non-rigid reconstruction.
First, we assume to be given a segmentation mask that filters out the background.
The depth value of background pixels is set to 2. 
We crop the foreground tightly and use bilinear sampling to isotropically rescale the crop to $256 \times 256$.
Given such a depth crop, we compute the average (foreground) depth value and subtract it from the input.
Such normalization necessitates normalizing the network output, as we will describe next.
\paragraph{Bodies: Meshes}
We first normalize out the global translation from the meshes by subtracting from each mesh vertex the average vertex position.
Since scale information is also lost, we fix the scale of the meshes by normalizing their approximate spine length.
To that end, we compute the approximate spine length of the template mesh and of each mesh in the dataset.
We then isotropically rescale all the meshes to the same spine length as the template mesh.
The depth-to-mesh body reconstruction errors in the main paper are reported for these normalized meshes.

\section{Standard Deviations in Table 2}\label{sec:std}

Due to space reasons, we could not fit standard deviations across the test set in Table 2 of the main manuscript.
Table~\ref{tab:std} contains the expanded version.

\begin{table}[]
\centering
\scalebox{0.7}{
\begin{tabular}{l | c | c | c | c | c | c | c | c |}
 & \multicolumn{2}{|c|}{DFaust} & \multicolumn{2}{|c|}{SynHand5M} & \multicolumn{2}{|c|}{Cloth} & \multicolumn{2}{|c|}{CoMA}\\
\cline{2-9}
                                   & 8   & 32   & 8  & 32  & 8  & 32  & 8  & 32  \\ \hline
\multicolumn{1}{|l|}{CA}           & 6.35 $\pm$ 2.40 & \textbf{2.07 $\pm$ 0.73} &  8.12 $\pm$ 1.77 & 2.60 $\pm$ 0.60 &  \textbf{11.21 $\pm$ 4.58} & 6.50 $\pm$ 1.85 &  \textbf{1.17 $\pm$ 0.47} & 0.72 $\pm$ 0.22  \\ %
\multicolumn{1}{|l|}{MCA}          & \textbf{6.21 $\pm$ 2.48} & 2.13 $\pm$ 0.79 &  \textbf{8.11 $\pm$ 1.77} & 2.67 $\pm$ 0.60 &  11.64 $\pm$ 4.58 & 6.59 $\pm$ 1.96 &  1.20 $\pm$ 0.46 & 0.71 $\pm$ 0.21  \\ %
\multicolumn{1}{|l|}{Ours}         & 6.69 $\pm$ 2.76 & 2.23 $\pm$ 0.99 &  8.12 $\pm$ 1.73 & \textbf{2.51 $\pm$ 0.59} &  11.28 $\pm$ 4.65 & 6.40 $\pm$ 1.96 &  1.23 $\pm$ 0.41 & 0.81 $\pm$ 0.22  \\ %
\multicolumn{1}{|l|}{FCA}          & 6.51 $\pm$ 2.45 & 2.17 $\pm$ 0.82 &  15.10 $\pm$ 4.06 & 2.95 $\pm$ 0.69 &  15.63 $\pm$ 7.18 & 5.99 $\pm$ 1.86 &  1.77 $\pm$ 0.57 & \textbf{0.67 $\pm$ 0.22}  \\ %
\multicolumn{1}{|l|}{FCED}         & 6.26 $\pm$ 2.35 & 2.14 $\pm$ 0.86 &  14.61 $\pm$ 3.95 & 2.75 $\pm$ 0.63 &  15.87 $\pm$ 7.73 & \textbf{5.94 $\pm$ 1.81} &  1.81 $\pm$ 0.71 & 0.73 $\pm$ 0.20  \\ \hline
\end{tabular}
}
\caption{Average per-vertex errors on the test sets of DFaust (in $cm$), SynHand5M (in $mm$), textureless cloth (in $mm$) and CoMA (in $mm$) for 8 and 32 latent dimensions, including standard deviations across the test sets.}
\label{tab:std}
\end{table}

\section{Mesh Hierarchy}\label{sec:hierarchy}

We use the code of \cite{Ranjan2018} to generate the mesh hierarchy. 
Then, in practice, the first or the second level of this automatically generated mesh hierarchy can be used as the embedded graph.

However, we propose to use more uniform embedded graphs.
Using MeshLab's \cite{Meshlab} implementation of quadric edge collapse decimation with default settings works well.
(Note that we decimate the mesh until we reach the same number of graph nodes as used by \cite{Ranjan2018}.)
We experimented with different ways of obtaining better embedded graphs, including hand-crafting them, but found no major differences, except that graphs generated by \cite{Ranjan2018} were too non-uniform.
Among the tested methods, MeshLab constitutes the least involved method of obtaining a uniform graph.

Since the embedded graph needs to be a subset of the mesh, graph nodes obtained this way need to be projected to their closest vertices. This may lead to multiple nodes projecting to the same vertex. We resolve this with a greedy assignment from nodes to vertices: looping over all nodes, the current node is assigned to its closest vertex that is not yet taken by another node. With the modification of the code of \cite{Ranjan2018} described in the main paper, we can then generate the mesh hierarchy for this embedded graph.

\begin{figure}
    \centering
    \includegraphics[width=\linewidth]{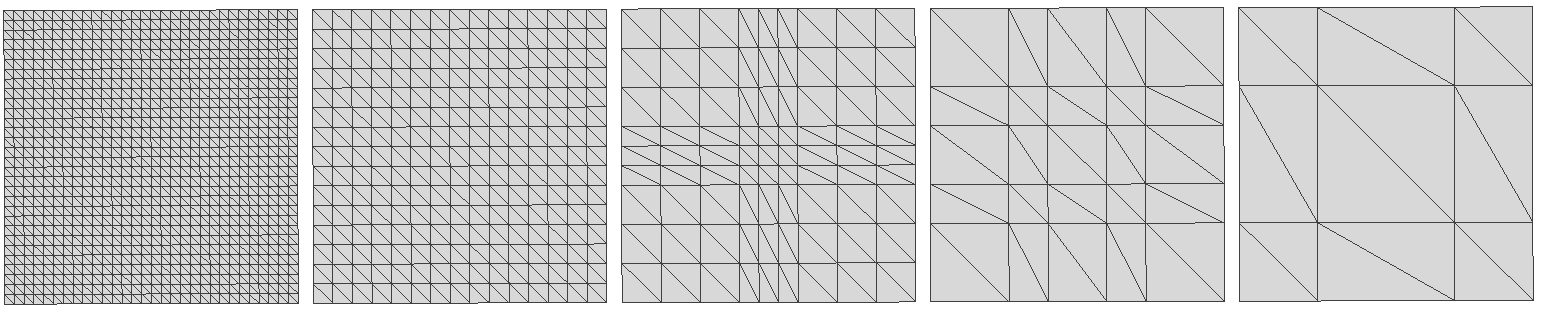}
    \caption{The Cloth hierarchy.}
    \label{fig:clothhierarchy}
\end{figure}

Figure~\ref{fig:clothhierarchy} shows the hierarchy we use for Cloth.

\section{Graph Convolutions}\label{sec:conv}

Our graph encoder-decoder architecture can work with multiple types of graph convolutions, which we show  in this section. 
The results of the main paper use spiral graph convolutions, as we found these to give slightly better results than spectral graph convolutions.
After defining the two types of graph convolutions, we report quantitative results.

Given an $F_{in}$-channel feature tensor $\mathbf{x} \in \mathbb{R}^{N \times F_{in}}$, where the features are defined at the $N$ graph nodes, let $\mathbf{x}_{*,i} \in \mathbb{R}^{N}$ denote the $i$-th input graph feature map. 
The complete output feature tensor, that stacks all $F_{out}$ feature maps, is denoted as $\mathbf{y} \in \mathbb{R}^{N \times F_{out}}$.
We apply the graph convolutions without stride, \textit{i.e.,} the input graph resolution equals the output resolution.

\subsection{Spiral Graph Convolutions}

We mainly work with spiral graph convolutions \cite{Bouritsas2019}. 
Let $\mathbf{x}_{n,*}^T\in \mathbb{R}^{F_{in}}$ denote the feature vector of graph node $n$. 
Assuming a fixed topology of the graph, we may choose an ordering of the neighboring nodes of graph node $n$: $n_0,\ldots,n_s,\ldots,n_{S-1}$. 
Bouritsas \textit{et~al.} pick a spiral ordering that starts with $n_0=n$ and proceeds along the 1-ring ($n^1_0, n^1_1, \ldots$), then the 2-ring ($n^2_0, n^2_1, \ldots$), and so on. 
The spiral ordering is thus given by: $n, n^1_0, n^1_1, \ldots, n^2_0, n^2_1, \ldots$
All the spirals of the graph are ultimately cut to a fixed length $S$ and zero-padded if necessary.
The output of the spiral convolution is then defined as:
\begin{equation}
    \mathbf{y}_{j,*}^T = \sum_{s=0}^{S-1}{G_s \cdot \mathbf{x}_{n_s,*}^T }\enspace{,}
\end{equation}
where $G_s \in F_{out} \times F_{in}$ is a trainable matrix.

\subsection{Spectral Graph Convolutions}
\label{ssec:spec_conv}
The second type of graph convolutions is based on fast localized spectral filtering \cite{Defferrard2016}, which Ranjan~\etal~use in CoMA~\cite{Ranjan2018}.
We compute the $j$-th output graph feature map $\mathbf{y}_{*,j} \in \mathbb{R}^{N}$ as follows:
\begin{equation}
    \mathbf{y}_{*,j} = \sum_{i=1}^{F_{in}}{g_{\theta_{i,j}}(\mathbf{L}) \cdot \mathbf{x}_{*,i} }\enspace{.}
\end{equation}
Here, $\mathbf{L}$ is the normalized Laplacian matrix of the graph and the filters $g_{\theta_{i,j}}(\mathbf{L})$ are parameterized using Chebyshev polynomials of order $K$.
More specifically, 
\begin{equation}
    g_{\theta_{i,j}}(\mathbf{L})=\sum_{k=0}^{K-1} \theta_{i,j,k} \cdot T_k(\mathbf{\widetilde{L}}), 
\end{equation}
where $\theta_{i,j,k}\in \mathbb{R}$ and $\mathbf{\widetilde{L}}=2\mathbf{L}/\lambda_{\textit{max}} - \mathbf{I}$, with $\lambda_{\textit{max}}=2$ being an upper bound on the spectrum of $\mathbf{L}$. The Chebychev polynomial $T_k$ is defined as $T_k(x) = 2 x \cdot T_{k-1}(x) - T_{k-2}(x)$, $T_1(x)=x$, and $T_0(x)=1$.

This leads to $K$-localized filters that operate on the $K$-neighbourhoods of the nodes.
Each filter $g_{\theta_{i,j}}(\mathbf{L})$ is parameterized by $K$ coefficients, which in total leads to $F_{in} \times F_{out} \times K$ trainable parameters for each graph convolution layer.

\subsection{Results}

We compare our proposed DEMEA with spiral convolutions against a version of DEMEA that uses spectral convolutions.
Table~\ref{tab:demeagraphconv} contains the results. 
Except for DFaust on latent dimension 8, spiral graph convolutions always perform at least slightly better than spectral graph convolutions.
These results show that EDL obtains similar accuracy with both graph convolutions, which further validates its robustness.

\begin{table}[]
\centering
\scalebox{0.8}{
\begin{tabular}{l | l  l | l  l | l  l | l  l |}
 & \multicolumn{2}{|c|}{DFaust} & \multicolumn{2}{|c|}{SynHand5M} & \multicolumn{2}{|c|}{Cloth} & \multicolumn{2}{|c|}{CoMA}\\
\cline{2-9}
                                & 8   & 32  & 8  & 32  & 8  & 32  & 8  & 32 \\ \hline
\multicolumn{1}{|l|}{Spiral}    & 6.69 & \textbf{2.23} & \textbf{8.12} & \textbf{2.51} & \textbf{11.28} & \textbf{6.40} & \textbf{1.23} & \textbf{0.81} \\ \hline
\multicolumn{1}{|l|}{Spectral}  & \textbf{6.56} & 2.40 & 8.74 & 3.83 & 11.76 & 6.52 & 1.40 & 0.98 \\ \hline
\end{tabular}
}
\caption{Average per-vertex errors on the test sets of DFaust (in $cm$), SynHand5M (in $mm$), textureless cloth (in $cm$) and CoMA (in $mm$). We compare two versions of DEMEA: one with spiral convolutions and one with spectral convolutions.}
\label{tab:demeagraphconv}
\end{table}

\section{FCA and CA Results}\label{sec:fca_ca}

Fig.~\ref{fig:fca} contains qualitative results for FCA and Fig.~\ref{fig:fca_artifacts} shows artifacts when using FCA.
Fig.~\ref{fig:ca} contains qualitative results for CA.
We show depth-to-mesh results in Fig.~\ref{fig:d2mbase}.

\begin{table}
\centering
\resizebox{\columnwidth}{!}{
\begin{tabular}{c|c|c|c|c}
Ground-truth & FCA 32 & DEMEA 32 & FCA 8 & DEMEA 8 \\\hline
\includegraphics[trim={300 0 300 0},clip,height=3.5cm]
{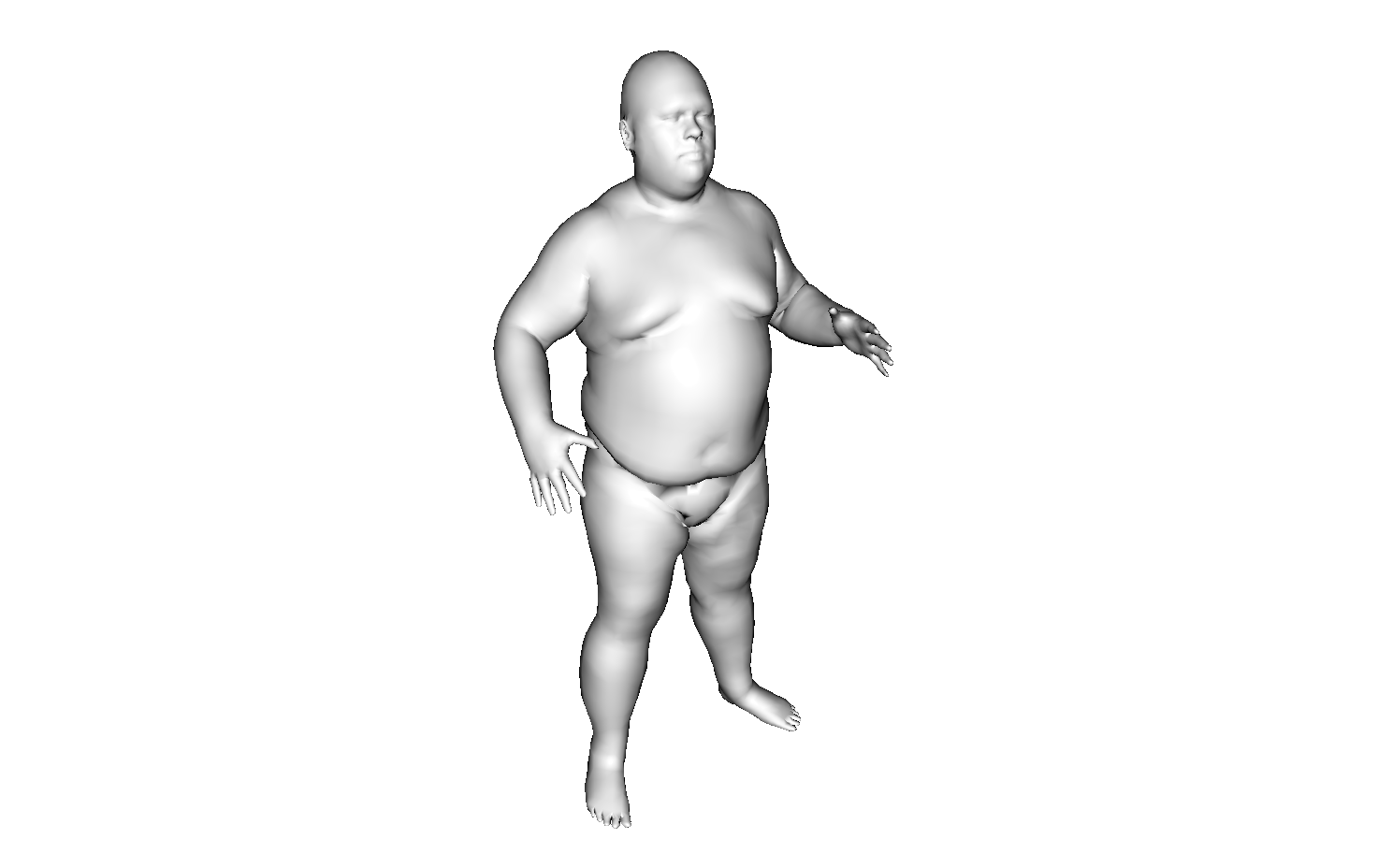}
&
\includegraphics[trim={300 0 300 0},clip,height=3.5cm]
{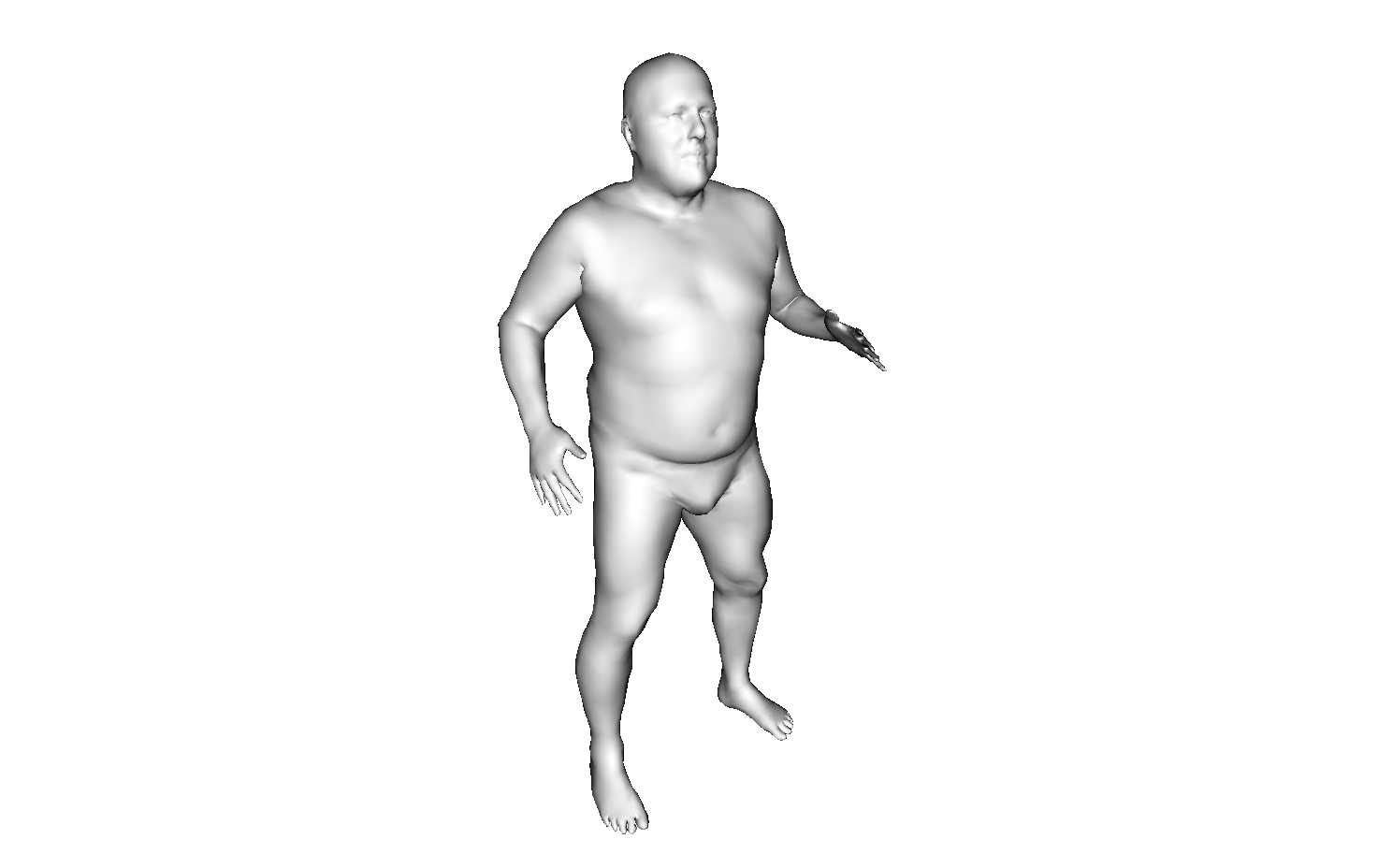}
&
\includegraphics[trim={300 0 300 0},clip,height=3.5cm]
{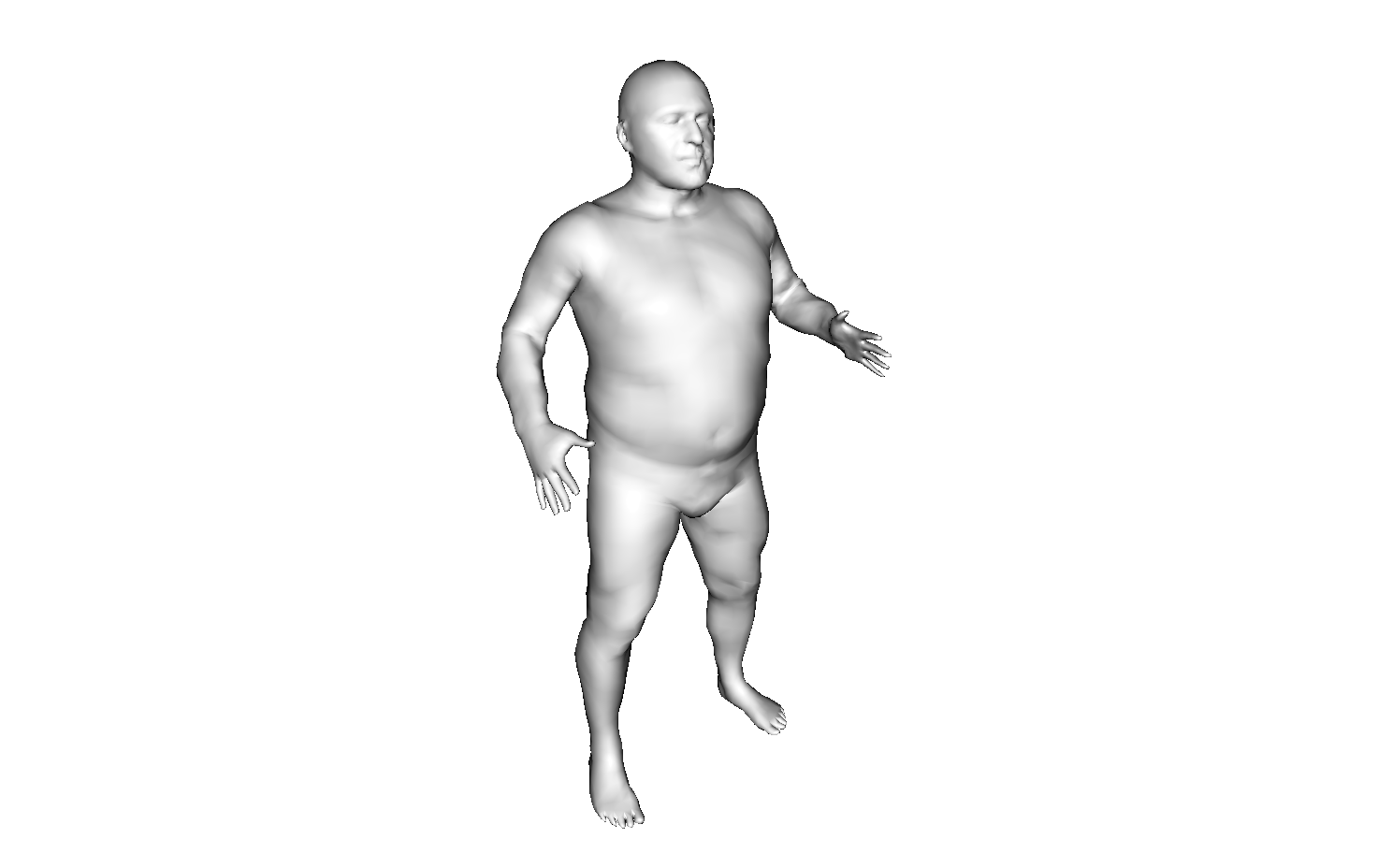}
&
\includegraphics[trim={300 0 300 0},clip,height=3.5cm]
{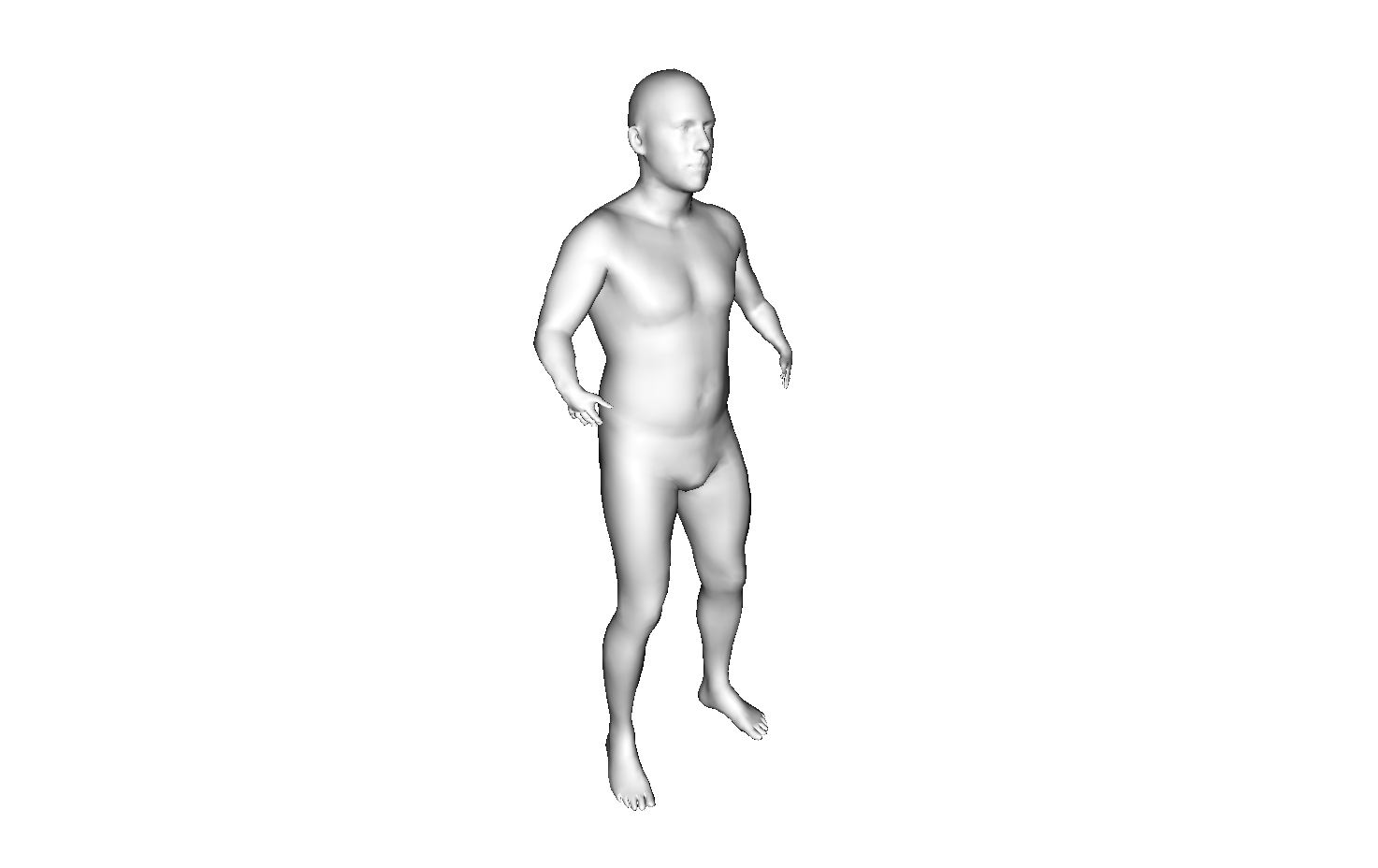}
&
\includegraphics[trim={300 0 300 0},clip,height=3.5cm]
{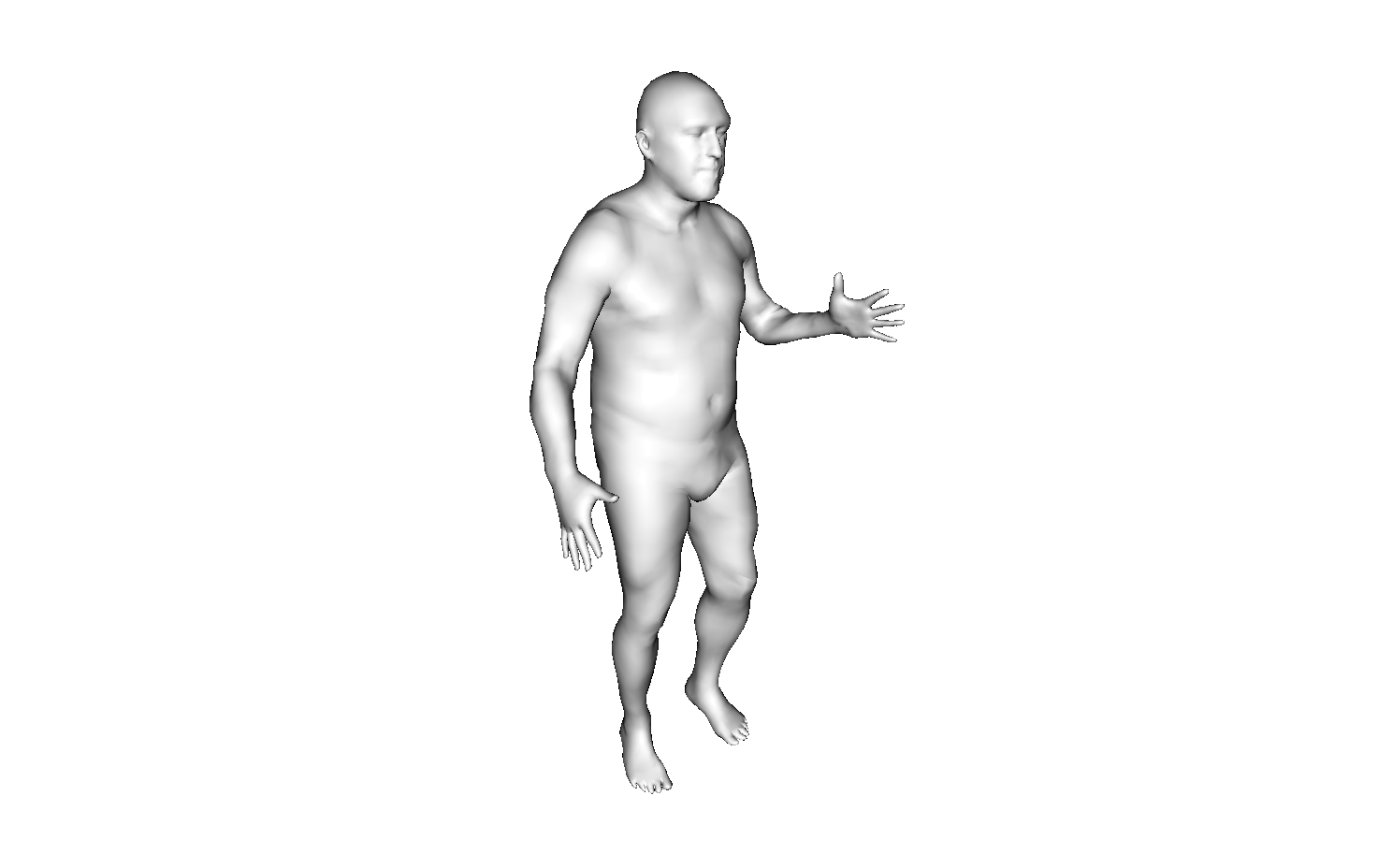}
\end{tabular}
}
\caption{Qualitative Results on FCA.}
\label{fig:fca}
\end{table}

\begin{table}
\centering
\resizebox{\columnwidth}{!}{
\begin{tabular}{c|c|c}
Ground-truth & FCA 32 & DEMEA 32 \\\hline
\includegraphics[trim={450 200 200 200},clip,height=3.5cm]
{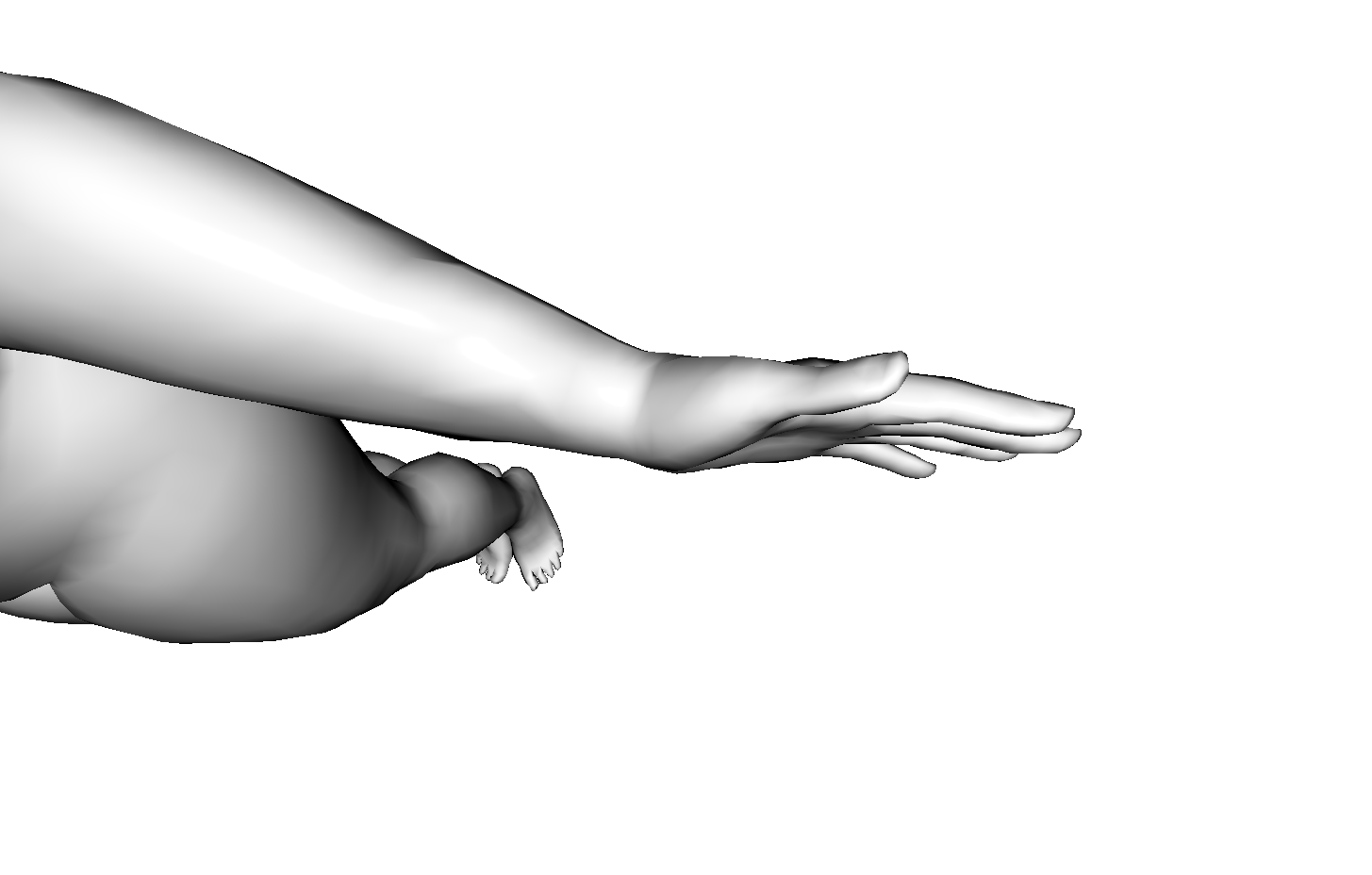}
&
\includegraphics[trim={450 200 200 200},clip,height=3.5cm]
{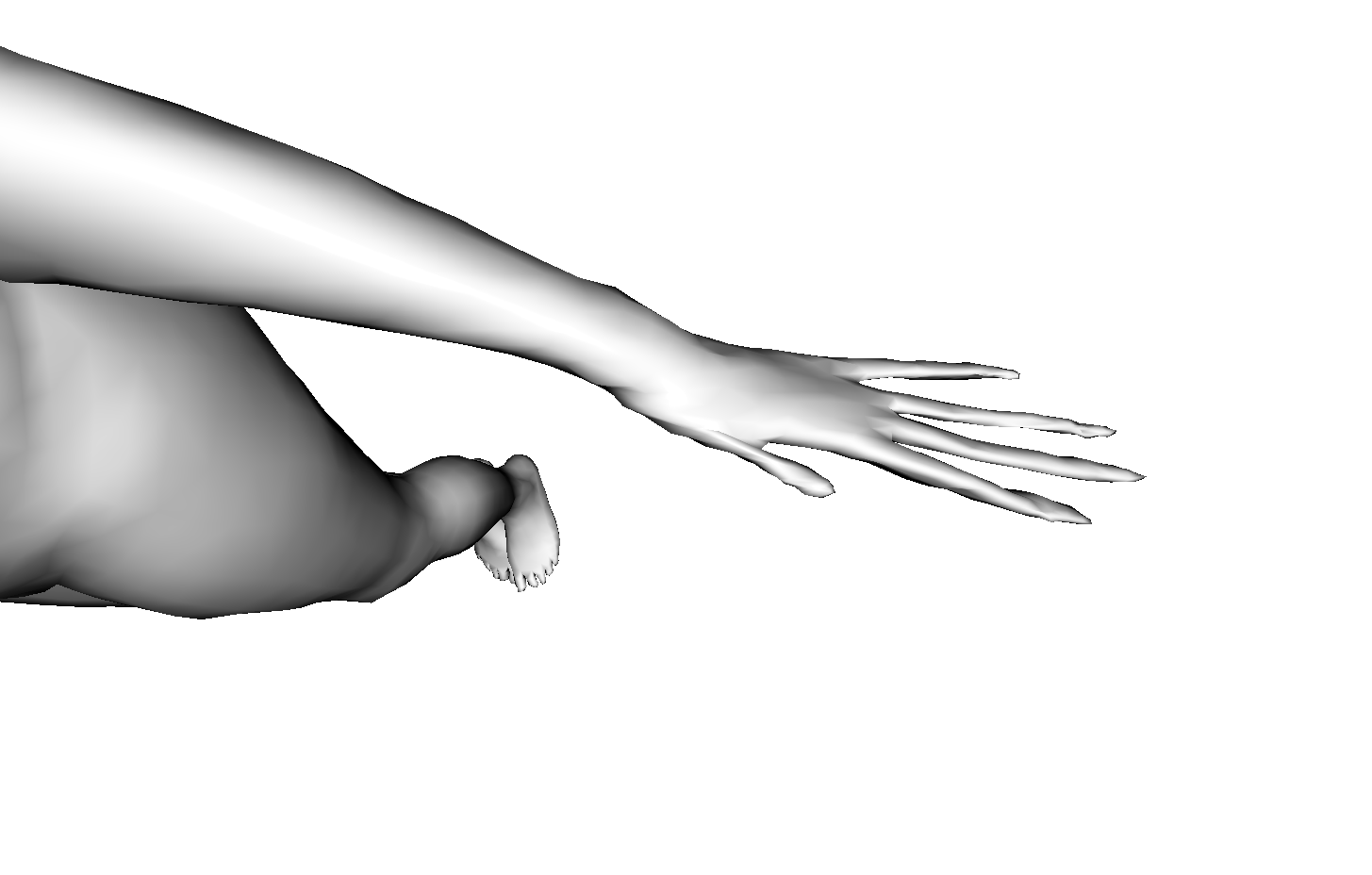}
&
\includegraphics[trim={450 200 200 200},clip,height=3.5cm]
{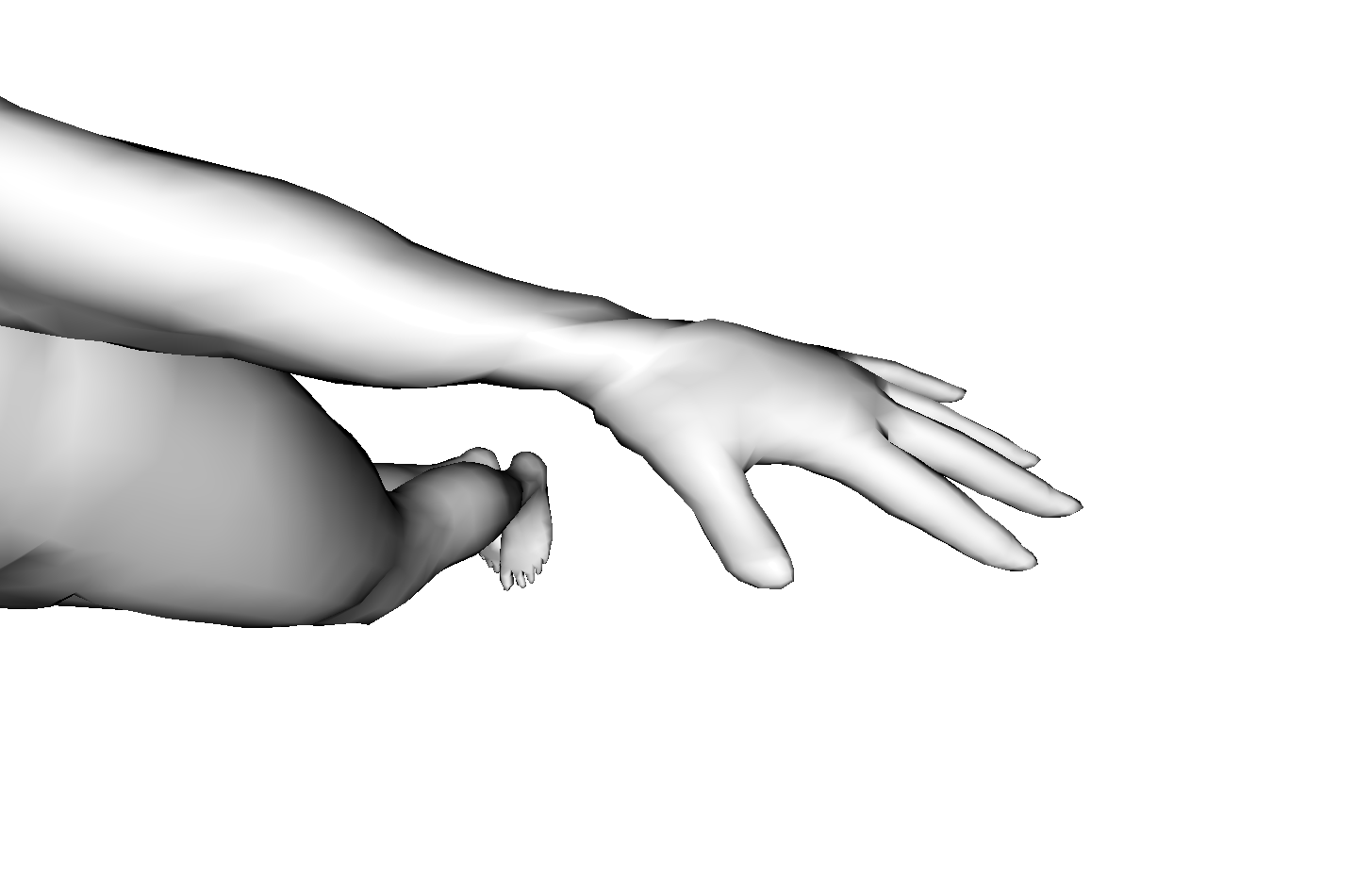}
\end{tabular}
}
\caption{Artifacts on FCA. While the reconstruction by DEMEA only matches the ground-truth as well as FCA, it is a significantly more plausible shape.}
\label{fig:fca_artifacts}
\end{table}

\begin{table}
\centering
\resizebox{\columnwidth}{!}{
\begin{tabular}{c|c|c|c|c}
Ground-truth & CA 32 & DEMEA 32 & CA 8 & DEMEA 8 \\\hline
\includegraphics[trim={300 0 300 0},clip,height=3.5cm]
{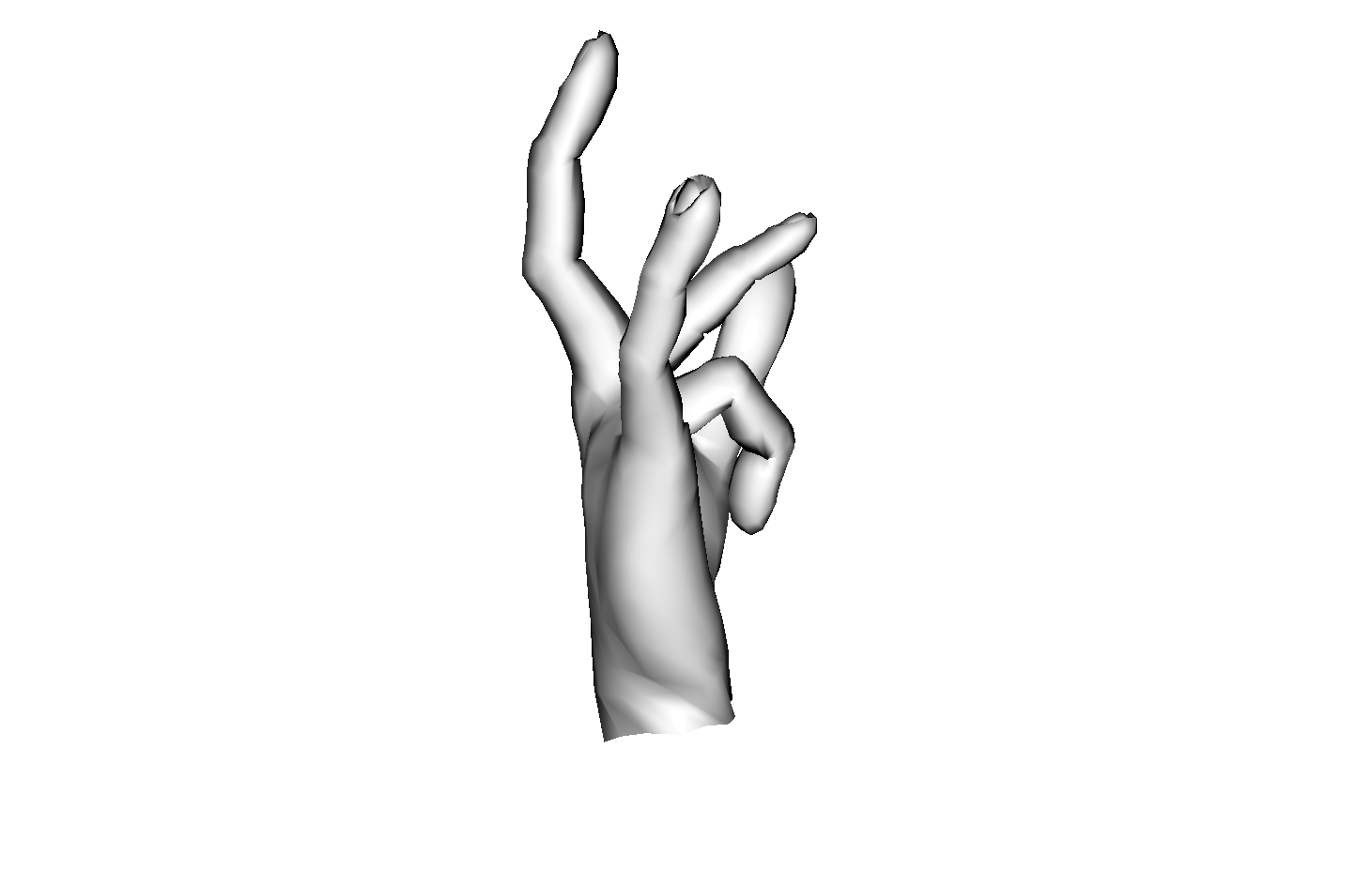}
&
\includegraphics[trim={300 0 300 0},clip,height=3.5cm]
{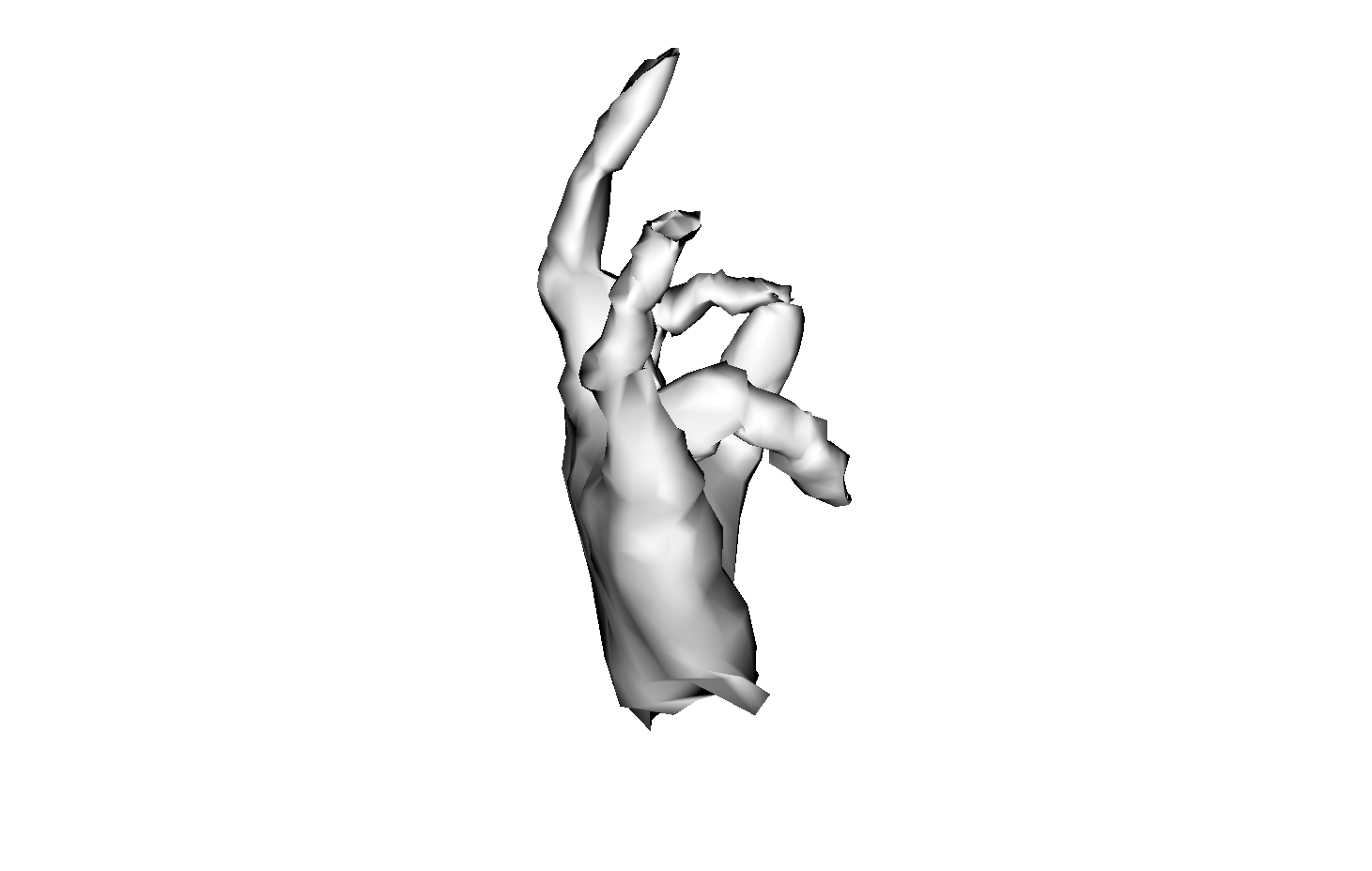}
&
\includegraphics[trim={300 0 300 0},clip,height=3.5cm]
{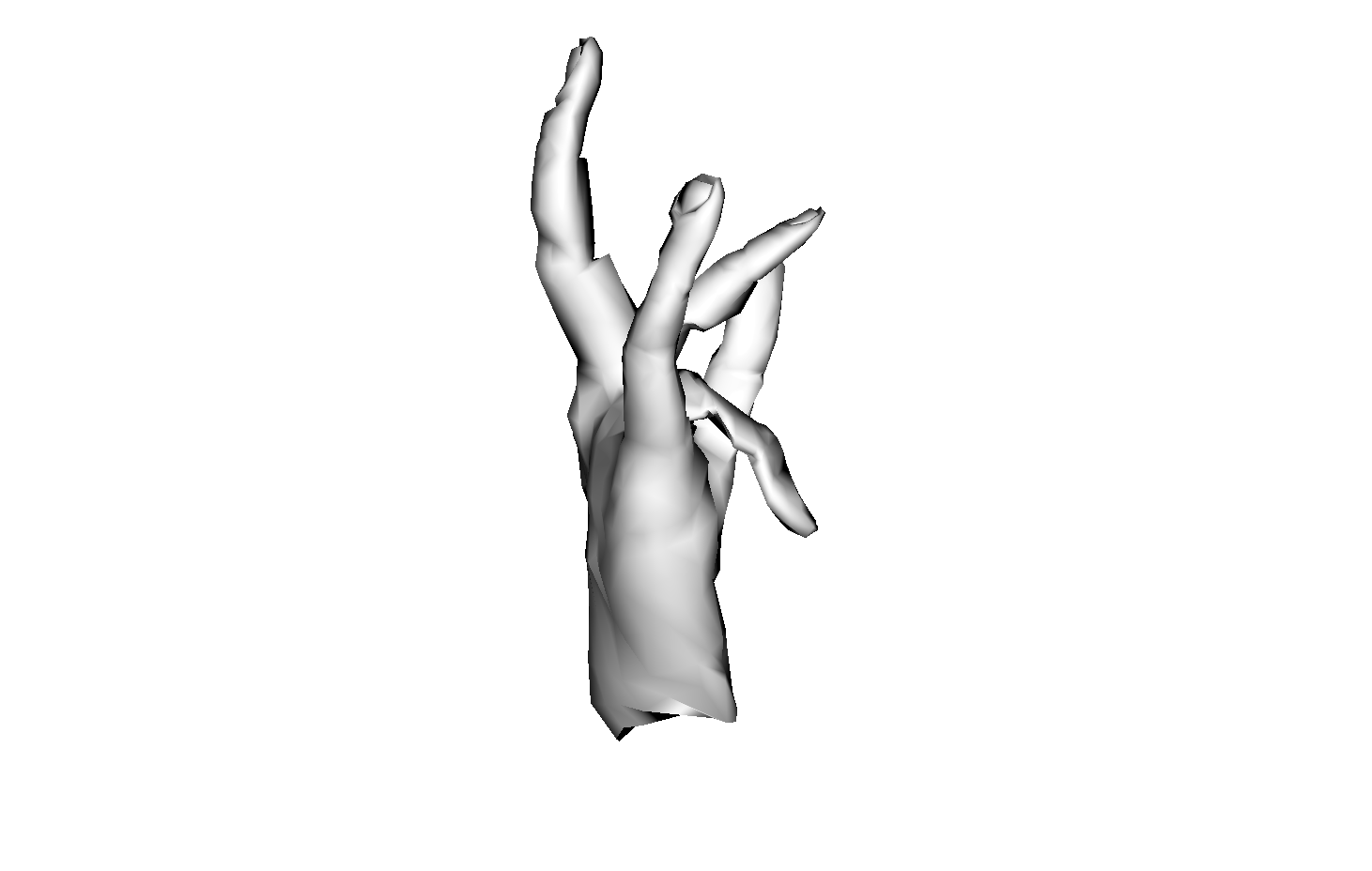}
&
\includegraphics[trim={300 0 300 0},clip,height=3.5cm]
{FIGURES/FCA_CA/hand-vertices-000000090038_____paper_2_missing_auto_shandh_split_meshlab_coma_ablation_spiral_hand_4_accounted_6_free_graph_l804.png}
&
\includegraphics[trim={300 0 300 0},clip,height=3.5cm]
{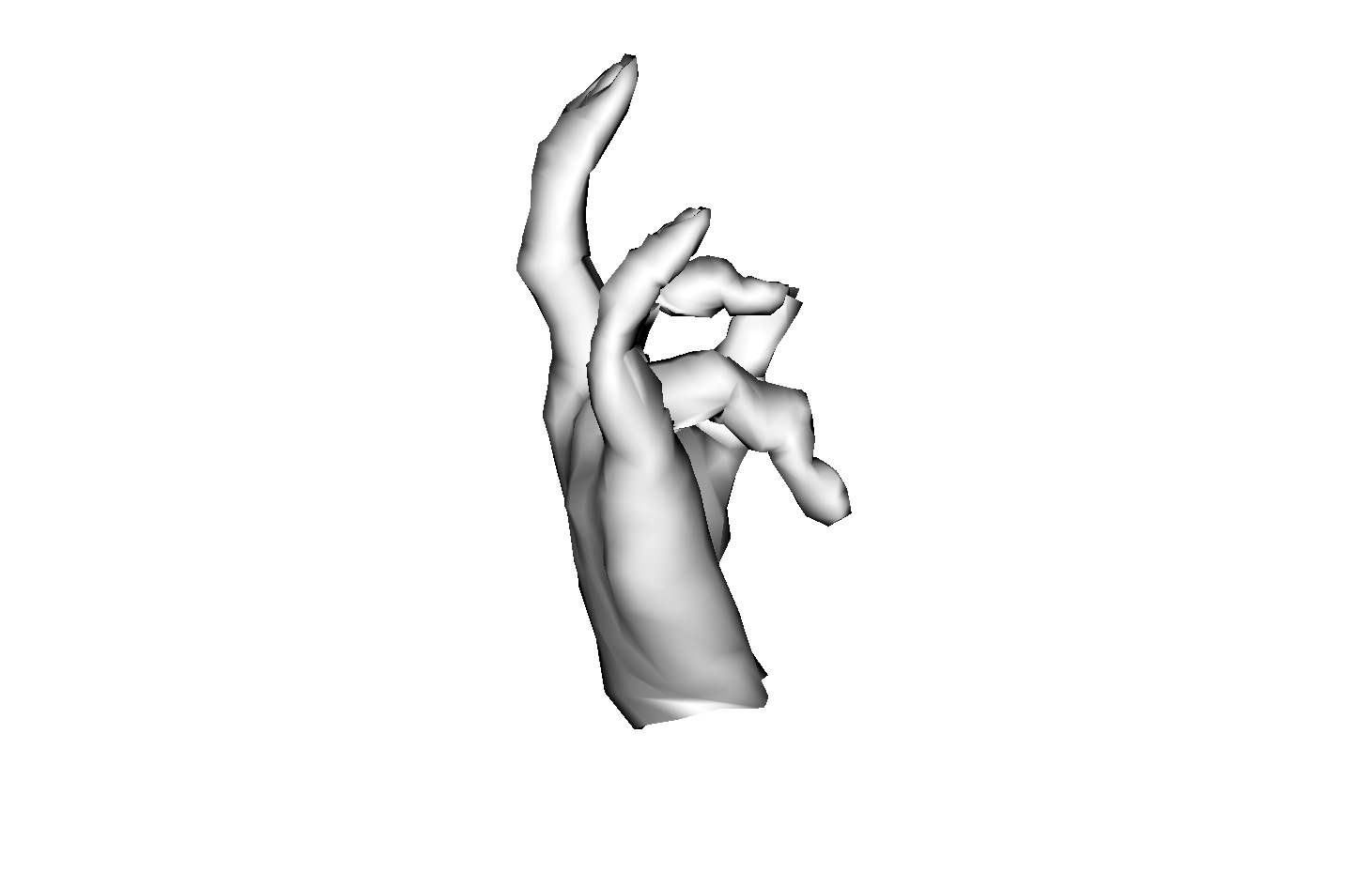}

\\

\includegraphics[trim={300 0 300 0},clip,height=3.5cm]
{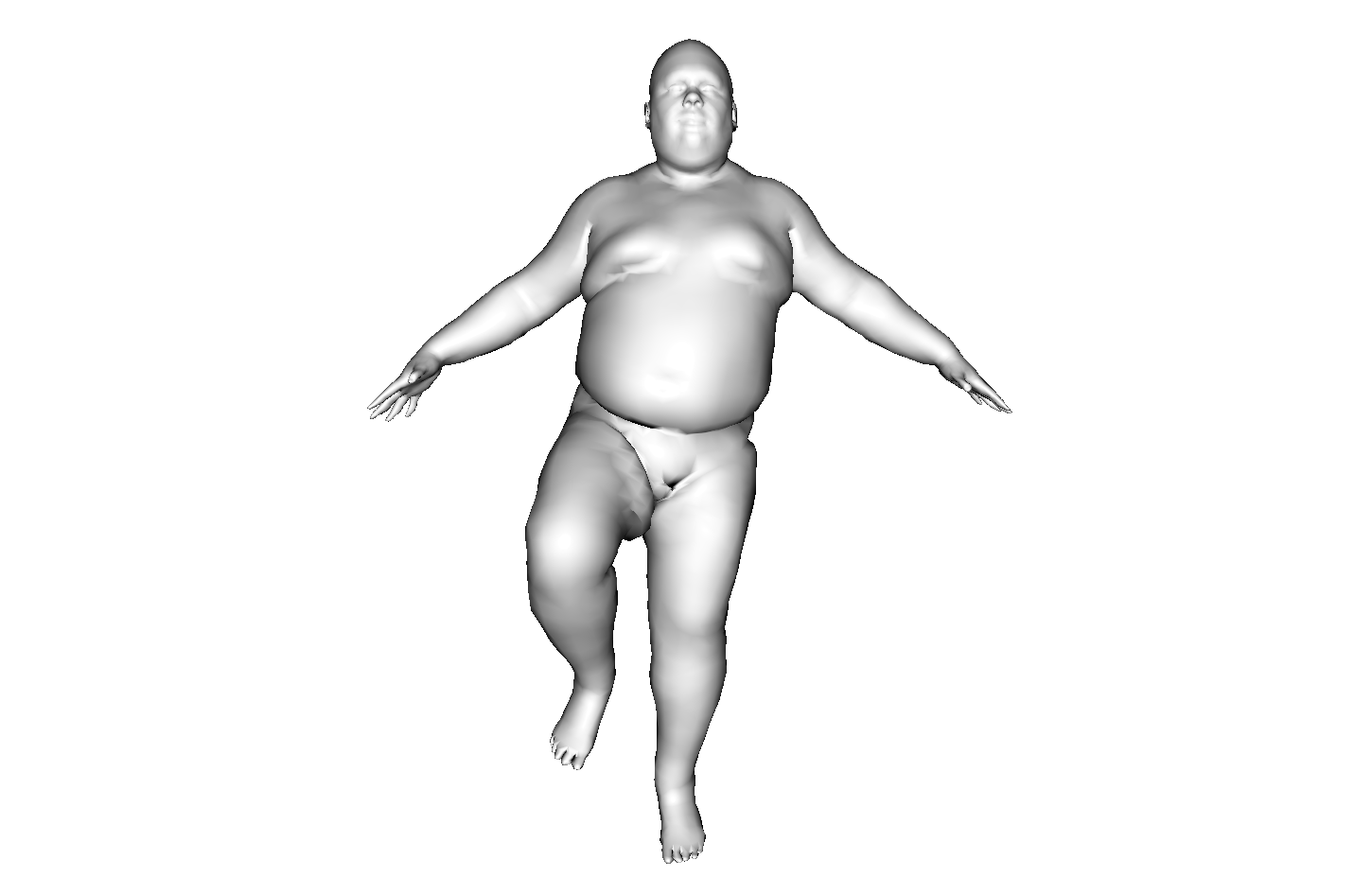}
&
\includegraphics[trim={300 0 300 0},clip,height=3.5cm]
{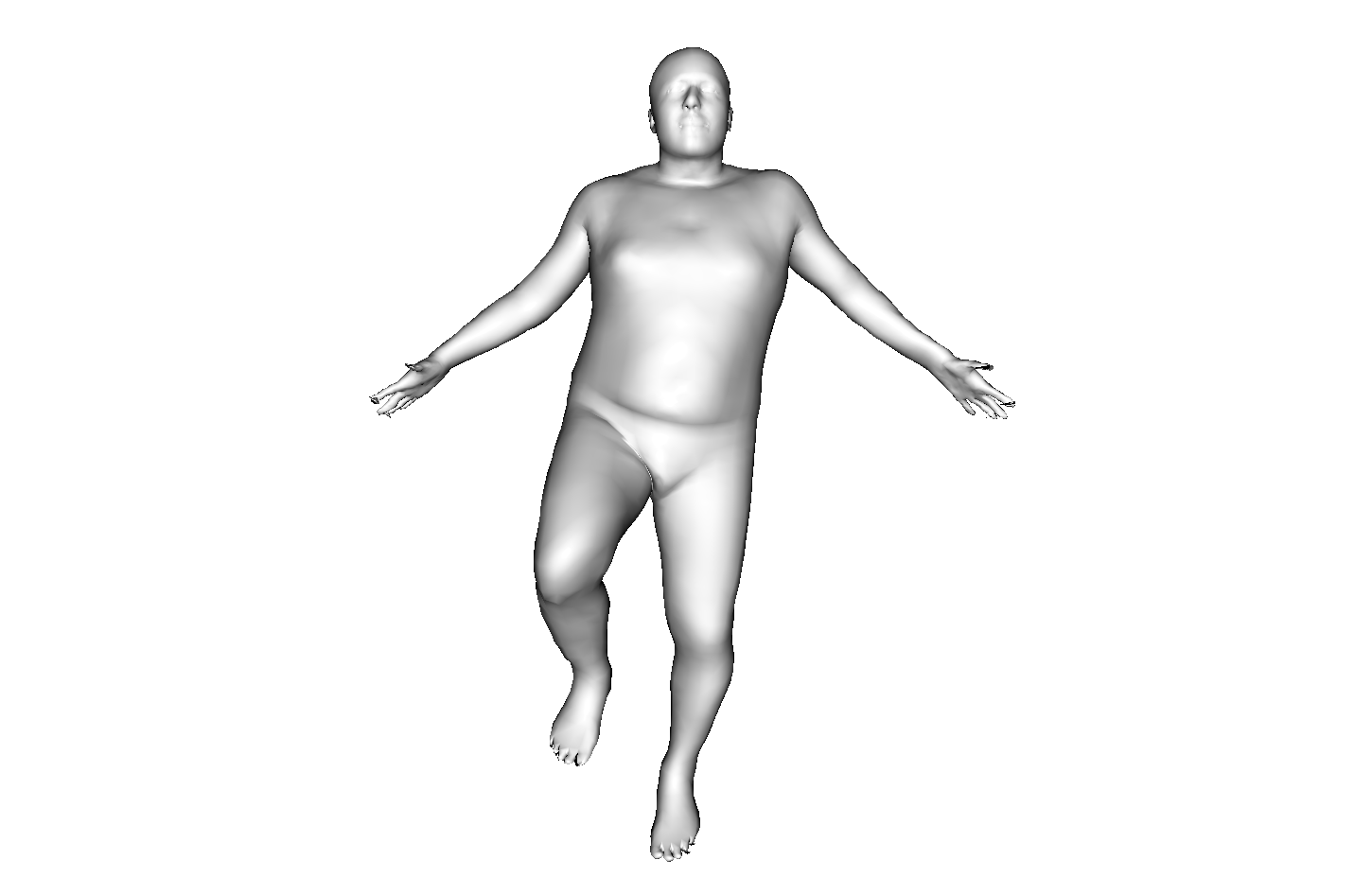}
&
\includegraphics[trim={300 0 300 0},clip,height=3.5cm]
{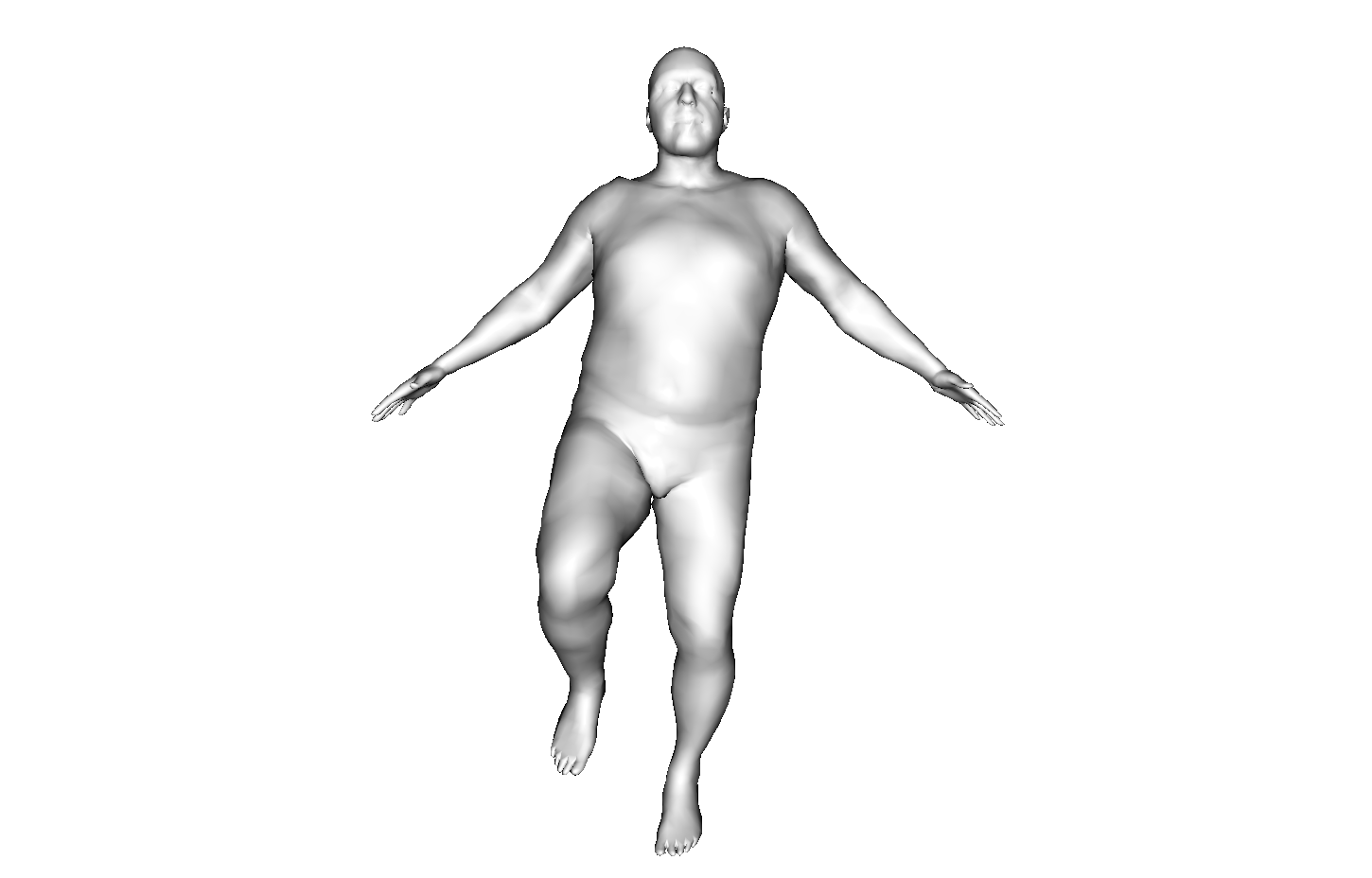}
&
\includegraphics[trim={300 0 300 0},clip,height=3.5cm]
{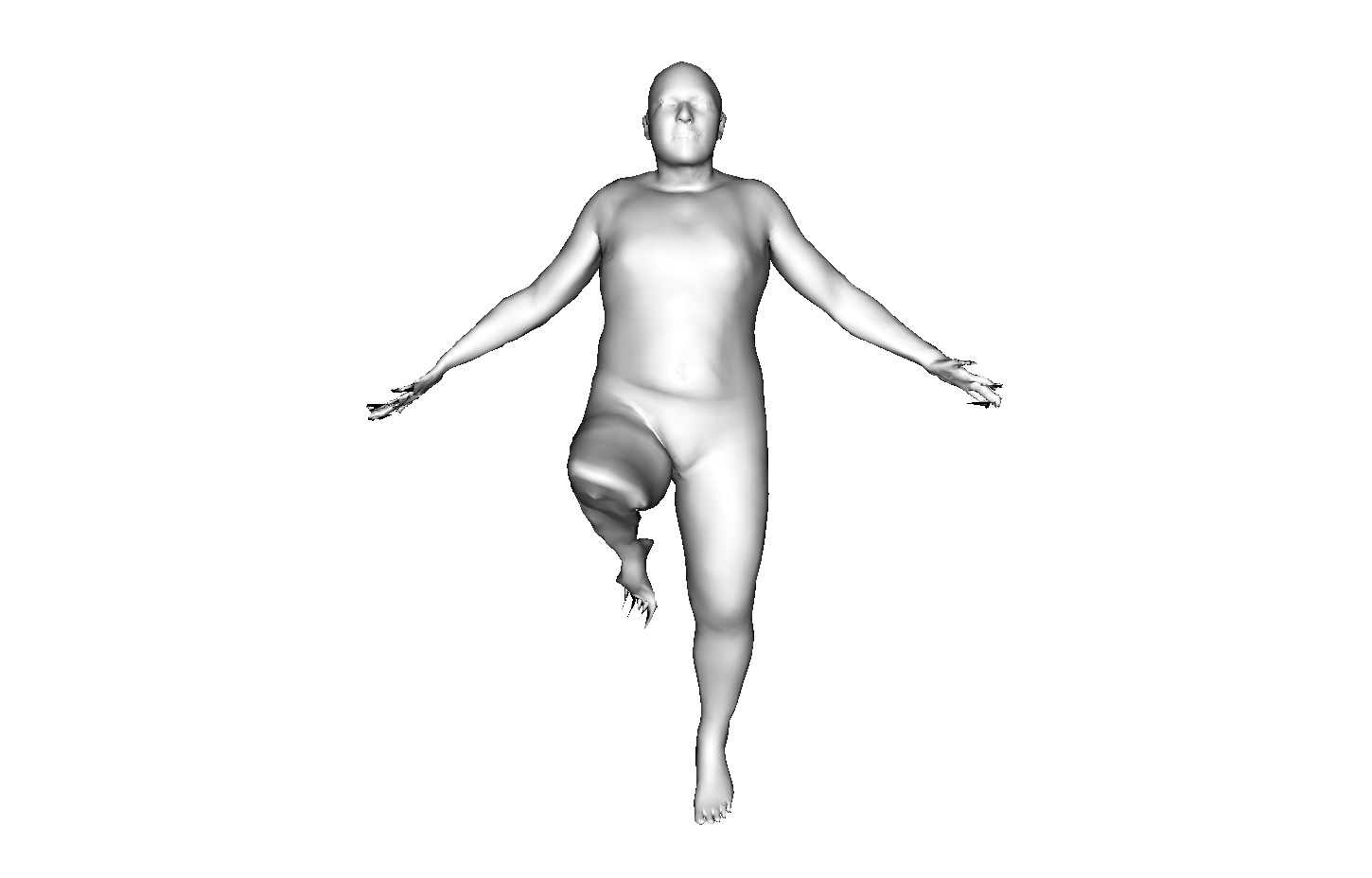}
&
\includegraphics[trim={300 0 300 0},clip,height=3.5cm]
{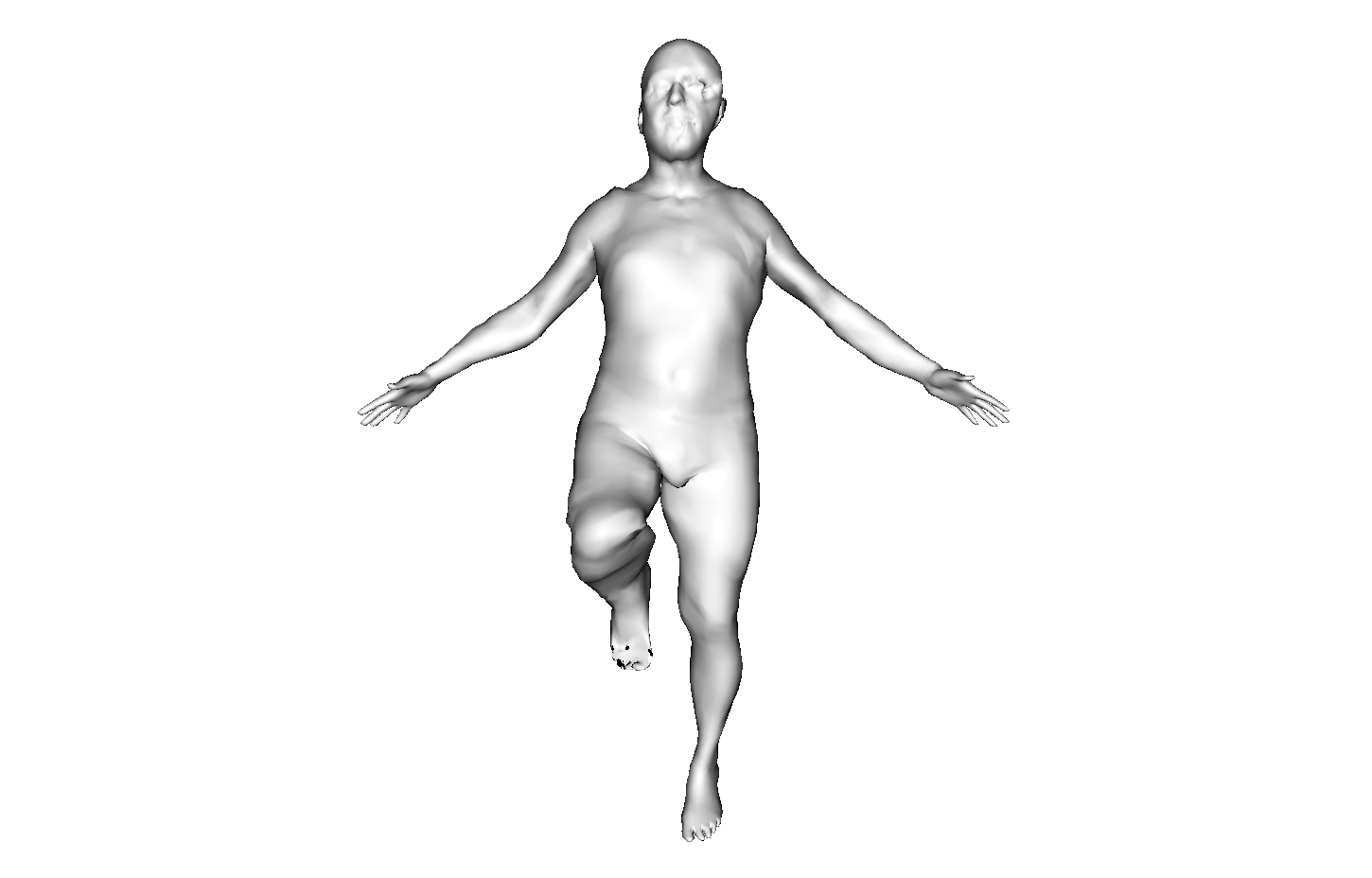}

\end{tabular}
}
\caption{Qualitative Results on CA.}
\label{fig:ca}
\end{table}

\begin{table}
\centering
\resizebox{\columnwidth}{!}{
\begin{tabular}{c|c|c|c}
Depth & DEMEA & CA & FCA \\\hline
\includegraphics[trim={200 50 100 0},clip,height=5cm]
{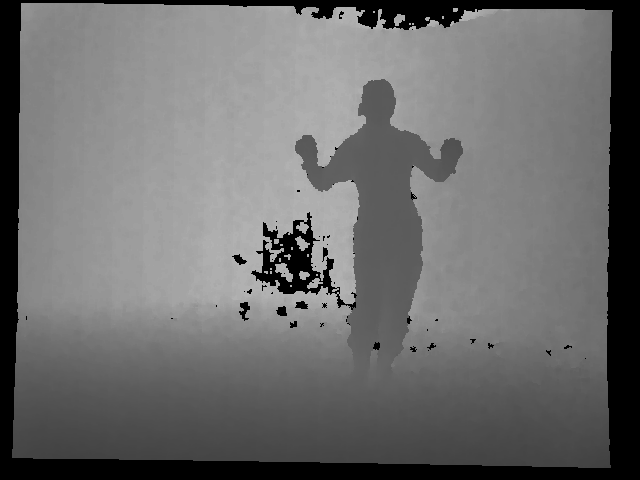}
&
\includegraphics[trim={400 50 400 0},clip,height=5cm]
{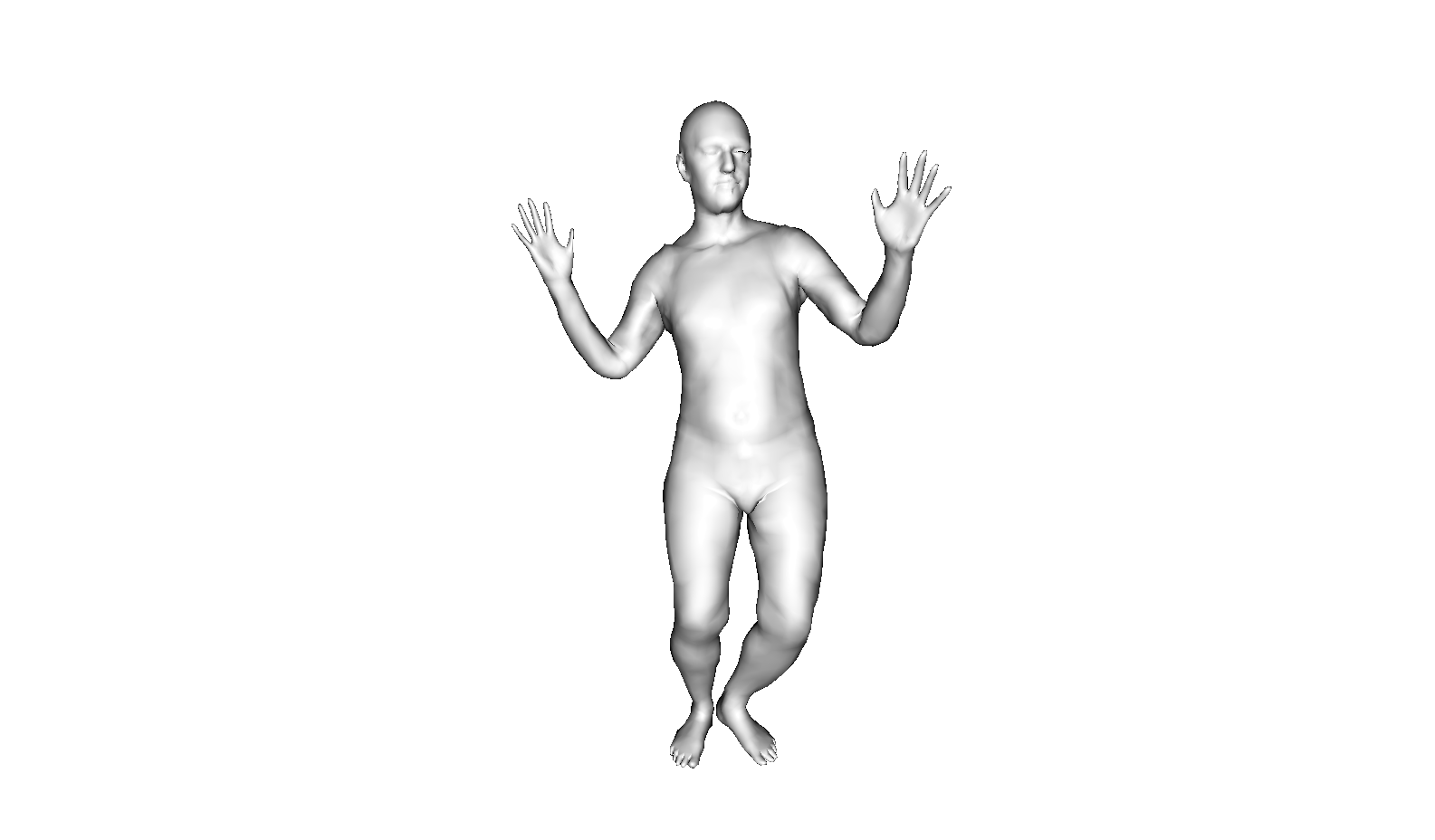}
&
\includegraphics[trim={400 50 400 0},clip,height=5cm]
{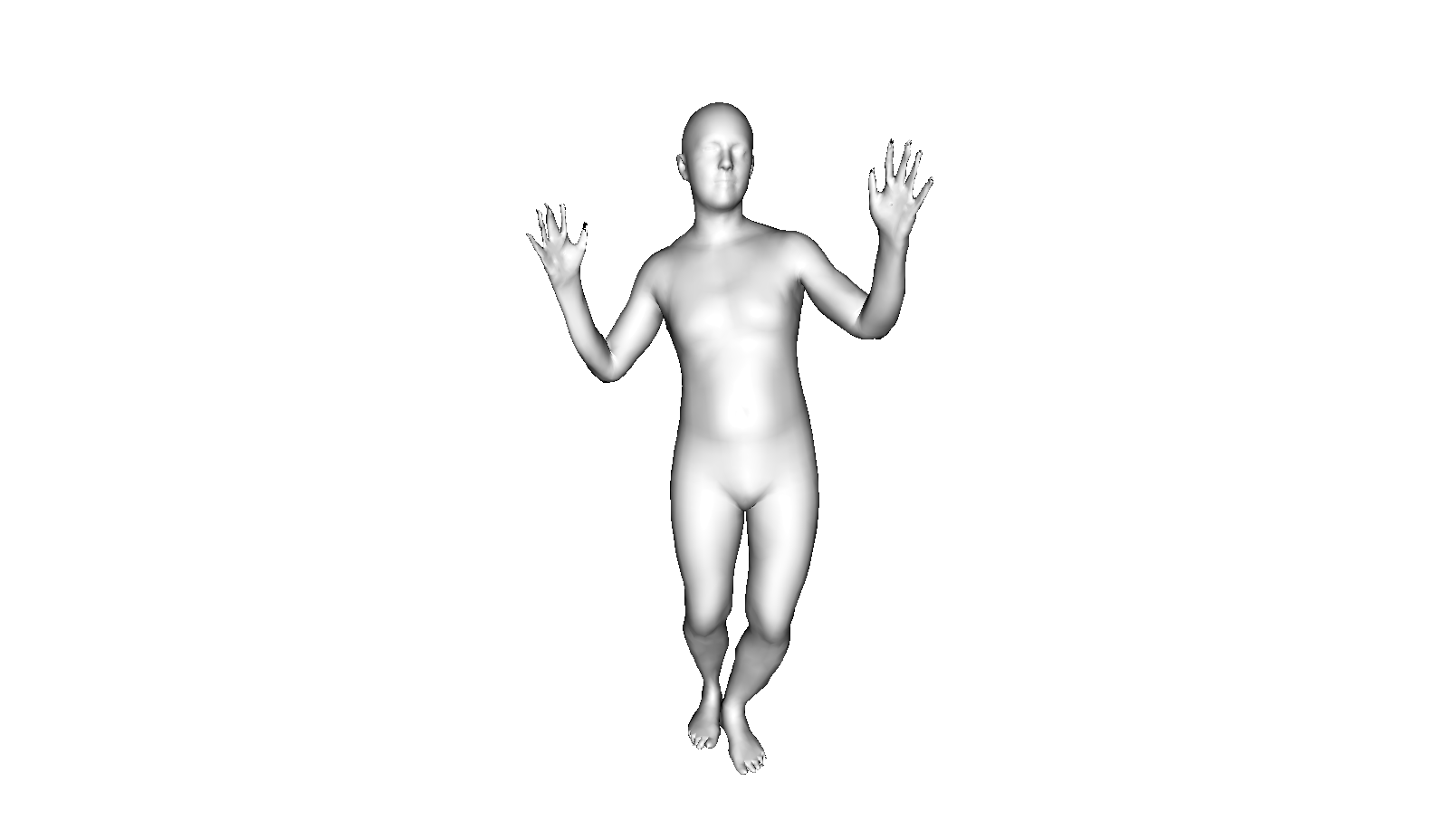}
&
\includegraphics[trim={400 50 400 0},clip,height=5cm]
{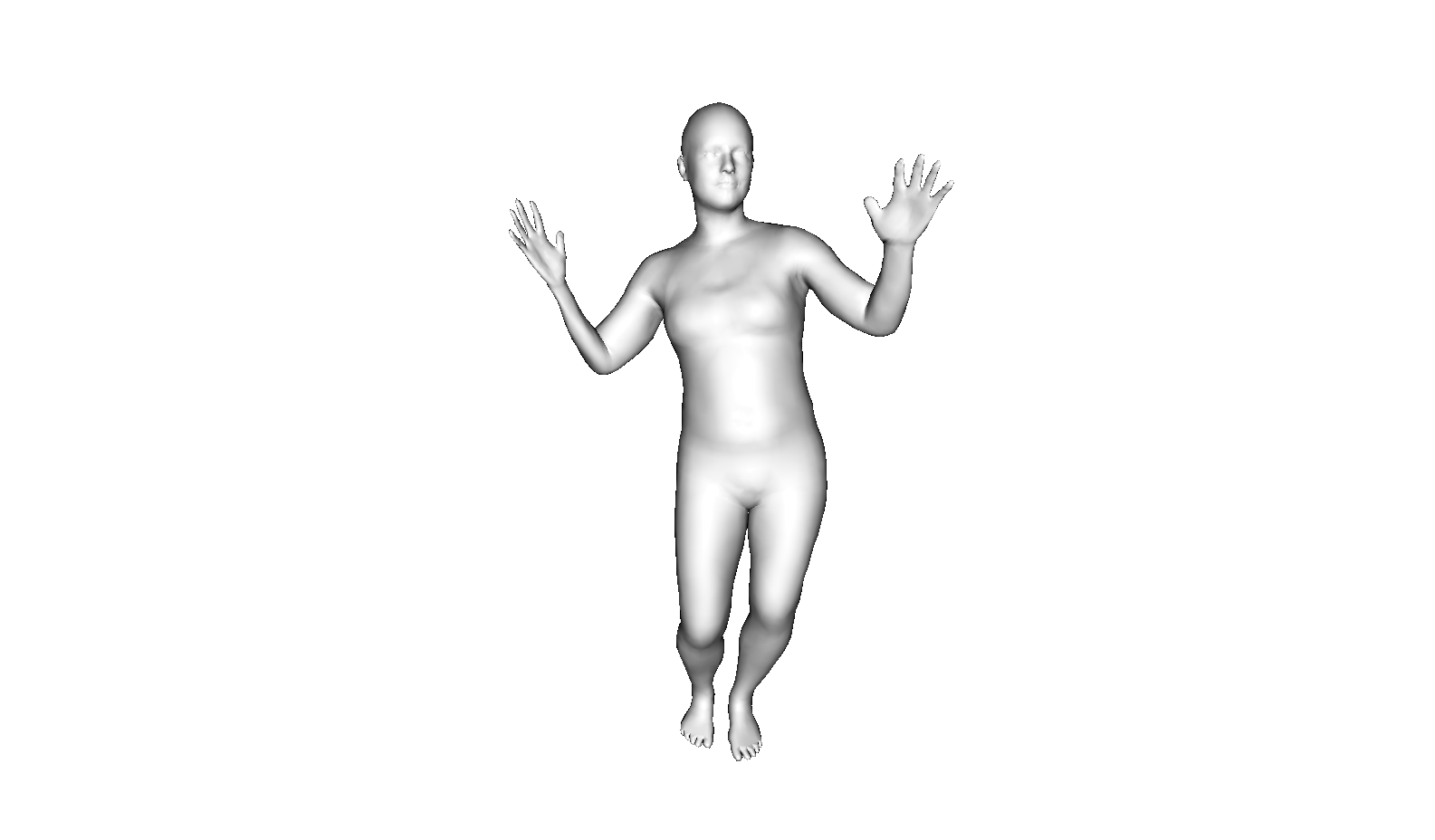}
\\
\includegraphics[trim={200 50 100 0},clip,height=5cm]
{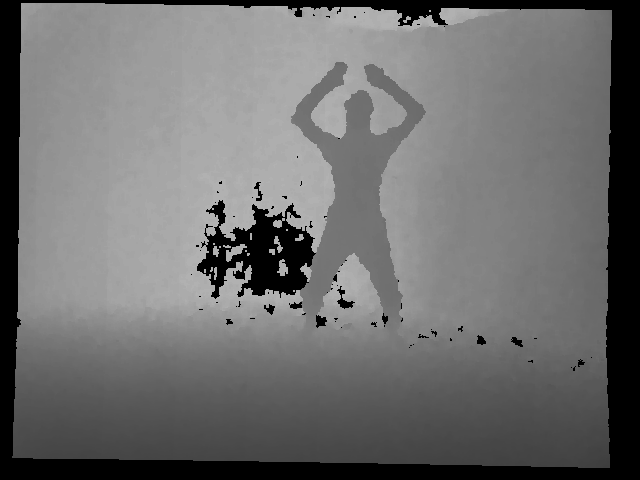}
&
\includegraphics[trim={400 50 400 0},clip,height=5cm]
{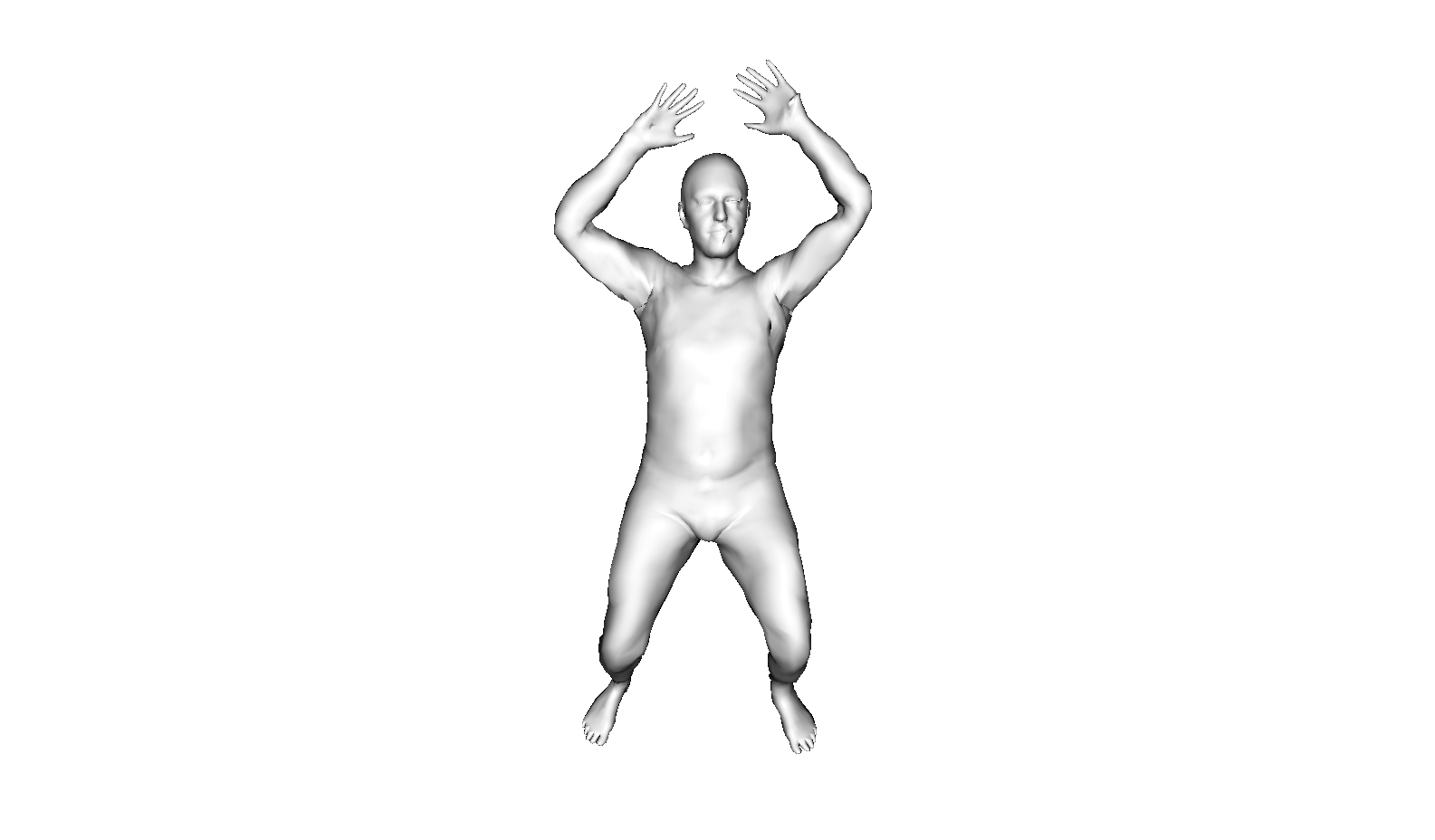}
&
\includegraphics[trim={400 50 400 0},clip,height=5cm]
{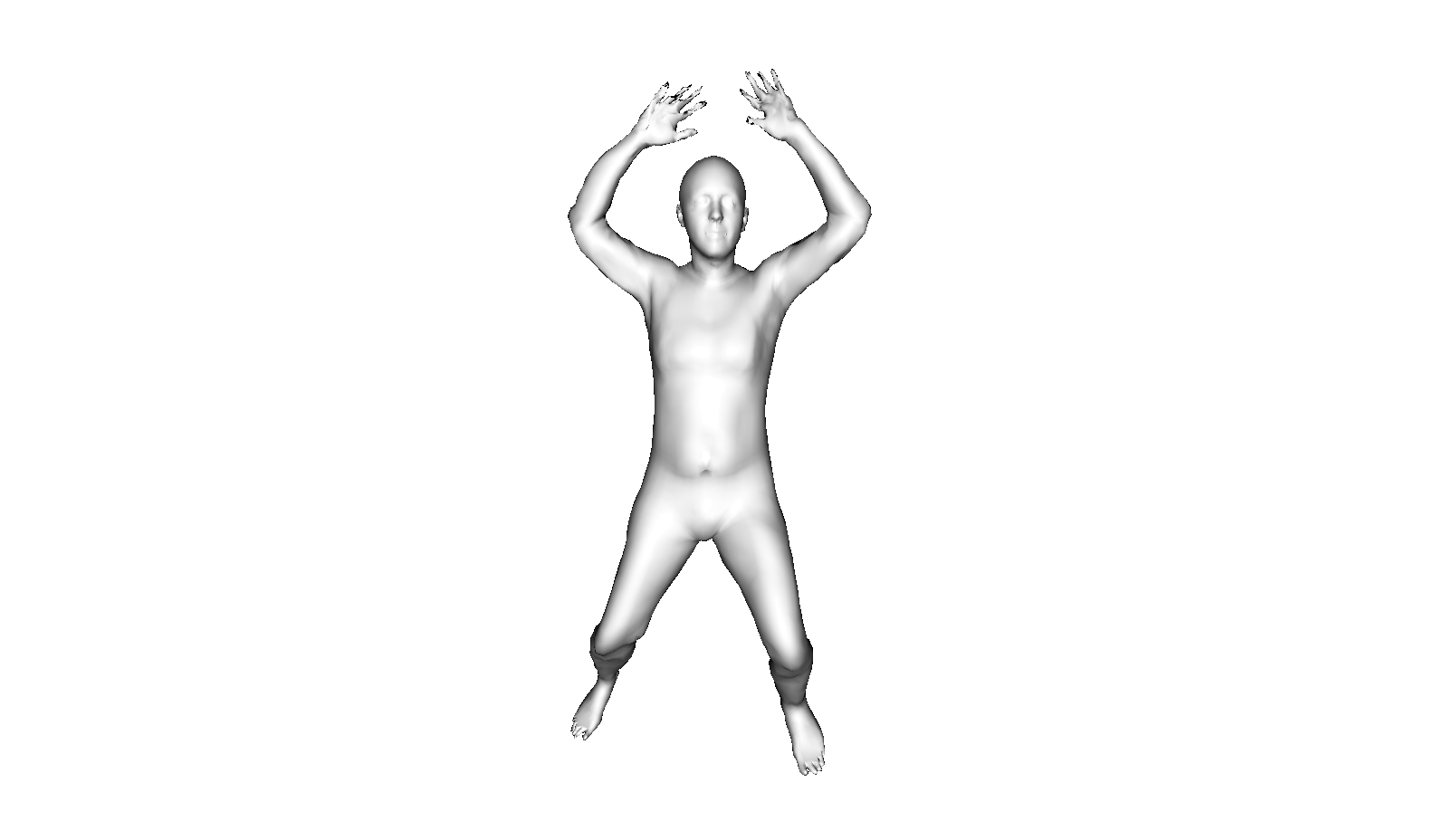}
&
\includegraphics[trim={400 50 400 0},clip,height=5cm]
{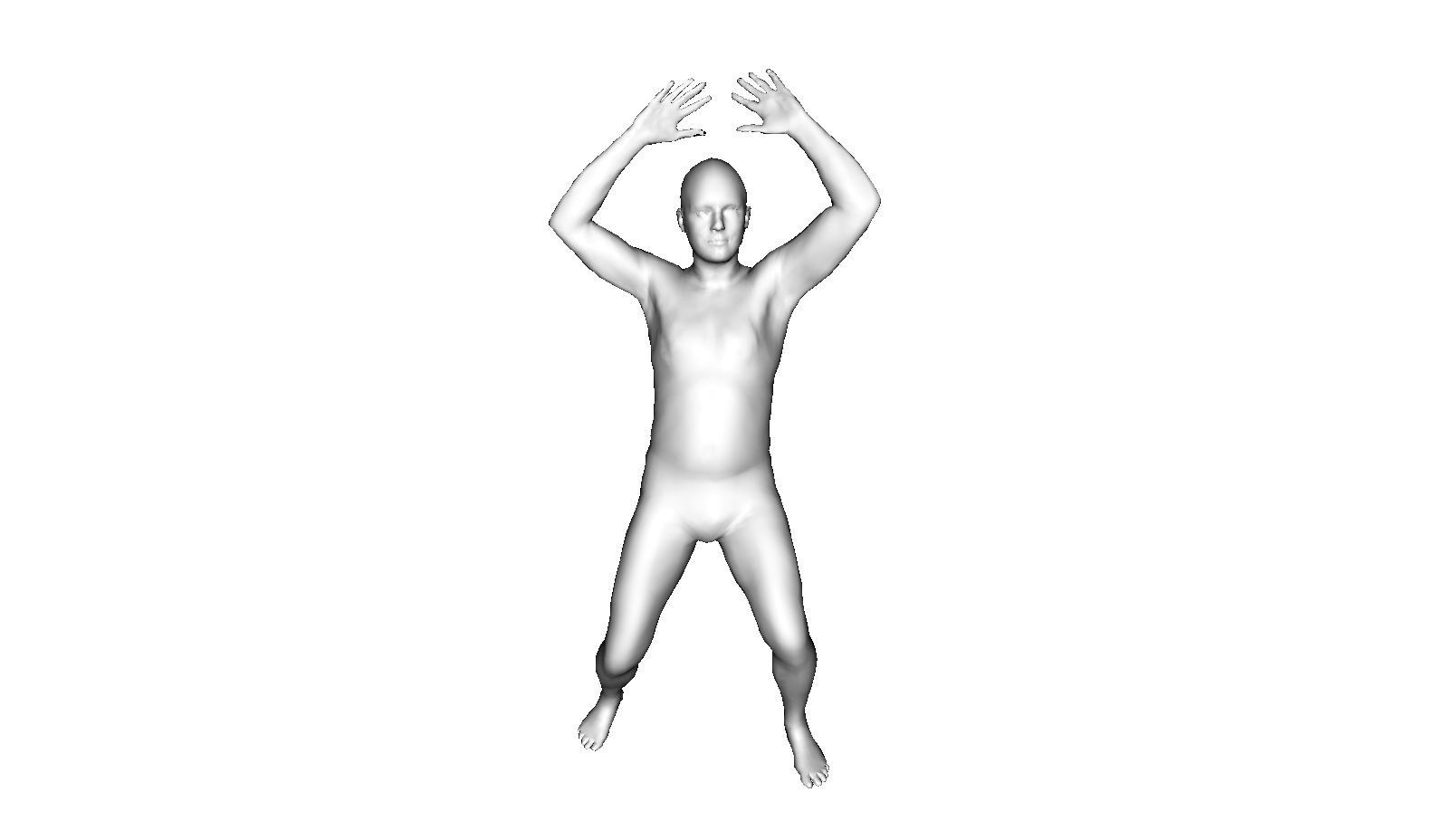}
\\
&
\includegraphics[trim={300 150 200 100},clip,height=3.5cm]
{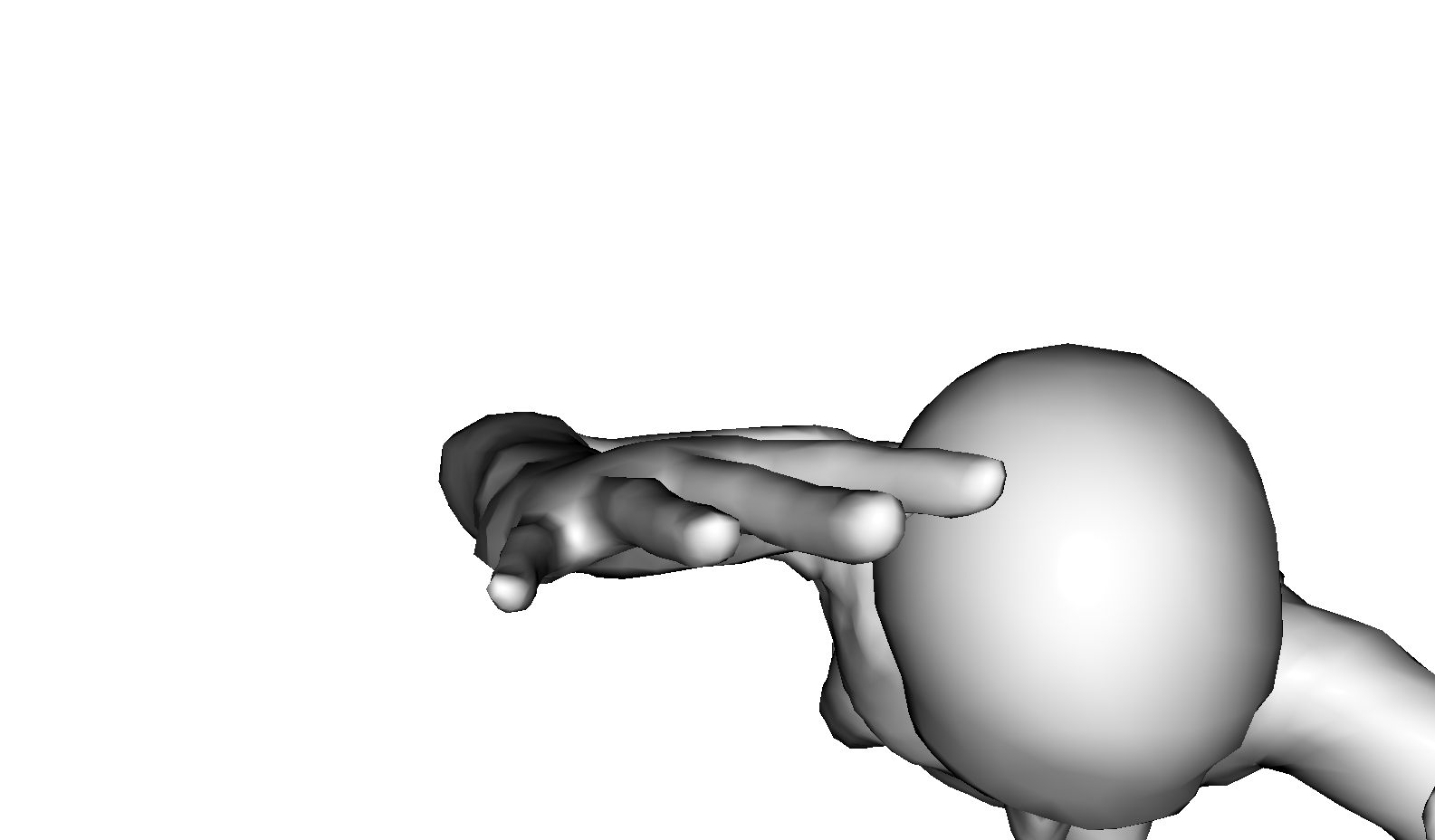}
&
\includegraphics[trim={300 150 200 100},clip,height=3.5cm]
{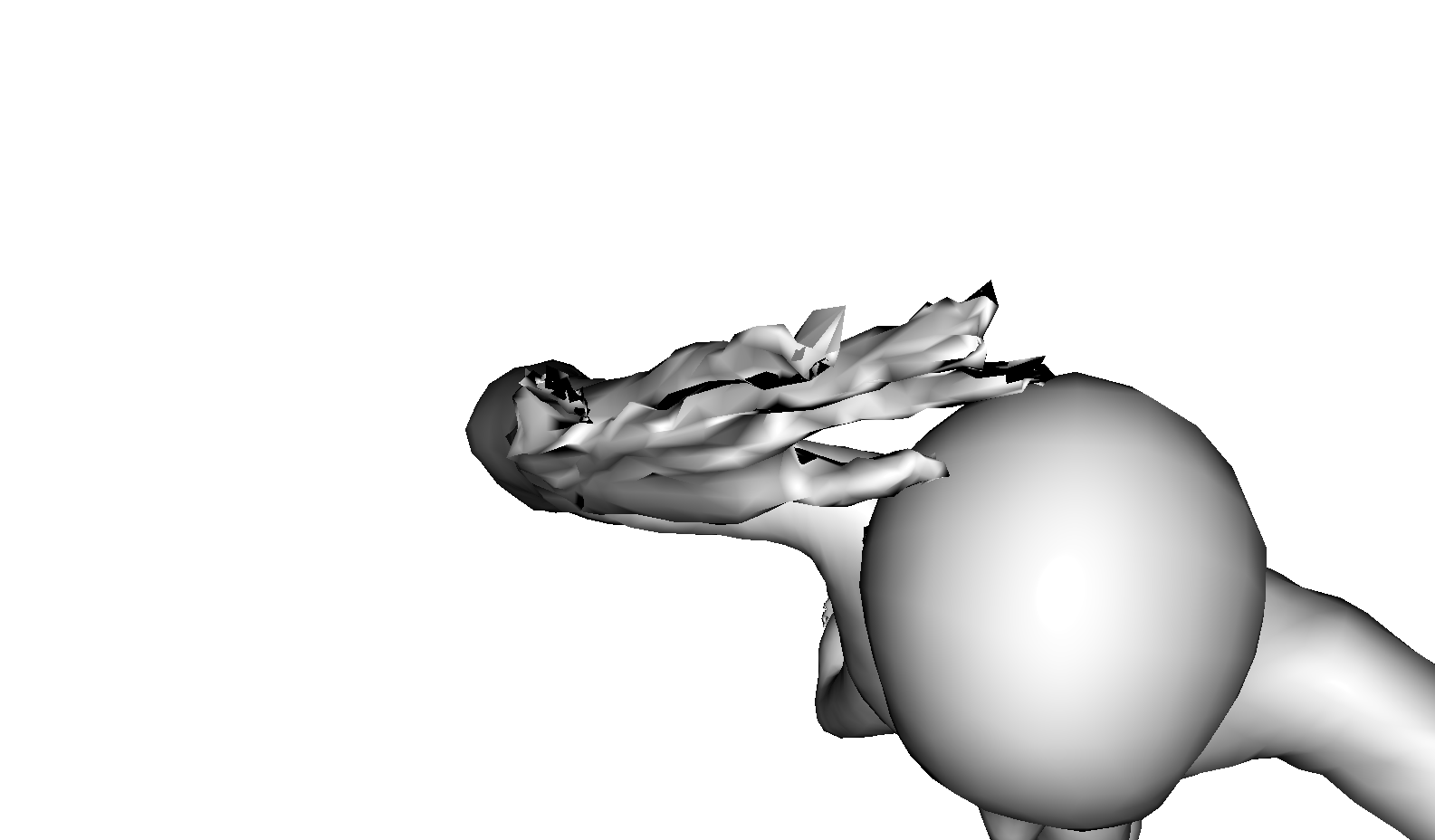}
&
\includegraphics[trim={300 150 200 100},clip,height=3.5cm]
{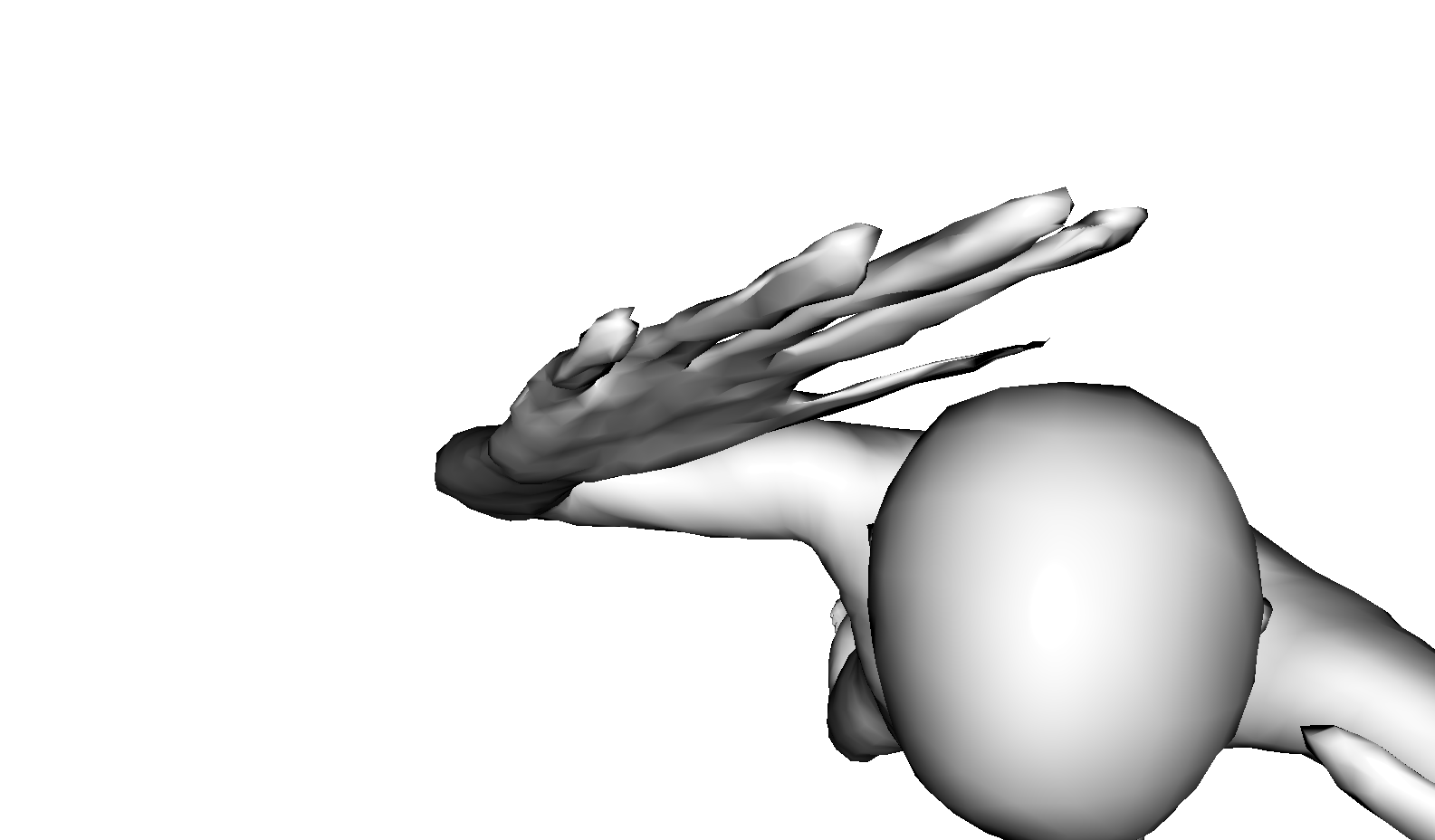}
\end{tabular}
}
\caption{Depth-to-Mesh Results for CA and FCA. The bottom row shows artifacts that DEMEA avoids.}
\label{fig:d2mbase}
\end{table}

\section{Coarse Embedded Graphs}\label{sec:coarse}

In Fig.~\ref{fig:coarse}, we show how an embedded graph can lead to over-smoothing and a loss of detail.

\begin{table}
\centering
\resizebox{\columnwidth}{!}{
\begin{tabular}{c|c|c}
Second Level & Ground-truth & First Level \\\hline
\includegraphics[trim={400 50 300 100},clip,height=3.5cm]
{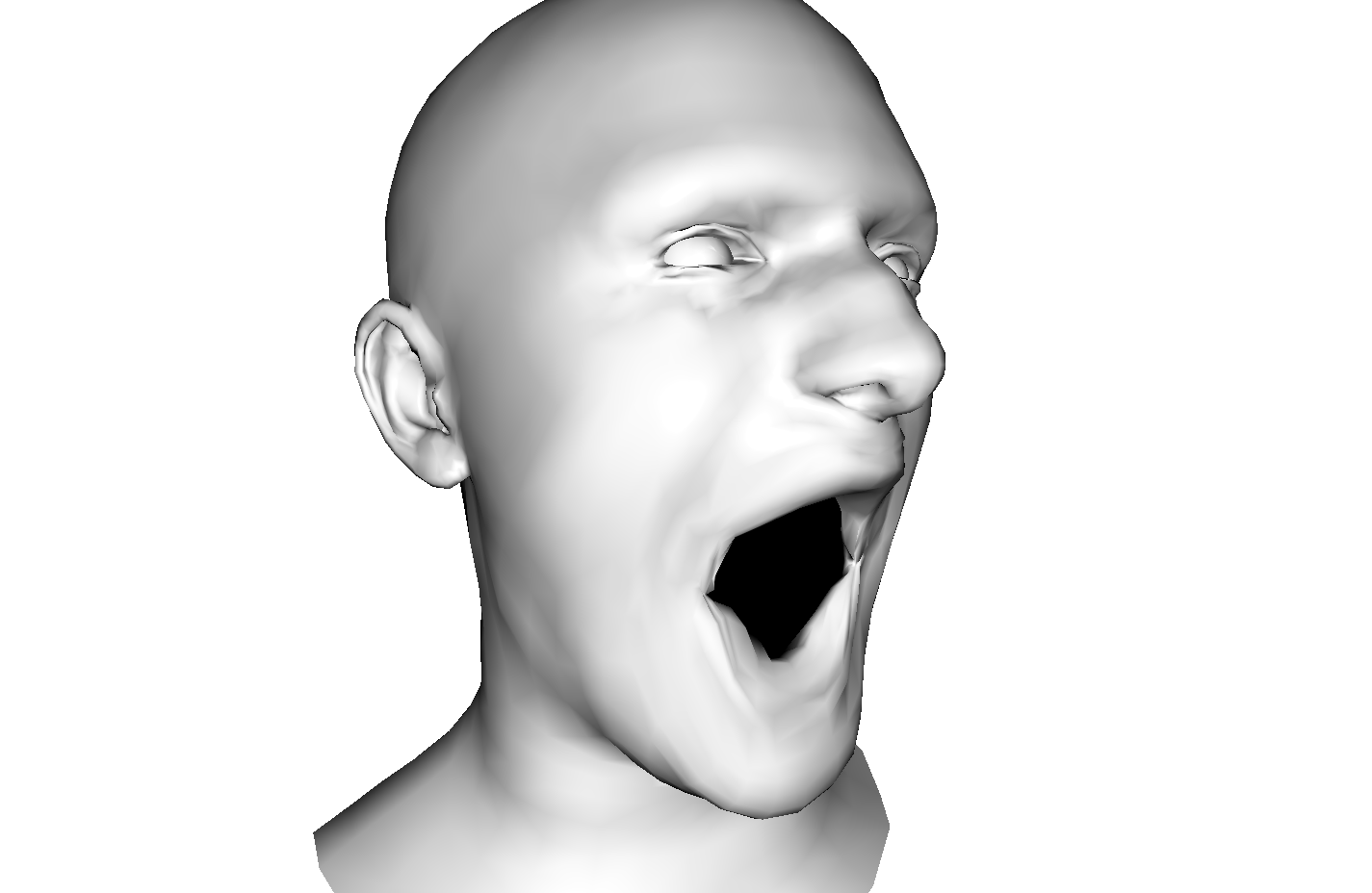}
&
\includegraphics[trim={400 50 300 100},clip,height=3.5cm]
{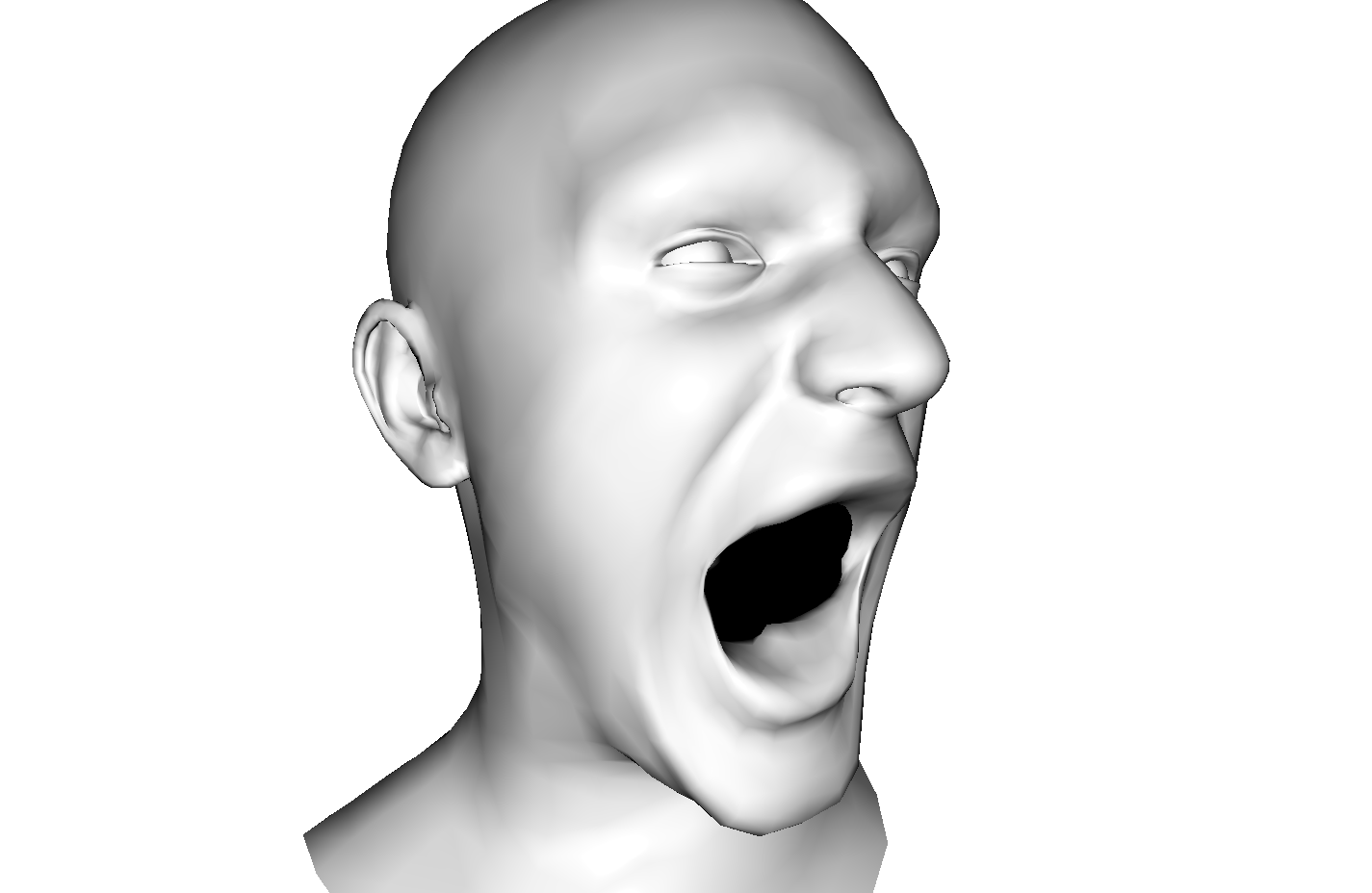}
&
\includegraphics[trim={400 50 300 100},clip,height=3.5cm]
{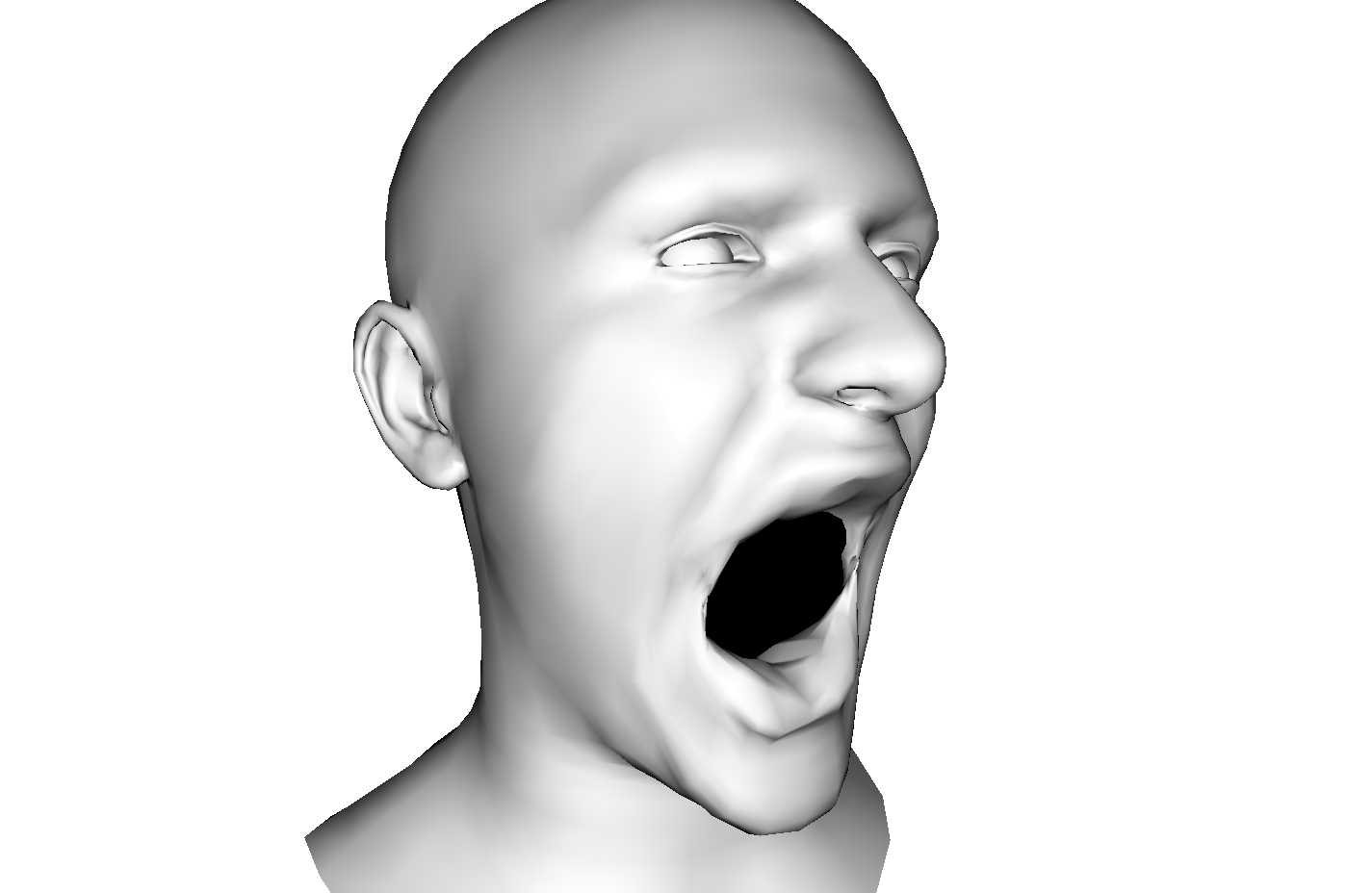}
\end{tabular}
}
\caption{Coarse Embedded Graphs. Note the lips. An embedded graph on the second level of the mesh hierarchy instead of the first level can lead to over-smoothing.}
\label{fig:coarse}
\end{table}

\end{document}